\documentclass[preprint,times,5p,authoryear]{elsarticle}

\usepackage{amssymb}
\usepackage{amsmath}
\usepackage{url}
\usepackage{color}
\usepackage{algorithm} 
\usepackage{algorithmic} 
\usepackage{amsmath,amssymb,amsfonts} 
\usepackage{mathrsfs} 
\usepackage{graphicx}
\usepackage{textcomp}
\usepackage{relsize} 
\usepackage{cuted}
\usepackage{dblfloatfix} 
\usepackage[bookmarks,bookmarksnumbered]{hyperref} 
\hypersetup{colorlinks = true,
    linkcolor=blue,
    filecolor=magenta,      
    urlcolor=cyan}
\usepackage[labelfont=bf]{caption} 
\usepackage[nameinlink,capitalize]{cleveref} 
\usepackage{flushend} 
\usepackage{booktabs} 
\usepackage[bb=boondox]{mathalfa} 
\usepackage{float} 
\usepackage{makecell} 
\usepackage{setspace}
\usepackage[english]{babel}	
\usepackage{color, colortbl}
\usepackage{nicematrix} 
\usepackage{arydshln} 
\usepackage{forest} 
\usepackage{multirow} 
\usepackage{pgfplots} 
\usetikzlibrary{secdia}
\usepackage{tikz}
\usetikzlibrary{patterns} 
\pgfplotsset{compat=1.9}
\usepackage[labelformat=simple]{subcaption}
\usepgfplotslibrary{groupplots}
\usepackage{tkz-kiviat}
\usepackage{pgfplotstable} 
\usepackage{enumitem} 
\usepackage[bottom]{footmisc} 
\biboptions{sort&compress} 
\usepackage{threeparttable} 

\usepackage{ifthen}
\usetikzlibrary{matrix,calc}

\def\classNames{{"Class 0","Class 1","Class 2","Class 3","Class 4"}} 
\def\numClasses{5} 
\def\myScale{0.83} 

\definecolor{sblue}{rgb}{0, 0.44, 1}
\definecolor{lightgray}{rgb}{0.94, 0.94, 0.94}
\definecolor{lightskyblue}{rgb}{0.65, 0.85, 0.9}
\definecolor{inchworm}{rgb}{0.7, 0.93, 0.36} 
\definecolor{horg}{rgb}{1.0, 0.27, 0.0} 

\definecolor{txt}{HTML}{008A49}

\definecolor{myDarkBlue}{HTML}{08306B} 

\definecolor{1a}{rgb}{0.15, 0.73, 0.62}
\definecolor{1b}{rgb}{1.00, 1.00, 0.24}
\definecolor{1c}{rgb}{0.43, 0.36, 0.88}
\definecolor{1d}{rgb}{1.00, 0.36, 0.29}
\definecolor{1e}{rgb}{0.16, 0.64, 0.97}
\definecolor{1f}{rgb}{1.00, 0.58, 0.11}
\definecolor{1g}{rgb}{0.60, 0.89, 0.09}
\definecolor{1h}{rgb}{1.00, 0.39, 0.70}
\definecolor{1i}{rgb}{0.71, 0.71, 0.71}
\definecolor{1j}{rgb}{0.76, 0.20, 0.77}
\definecolor{1k}{rgb}{0.60, 0.59, 0.23}
\definecolor{1l}{rgb}{1.00, 0.73, 0.00}

\definecolor{l1}{rgb}{0.15, 0.73, 0.62} 
\definecolor{l2}{rgb}{0.94, 0.47, 0.00} 
\definecolor{l3}{rgb}{0.29, 0.41, 0.94} 
\definecolor{l4}{rgb}{0.87, 0.18, 0.38} 
\definecolor{l5}{rgb}{0.10, 0.61, 0.47} 
\definecolor{l6}{rgb}{0.98, 0.75, 0.13} 
\definecolor{l7}{rgb}{0.63, 0.13, 0.94} 
\definecolor{l8}{rgb}{0.75, 0.22, 0.17} 
\definecolor{l9}{rgb}{0.00, 0.50, 0.75} 
\definecolor{l10}{rgb}{0.83, 0.00, 0.00} 
\definecolor{l11}{rgb}{0.38, 0.83, 0.31} 
\definecolor{l12}{rgb}{0.91, 0.41, 0.16} 
\definecolor{l13}{rgb}{0.45, 0.00, 0.72} 

\definecolor{acc}{HTML}{FE4A49}
\definecolor{pre}{HTML}{FED766}
\definecolor{rec}{HTML}{009FB7}
\definecolor{f1}{HTML}{9E768F}

\definecolor{cc1}{HTML}{FE4A49}
\definecolor{cc2}{HTML}{FED766}
\definecolor{cc3}{HTML}{009FB7}
\definecolor{cc4}{HTML}{9E768F}

\newcommand\eatpunct[1]{} 
\addto\extrasenglish{

} 
\addto\captionsenglish{} 

\begin{document}

\begin{titlepage}
\begin{center}

\vspace*{10pt}
\doublespacing
{\Large H-SemiS: Hierarchical Fusion of Semi and Self-Supervised Learning for \\ Knee Osteoarthritis Severity Grading}
\vspace*{15pt}

Chandravardhan Singh Raghaw$^{a}$,
Anushka Parwal$^{b}$, 
Shahid Shafi Dar$^{a}$,
Prajakta Darade$^{a}$,
Nagendra Kumar$^{a}$

\vspace{15pt}

\small  
$^a$Department of Computer Science and Engineering, Indian Institute of Technology Indore, Indore 453552, India \\
$^b$Department of Computer Science and Engineering, National Institute of Technology Tiruchirappalli, Tiruchirappalli 620015, India \\
\end{center}
\vspace{40pt}
\begin{flushleft}
{\large Highlights}

\begin{itemize}
    \item A semi-supervised framework with self-supervision for knee osteoarthritis
    \item Hierarchical mechanism decompose multi-class into binary classes to address imbalance
    \item Adversarial-inspired self-supervision enhance semi-supervise teacher-student learning
    \item Quantum principles capture nonlinear knee patterns to enhance feature discrimination
    \item Comprehensive experiments validate the superiority of our H-SemiS framework
\end{itemize}

\vspace{50pt}

\noindent This is the preprint version of the accepted paper.\\
\noindent This paper is accepted in \textbf{Expert Systems with Applications, 2026.}
\\
DOI: \url{https://doi.org/10.1016/j.eswa.2026.132279}

\end{flushleft}
\end{titlepage}


\begin{frontmatter}

\title{H-SemiS: Hierarchical Fusion of Semi and Self-Supervised Learning for \\ Knee Osteoarthritis Severity Grading}

\date{}

\author[1]{Chandravardhan Singh Raghaw}
\ead{phd2001201007@iiti.ac.in}

\author[2]{Anushka Parwal}
\ead{anushkaparwal.nitt@gmail.com}

\author[1]{Shahid Shafi Dar}
\ead{phd2201201004@iiti.ac.in}

\author[1]{Prajakta Darade}
\ead{cse210001052@alum.iiti.ac.in}

\author[1]{Nagendra Kumar\corref{cor1}}
\cortext[cor1]{Corresponding author email: \textit{nagendra@iiti.ac.in}}


\address[1]{Department of Computer Science and Engineering, Indian Institute of Technology (IIT) Indore, Indore 453552, India}
\address[2]{Department of Computer Science and Engineering, National Institute of Technology Tiruchirappalli, Tiruchirappalli 620015, India}

\begin{abstract}
Knee osteoarthritis (KOA) is a degenerative joint disease that can lead to chronic pain, reduced mobility, and long-term disability. Automated severity grading from knee radiographs can support early assessment, but current methods heavily depend on large labeled datasets and remain sensitive to class imbalance, noisy samples, and variability in clinical annotations. To alleviate these limitations, we propose a \textbf{\underline{H}}ierarchical fusion of \textbf{\underline{Semi}}-Supervised framework with \textbf{\underline{S}}elf-Supervision (H-SemiS) for KOA severity grading in knee X-ray samples using limited annotated data. Rather than treating severity grading as a flat multi-class problem, H-SemiS decomposes the task into a sequence of binary sub-tasks within a semi-supervised teacher–student architecture, directly mitigating the impact of class imbalance. To further enhance feature learning from unlabeled data, the framework integrates an adversarial self-supervised reconstruction module that encourages the network to capture robust anatomical structures. In parallel, a teacher–student design with quantum-inspired feature mixing improves discrimination boundaries between adjacent grades when pseudo-labels are noisy. We comprehensively evaluate H-SemiS on two challenging multi-class datasets and assess its generalizability on two binary-class datasets. Our experimental results demonstrate the superiority of the proposed H-SemiS framework across multiple evaluation metrics, consistently outperforming several competing baselines and state-of-the-art methods. 
The code is publicly available at \url{https://github.com/chandravardhan-singh-raghaw/H-SemiS}.
\end{abstract}

\end{frontmatter}

\section{Introduction}
\label{sec:introduction}
Knee Osteoarthritis (KOA) is a musculoskeletal disorder where the protective cartilage at the ends of bones gradually breaks down, resulting in chronic pain and functional disability \citep{Wang2025oag}. KOA affects more than 240 million people worldwide, with the highest cases in China, India, the United States, and Japan~\citep{Wang2025oag}. Radiographic signs appear in nearly 80\% of individuals over fifty, and the disease places a substantial burden on middle-aged and older adults, with a higher incidence in women~\citep{Zhuang2023kcd}. Current  KOA treatment includes physiotherapy, medication, and joint arthroplasty~. Physiotherapy strengthens surrounding muscles and reduces joint stress, medications control pain and inflammation, and arthroplasty provides surgical relief but has a limited lifespan and restricts physical activity~\citep{Zhuang2023kcd}. These limitations motivate early diagnosis, for which we adopt the KL grading system~\citep{Kellgren1957} as the reference standard. The KL system categorizes KOA severity into five grades based on radiographic features, as summarized in~\cref{tab:klgrade}.

\begin{table}[!ht] \footnotesize
\setlength{\tabcolsep}{3pt}
\centering
    \caption{Description of Kellgren-Lawrence (KL) system for grading severity}
    \begin{NiceTabular*}{\columnwidth}{@{\extracolsep{\fill}} l l p{0.65\columnwidth}}    
        \toprule
        \midrule
        \textbf{KL Grade} & \textbf{Severity} & \textbf{Definition} \\
        \midrule
        KL Grade 0 & None     & No detectable signs of osteoarthritis \\
        KL Grade 1 & Doubtful & Possible osteophytes \\
        KL Grade 2 & Minimal  & Clear presence of osteophytes with possible joint space narrowing (JSN) \\
        KL Grade 3 & Moderate & Multiple moderate osteophytes, confirmed JSN, and early sclerosis \\
        KL Grade 4 & Severe   & Large osteophytes, definite JSN, severe sclerosis \\        
        \midrule
        \bottomrule
    \end{NiceTabular*}
\label{tab:klgrade} 
\end{table}

\subsection{Challenges in Knee Osteoarthritis Diagnosis}
\begin{figure*}[!ht]
    \centering
    \includegraphics[width=\textwidth]{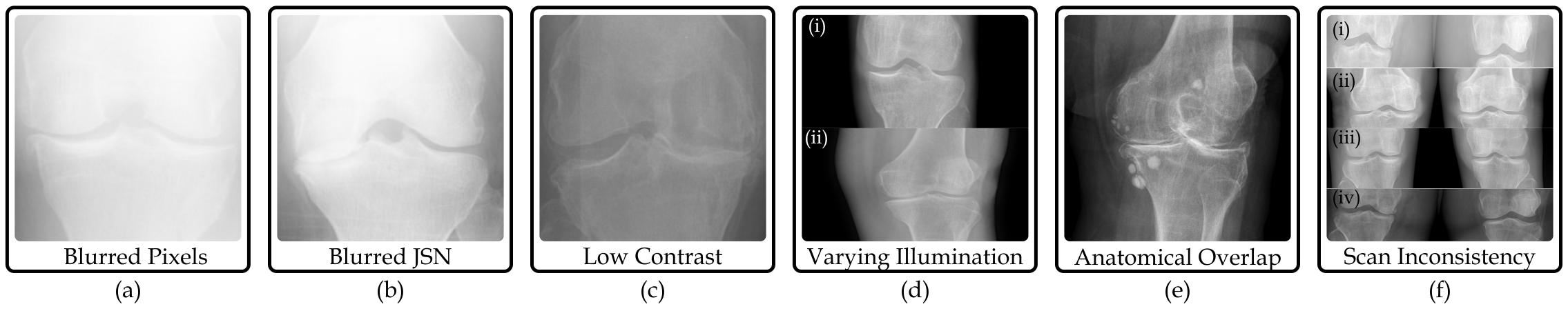}
    \caption{Diagnostic challenges in KXR analysis for KOA are highlighted from (a) to (f), including blurred pixels, indistinct joint space narrowing (JSN), low contrast, varying illumination, overlapping anatomical structures, and inconsistencies in scan composition (single or dual joint).}
    \label{fig:ch-kxr}
\end{figure*}

The diagnosis of KOA depends on radiologist expertise, which is time-consuming and varies across practitioners~\citep{Pan2024autok}. Computer-aided diagnosis (CAD) provides a practical alternative for grading KOA severity from knee X-rays (KXR), yet several challenges in KXR, illustrated in \cref{fig:ch-kxr}, limit accurate assessment. Blurred regions around joint space narrowing (JSN) caused by bone spurs dilute key structural information, while the grayscale nature of KXR, with low contrast and poor illumination, limits the visibility of subtle variations. Overlapping anatomical structures, including cartilage and surrounding soft tissues, further complicate the extraction of discriminative patterns. Dataset inconsistency also affects performance, as some scans include both knees, whereas most capture only one, resulting in less uniform feature extraction and interpretation.

\begin{figure*}[!b]
    \centering
    \includegraphics[width=\textwidth]{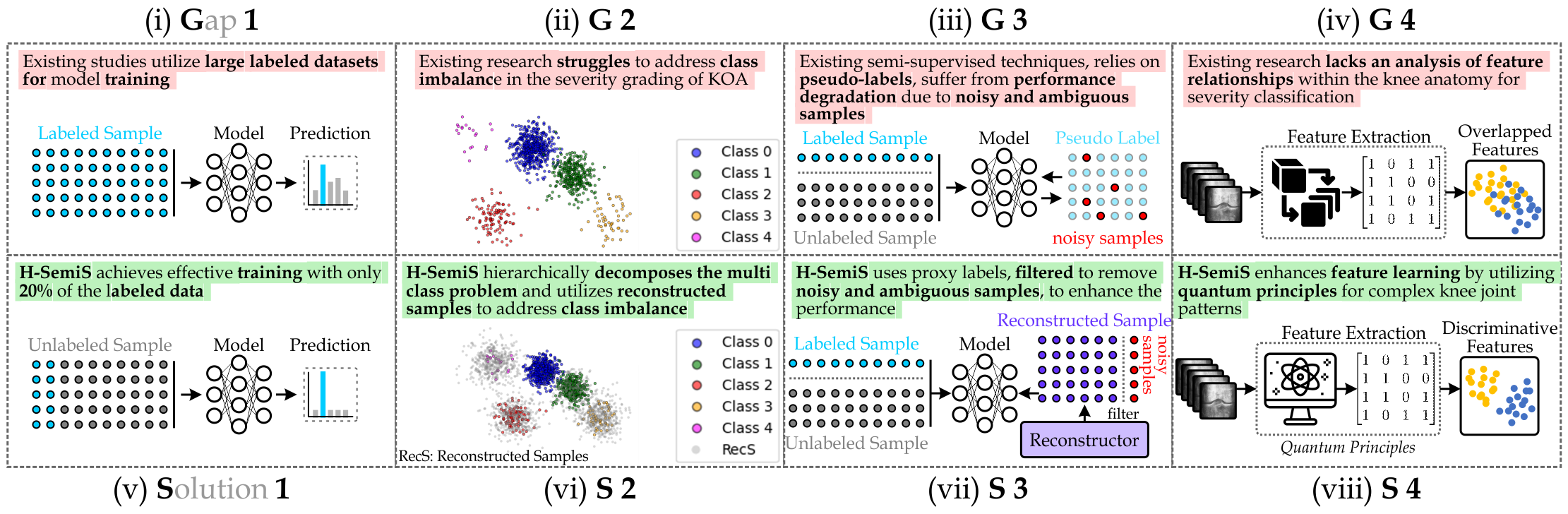}
    \caption{Gaps in existing research on severity grading of knee osteoarthritis are highlighted in the first row (i) to (iv). The second row (v) to (viii) presents the solutions offered by the proposed H-SemiS framework.}
    \label{fig:ch-exis}
\end{figure*}

\subsection{Existing Research and Gaps}
Several machine learning approaches have been explored for KOA diagnosis from KXR~\citep{Sohail2025tl, SCBose2024, Du2018ml}, but noise and variability in radiographs limit performance and generalizability. To address this, researchers have introduced supervised methods~\citep{Nguyen2024mat, Lo2023dlseptic, Teoh2023dhl, Hu2022enn} that capture complex knee patterns. Despite their potential, supervised techniques rely on large, labeled datasets, which are costly and time-consuming to obtain. To reduce the dependency on labeled data, semi-supervised methods~\citep{Wu2023, Azizi2023culp, Cai2023, Azizi2021} leverage limited labeled samples for model training. While semi-supervised methods demonstrate comparable performance with fewer labels, they still depend on pseudo-labels. This dependency leads to incorrect predictions because of noisy and ambiguous samples, which compromise the overall performance and consistency. The limited availability of labeled data restricts semi-supervised methods from capturing the fine-grained features needed for robust and discriminative knee anatomy analysis.

\subsection{Research Aim}
We designed this study to address diagnostic challenges in knee X-ray (KXR) samples and the gaps identified in existing research. To tackle the issues shown in~\cref{fig:ch-kxr}: (a) We preserve JSN by using the Random Forest Regression Voting Constrained Local Model \citep{Lindner2013bonefinder}, which detects knee key points for precise joint space mapping and reduces errors from blurred pixels, ensuring reliable JSN measurements; (b) We apply the contrast enhancement technique~\citep{Raghaw2024xccnet} during preprocessing to improve contrast and illumination, strengthening feature extraction; (c) We integrate quantum principles to address complications from overlapping structures, capturing nonlinear relationships between knee points and improving the learning of complex knee patterns; (d) We maintain consistency across KXR samples by using region-of-interest (ROI) localization~\citep{Redmon2017yolo} to extract knee pairs from bilateral scans, standardizing them into single-knee KXRs.

Second, to resolve the limitations within existing research (refer to~\cref{fig:ch-exis}): While existing methods require extensive labeled datasets (G1), our framework functions effectively with only 20\% labeled samples (S1), reducing annotation overhead. Current research struggles with skewed KOA severity grading (G2); we mitigate this by hierarchically decomposing the multi-class problem and incorporating reconstructed samples to ensure balanced learning (S2). (c) Unlike semi-supervised techniques compromised by noisy pseudo-labels (G3), we leverage filtered proxy labels to eliminate noise and enhance diagnostic stability (S3). (d) Existing studies overlook anatomical feature relationships (G4); in contrast, we map fine-grained features into quantum space to analyze nonlinear interactions, improving severity discrimination (S4).

\subsection{Contribution}
Addressing the discussed challenges, we propose H-SemiS, a \textbf{\underline{H}}ierarchical \textbf{\underline{Semi}}-Supervised framework with \textbf{\underline{S}}elf-Supervision for Knee Osteoarthritis (KOA) severity grading (KL Grades $0{-}4$). The methodological objective is to develop an enhanced KOA assessment framework that can classify severity levels from KL Grade 0 to KL Grade 4. H-SemiS integrates self-supervised and semi-supervised learning, where the former generates high-quality proxy data to directly support the latter, especially when annotated data is limited. H-SemiS performs hierarchical severity classification and incorporates reconstructed samples (proxy labels) to increase feature diversity. The experimental objective is to analyze the quantitative and qualitative performance of H-SemiS and compare it to state-of-the-art models on the OAI~\citep{ChenOAI} and DKXI~\citep{GornaleDKXI} datasets. An ablation study evaluates performance across different modules, and a generalizability analysis examines performance under various training and testing combinations. The main contributions of this research are outlined as follows:

\begin{itemize}
    \item We propose a novel semi-supervised framework incorporating self-supervised learning for grading knee osteoarthritis severity in knee X-ray images with limited annotated samples. 
    \item A dual approach is adopted to address multi-class imbalance. First, we propose a hierarchical mechanism to decompose multi-class problems into multiple binary classes, ensuring balanced learning across all severity levels. Second, we design an adversarial-inspired self-supervised module for masked image reconstruction (MI-Rec) to generate reconstructed samples. These samples are utilized at each decision node in the binary classification to balance sample distribution across severity levels and enhance feature diversity.
    \item We introduce a Similarity-aware Reconstructed Image Labeler (SiRL) that assigns labels to unlabeled reconstructed samples while filtering out noisy and ambiguous samples. This process improves feature discriminability and enhances overall performance.
    \item To the best of our knowledge, we are the first to incorporate quantum principles within semi-supervised learning to learn complex knee patterns despite the presence of overlapping structures, enabling a deeper understanding of knee anatomy.
    \item H-SemiS outperforms thirteen existing baselines across the benchmark datasets, demonstrating an effective and annotation-efficient approach for reliable KOA severity assessment.
\end{itemize}

The paper is organized as follows: \cref{sec:related} reviews the existing literature, and \cref{sec:method} details the proposed methodology. \cref{sec:dexp} outlines the datasets and experimental setup, followed by performance evaluation in \cref{sec:results}. \cref{sec:discussion} addresses significance, limitations, and future work, while \cref{sec:conclusion} concludes the study.

\section{Related Work}
\label{sec:related}
\subsection{Supervised learning-based Classification}
Supervised learning leverages labeled datasets to identify complex patterns and correlations in imaging data. It has gained traction in medical imaging for classification and segmentation. \cite{Nguyen2024mat} introduces CLIMATv2, a dual-transformer framework for one-to-many prognosis prediction using images and clinical data. \cite{Lo2023dlseptic} employs a pre-trained ViT to extract features from knee ultrasound scans for septic arthritis diagnosis. \cite{Teoh2023dhl} explores a CNN with a feature extraction strategy using global average pooling for KOA analysis. \cite{Hu2022enn} proposes A-ENN, integrating evolutionary traces and adversarial training for longitudinal KOA grading. Although supervised models have advanced KOA detection, they require large datasets for effective training and show limited generalizability across diverse datasets. In contrast, H-SemiS captures fine-grained details and complex patterns while relying on fewer labeled samples in a semi-supervised setting.

\subsection{Self-Supervised learning-based Classification}
The limited availability of labeled medical samples poses a major challenge for supervised learning, which self-supervised techniques address by extracting features from unlabeled data~\citep{Huang2023}. Methods such as Masked Autoencoder (MAE)~\citep{He2022mae} and SimCLR~\citep{Chen2020SimCLR} employ pretext tasks, including masked reconstruction and contrastive learning, to derive meaningful representations. \cite{Azizi2023culp} combines supervised transfer learning with contrastive self-supervision to improve data-efficient generalization. \cite{Cai2023} introduces Dual-distribution Discrepancy for Anomaly Detection (DDAD), a one-class semi-supervised learning (OC-SSL) method that models distributions of unlabeled and normal samples using reconstruction networks. \cite{Wu2023} develops a Self-supervised Multimodal Fusion Network (S-MFN), a multimodal self-supervised framework integrating X-ray and MRI for KOA grading. \cite{Wang2022MI-Se} integrates self-supervised and semi-supervised learning to learn complementary information across multi-sequence images. \cite{Azizi2021} introduces the Multi-Instance Contrastive Learning (MICLe) method that improves robustness to distribution shifts. In contrast, H-SemiS distinguishes itself from existing self-supervised methods by integrating generative adversarial learning with a masked autoencoder to reconstruct masked patches without labels. This reconstruction-driven augmentation mitigates class imbalance, enhances feature discriminability, and improves the performance of KOA severity grading.

\begin{figure*}[!b]
    \centering
    \includegraphics[width=\linewidth]{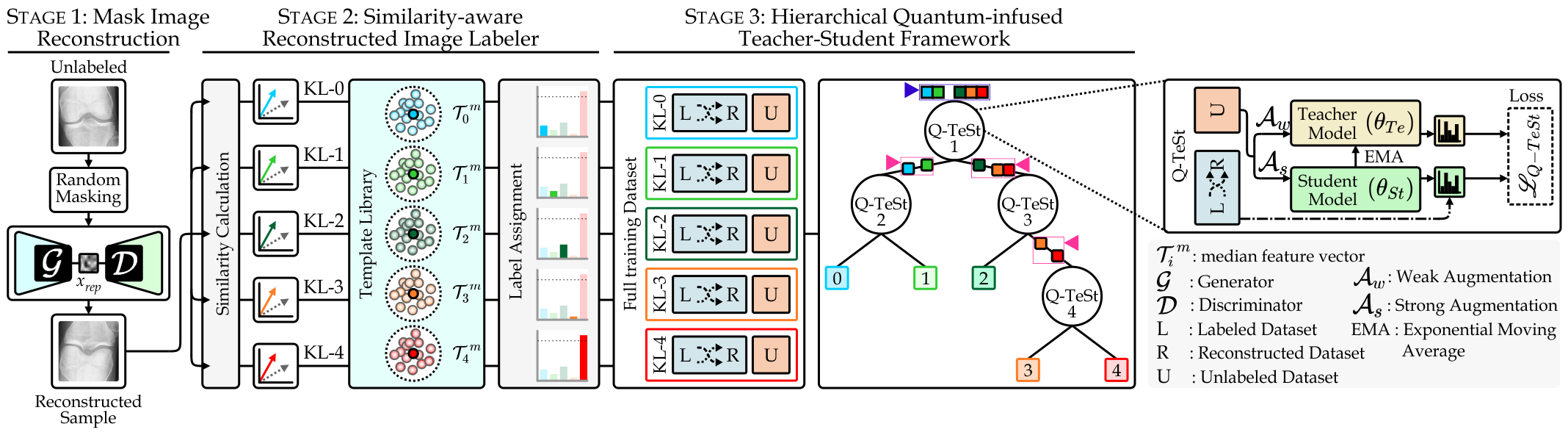}
    \caption{Overview of the proposed Hierarchical Semi-Supervised framework with Self-Supervision (H-SemiS) for knee osteoarthritis severity grading, consisting of three stages. \textsc{Stage}~1: Mask Image Reconstruction (MI-Rec)~(refer~\cref{sec:mirec}) for KXR reconstruction; \textsc{Stage}~2: Similarity-aware Reconstructed Image Labeler (SiRL) (refer~\cref{sec:sirl}), assigns labels (KL-0 to KL-4) to reconstructed samples; and \textsc{Stage}~3: Hierarchical Quantum-infused Teacher-Student Framework (HQ-TeSt)~(refer~\cref{sec:hqtest}), decomposes multi-class classification into binary class and refines predictions through teacher-student framework.}
    \label{fig:overview}
\end{figure*}

\subsection{Semi-Supervised learning-based Classification}
Semi-supervised learning (SSL) enhances model performance by utilizing both labeled and unlabeled samples, particularly when labeled data is scarce. Among SSL strategies, consistency regularization is widely used, encouraging consistent predictions under data perturbations~\citep{Ghosh2024cregssl}. The $\pi$-Model~\citep{Laine2017pi} applies weak-strong augmentation to generate pseudo-labels and predict from strongly augmented inputs. Its variant, TemporalEmbedding~\citep{Laine2017pi}, enhances pseudo-label stability by utilizing an exponential moving average (EMA). Mean Teacher~\citep{Tarvainen2017mt} further stabilizes pseudo-labels by updating the teacher model through EMA. In medical imaging, \cite{Berrimi2024mri} proposes a CNN-based SSL framework for osteoarthritis detection, while \cite{Farooq2023dcaae} introduces a dual-channel adversarial autoencoder for KL-grade classification. \cite{Huo2022dcmt} develops a teacher-student framework leveraging both labeled and unlabeled samples. \cite{Burton2020} uses a semi-supervised approach that combines CNNs with Monte Carlo patch sampling to improve knee MRI segmentation accuracy. Building on expanding consistency regularization, \cite{Nguyen2020semix} proposes the Semixup algorithm. However, current methods often underutilize unlabeled data. In contrast, H-SemiS fuses self-supervision with semi-supervision, leveraging proxy labels to maximize feature learning from unlabeled samples. Through consistency training on weak and strong augmentations, the framework improves performance and generalizability across diverse distributions.

\section{Methodology}
\label{sec:method}
\subsection{Problem Definition and Framework Overview}
\label{sec:prodef-overview}
We aim to improve the severity grading of knee osteoarthritis (KOA) in a semi-supervised setting by jointly leveraging self-supervised and semi-supervised learning. We utilize a small labeled set $\mathcal{D}_l=\{x_l,y_l\}_{i=1}^{n}$ and a large unlabeled set $\mathcal{D}_u=\{x_u\}_{j=1}^{m}$, with $m \gg n$, where $x_l,x_u \in \mathbb{R}^{h \times w \times ch}$ with $h$, $w$, and $ch$ denote the height, width, and number of channels for each KXR sample. The task is formulated as a multi-class classification problem where labels $y_i \in \{1,2,\ldots,\mathcal{C}\}$ correspond to KL grades. Our objective is to learn a classifier $f: x \rightarrow [0,1]^{\mathcal{C}}$ that minimizes the error between predicted and ground-truth severity grades by utilizing both labeled and unlabeled samples.

As illustrated in \autoref{fig:overview}, the framework operates in a three-stage pipeline to address data scarcity and class imbalance. \textsc{Stage}~1 (Masked Image Reconstruction) generates proxy data by utilizing unlabeled samples from $\mathcal{D}_u$ and partitioned into non-overlapping patches $x^{k}_u=\{x_u^{(1)},\ldots,x_u^{(p)}\}$, where each patch $x^{k} \in \mathbb{R}^{p^2 \times ch}$. The positional information is added and 75\% of patches are randomly masked, after which the model reconstructs the missing regions by minimizing the reconstruction loss $(\mathcal{L}_{rec})$, producing reconstructed samples $x_{rec}^u$. Next, \textsc{Stage}~2 (Similarity-aware Reconstructed Image Labeler) assigns proxy labels $y_{rec}$ to these samples by maximizing a combined cosine and Euclidean similarity metric ${\ddot{\mathbb{S}}}_i(x_{rec}^u,\mathcal{T}_i)$ against class-specific templates $\mathcal{T}_i$. Finally, in \textsc{Stage}~3 (Hierarchical Quantum-infused Teacher-Student Framework) the proxy-labeled reconstructed samples and weakly and strongly augmented views of unlabeled data, $x_w=\mathcal{A}_w(x_u)$ and $x_s=\mathcal{A}_s(x_u)$, are jointly used within a hierarchical quantum-infused teacher–student framework. The teacher parameters $\theta_{Te}$ are updated using an exponential moving average of the student parameters $\theta_{St}$, and the trained teacher model is finally used to predict KOA severity. 

\begin{figure*} [!ht]
    \centering
    \includegraphics[width=\linewidth]{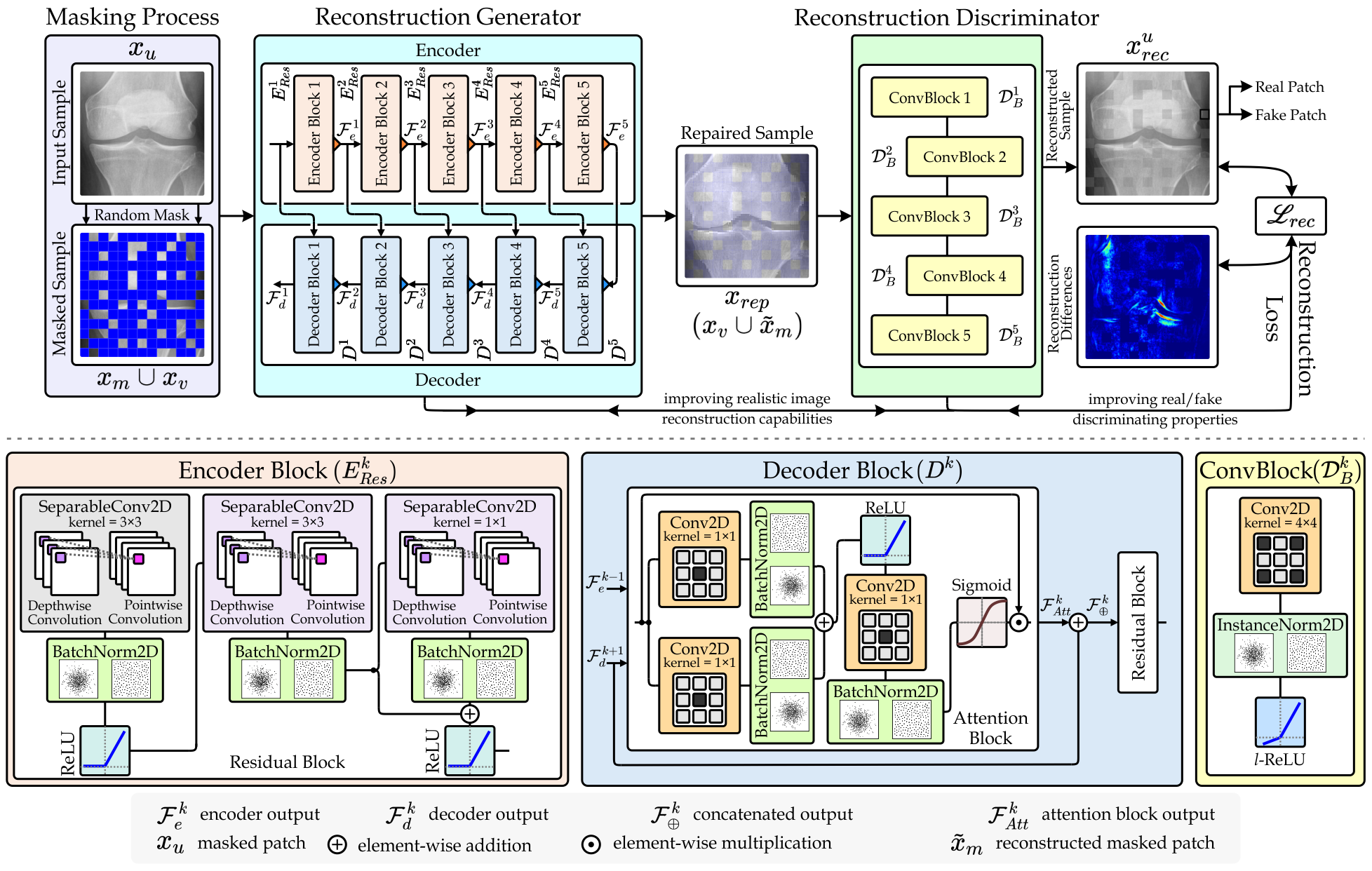}
    \caption{Overview of our Masked Image Reconstruction (MI-Rec). The MI-Rec process begins with applying random masking to the input sample, dividing the image into visible and masked patches. Next, a reconstruction generator reconstructs the masked patches. Finally, the reconstruction discriminator evaluates the generated masked patches, classifying them as real or fake to produce the final unlabeled reconstructed sample.}
    \label{fig:mirec}
\end{figure*}

\subsection{Stage 1: Masked Image Reconstruction}
\label{sec:mirec}
Inspired by recent advances in self-supervised masked image reconstruction~\citep{Song2024unitpathssl, Fei2023maegan, He2022mae}, we employ self-supervised learning within a semi-supervised framework for grading osteoarthritis severity. The proposed Masked Image Reconstruction (MI-Rec) module builds on Masked Autoencoders~\citep{He2022mae} and Generative Adversarial Networks~\citep{Goodfellow2020gan}. Through adversarial reconstruction of masked images, MI-Rec learns robust anatomical features from unlabeled data and generates high-fidelity proxy samples that (i) alleviate data scarcity, (ii) balance class distributions, and (iii) inject diverse discriminative features into the training set. As shown in \autoref{fig:mirec}, MI-Rec comprises a reconstruction generator $(\mathcal{G}_{MI\text{-}Rec})$ and a reconstruction discriminator $(\mathcal{D}_{MI\text{-}Rec})$, where the generator reconstructs masked regions, and the discriminator distinguishes real patches from reconstructed ones, ensuring anatomical correctness and structural consistency.

\subsubsection{Masking Process}
Given an input sample ${x}_u \in \mathbb{R}^{h \times w \times ch}$, where $h$, $w$, and $ch$ denote the height, width, and number of channels. We partition the input sample into $N_p$ non-overlapping patches of size $p \times p$, with $N_p = \frac{h \times w}{p^2}$ patches. Each patch $p_i \in \mathbb{R}^{p \times p \times ch}$ is flatten into a vector $z_i \in \mathbb{R}^{p^2 \cdot ch}$ forming the patch sequence $\mathbf{z} = \{z_1, \ldots , z_{N_p}\}$. We add positional encoding $PE_i$ to each patch $z_i$ to preserve the spatial information, where $\mathbf{PE} = \{PE_1, \ldots ,PE_{N_p} \}$. The resulting patch representations with positional encoding are defined in \cref{eq:pos}.
\begin{equation}
\label{eq:pos}
    \tilde{z}_i = z_i + PE_i ; \qquad
    \mathbf{\tilde{z}} = \{\tilde{z}_1, \ldots ,\tilde{z}_{N_p} \}
\end{equation}
After patching and adding positional encoding, we follow a uniform distribution to apply random masking to create a uniformly distributed random set $\mathbf{z_m}$. A masking ratio of 75\% to partition the sequence of patches into the masked subset $x_m$ and visible subset $x_v$, which are defined in \cref{eq:mask_a} and \cref{eq:vis_b}.
\begin{subequations}
    \begin{gather}
    \label{eq:mask_a}
       x_m = \{z_i\,|\,i \in \mathbf{z_m} \}, \quad \mathbf{z_m} = RandomSample \left(\{1, \ldots, N_p \}, N_m \right) \\
    \label{eq:vis_b}
        x_v = \{z_i\,|\,i \notin \mathbf{z_m} \}
    \end{gather}
\end{subequations}
Here, $N_m = 0.75 N_p$ represents the number of masked patches, and $x_m \cup x_v = \{1, \ldots , N_p \}$ and $x_m \cap x_v = \varnothing$.
\paragraph[]{Why high masking ratio? \eatpunct}
When masking is high, only a few patches remain visible, forcing the model to learn deeper patterns instead of relying on random predictions from neighboring pixels. Uniform random masking ensures the model learns to reconstruct missing patches from all parts of the image equally, preventing any bias towards specific regions like edges or the center. Since most patches are masked, the input to the reconstruction generator is primarily sparse. This sparsity encourages the reconstruction generator (described next) to process information from the visible patches efficiently.

\subsubsection{Reconstruction Generator}
Our reconstruction generator $\mathcal{G}_{MI\text{-}Rec}$ follows an encoder–decoder architecture inspired by Masked Autoencoders (MAE). Its primary objective is to reconstruct the masked patches $\tilde{x}m$ from the visible patches $x_v$. The encoder comprises five residual blocks ${E_{Res}^k}, \{k=1 \ldots 5\}$, where each block applies separable convolutions $(\mathbb{C}_{Sep})$, batch normalization $(\beta_n)$, and ReLU activation $(\alpha_{ReLU})$, followed by downsampling to reduce spatial resolution. The encoder progressively reduces the spatial resolution while increasing feature dimensionality, producing encoded features $\mathcal{F}_e^k$ as defined in \cref{eq:en_feat}. The decoder then uses the encoded representations to reconstruct the masked patches $\tilde{x}_m$, which are validated by the reconstruction discriminator.
\begin{equation}
    \label{eq:en_feat}
     \mathcal{F}_e^k = E_{Res}^k \left(\mathcal{F}_e^{k-1}\right), \quad k \in \{1, \ldots ,5\}
\end{equation}
Here, the initialization $\mathcal{F}_e^0 = x_v$ and $E_{Res}^k$ takes the encoded representation as input from the previous block's output. $\mathcal{F}_e^5$ is the final output of the last encoder block $E_5$.

After encoding, the highest-level encoded feature maps are passed to the decoder to reconstruct the masked patches. The decoder mirrors the encoder and consists of five decoder blocks $D^k$, $k \in {1,\ldots,5}$. Each block applies upsampling, an attention unit $(D_{Att}^k)$, and a residual block $(D_{Res}^k)$ to progressively recover spatial resolution. A transpose convolution upsamples the features from the previous block. Next, an attention unit fuses the upsampled features with the corresponding encoder output $\mathcal{F}_e^{k-1}$ to produce an attention-refined representation $\mathcal{F}_{Att}^k$. The attention output is concatenated with the upsampled features to form $\mathcal{F}_\oplus^k$, which is then refined by $D_{Res}^k$ to generate the decoded features $\mathcal{F}_d^k$, as defined in~\cref{eq:dec_feat}:
\begin{equation}
    \label{eq:dec_feat}
     \mathcal{F}_d^k = D^k \left(\mathcal{F}_d^{k+1}\right), \quad k \in \{1, \ldots ,5\}
\end{equation}
For the final decoding stage, $D^5$ receives the encoder output $\mathcal{F}_e^5$ (i.e., $\mathcal{F}_d^6=\mathcal{F}_e^5$), and the output $\mathcal{F}_d^1$ represents the reconstructed masked image $\tilde{x}_m$.

\subsubsection{Reconstruction Discriminator}
Next, the reconstruction discriminator $\mathcal{D}_{MI\text{-}Rec}$ predicts whether each reconstructed patch is real or synthetic. It takes a repaired image $x_{rep}$ as input, formed by replacing the masked patches $x_m$ in the visible image $x_v$ with reconstructed patches $\tilde{x}_m$, such that $x_{rep}=x_v \cup \tilde{x}_m$ and $x_m \cap x_v=\varnothing$. Given the ground-truth label sequence $\mathbf{\tilde{y}}=\{\tilde{y}_1,\ldots,\tilde{y}_{N_p}\}$, where $\tilde{y}_k \in \{0,1\}$ denotes real $(1)$ or synthetic $(0)$ patches, the discriminator learns to predict these labels from $x_{rep}$. The discriminator comprises five convolutional blocks $\mathcal{D}_B^k$, $k \in \{1,\ldots,5\}$, each applying convolution $\mathbb{C}$, instance normalization $(\beta_{in})$, and leaky ReLU activation $(\underset{lReLU}{\alpha})$, followed by progressive downsampling. The network outputs a scalar confidence score indicating the likelihood of each patch being real, as defined in~\cref{eq:discriminator}:
\begin{equation}
    \label{eq:discriminator}
     \mathcal{D}_{MI-Rec} = prob_{\mathcal{D}} \left(\tilde{y}_k|x_{rep},k \right), \quad k \in \{1, \ldots , N_p \}
\end{equation}
where $prob_{\mathcal{D}}$ denotes the discriminator’s confidence for classifying each patch as real or synthetic. When the confidence score is high, the repaired image is denoted as $x_{rec}^u$, representing the final unlabeled reconstructed sample.

\subsubsection{Loss Strategy}
The MI-Rec framework adopts a dual loss strategy to optimize the reconstruction generator $\mathcal{G}_{MI\text{-}Rec}$ and discriminator $\mathcal{D}_{MI\text{-}Rec}$. The generator loss $\mathscr{L}_{gen}$ combines adversarial loss (Binary Cross-Entropy (BCE)) and reconstruction loss (L1). The adversarial term $\mathscr{L}_{adv}$ encourages $\mathcal{G}_{MI\text{-}Rec}$ to produce patches classified as real by $\mathcal{D}_{MI\text{-}Rec}$, while the reconstruction term $\mathscr{L}_{rec}$ preserves consistency with masked regions. The generator objective is defined in~\cref{eq:gen_loss}:
\begin{equation}
    \label{eq:gen_loss}
    \mathscr{L}_{gen} = \underbrace{BCE \bigl(\mathcal{D}_{MI-Rec}(\tilde{x}_m),1 \bigr)}_{\mathscr{L}_{adv}} + \alpha \cdot \underbrace{\Vert \tilde{x}_m - x_m \Vert_1}_{\mathscr{L}_{rec}}    
\end{equation}
where, $\tilde{x}_m$ denotes the reconstructed patch, $x_m$ the original masked patch, $\mathcal{D}_{MI\text{-}Rec}$ the discriminator prediction for the reconstructed patch, and $\alpha$ a weighting factor that balances the two loss terms. The reconstruction discriminator loss $\mathscr{L}_{dis}$ measures its ability to classify real patches as real and reconstructed patches as fake using BCE loss. The discriminator objective is defined in~\cref{eq:dis_loss}.
\begin{equation}
    \label{eq:dis_loss}
    \mathscr{L}_{dis} = \underbrace{\frac{1}{2} \bigl(BCE (\mathcal{D}_{MI-Rec}(x_m),1) \bigr)}_{\mathscr{L}_{real}} + \alpha \cdot \underbrace{\frac{1}{2} \bigl(BCE (\mathcal{D}_{MI-Rec}(\tilde{x}_m),0) \bigr)}_{\mathscr{L}_{fake}}    
\end{equation}
where, $\mathcal{D}_{MI\text{-}Rec}(x_m)$ and $\mathcal{D}_{MI\text{-}Rec}(\tilde{x}_m)$ denote the discriminator predictions for real and generated patches. The real patch is assigned a label of $1$, while the fake patch is labeled $0$. The loss averages the two BCE terms to balance real and fake classifications. The final objective combines the generator and discriminator losses, as defined in~\cref{eq:mirec_loss}:
\begin{equation}
    \label{eq:mirec_loss}
    \mathscr{L}_{MI-Rec} = \mathscr{L}_{gen} + \mathscr{L}_{dis}
\end{equation}
where, $\mathscr{L}_{gen}$ promotes realistic and accurate patch reconstruction, while $\mathscr{L}{dis}$ enforces effective discrimination between real and reconstructed patches.

\subsection{Stage 2: Similarity-aware Reconstructed Image Labeler}
\label{sec:sirl}

\begin{figure*}[!ht]
    \centering
    \includegraphics[width=\linewidth]{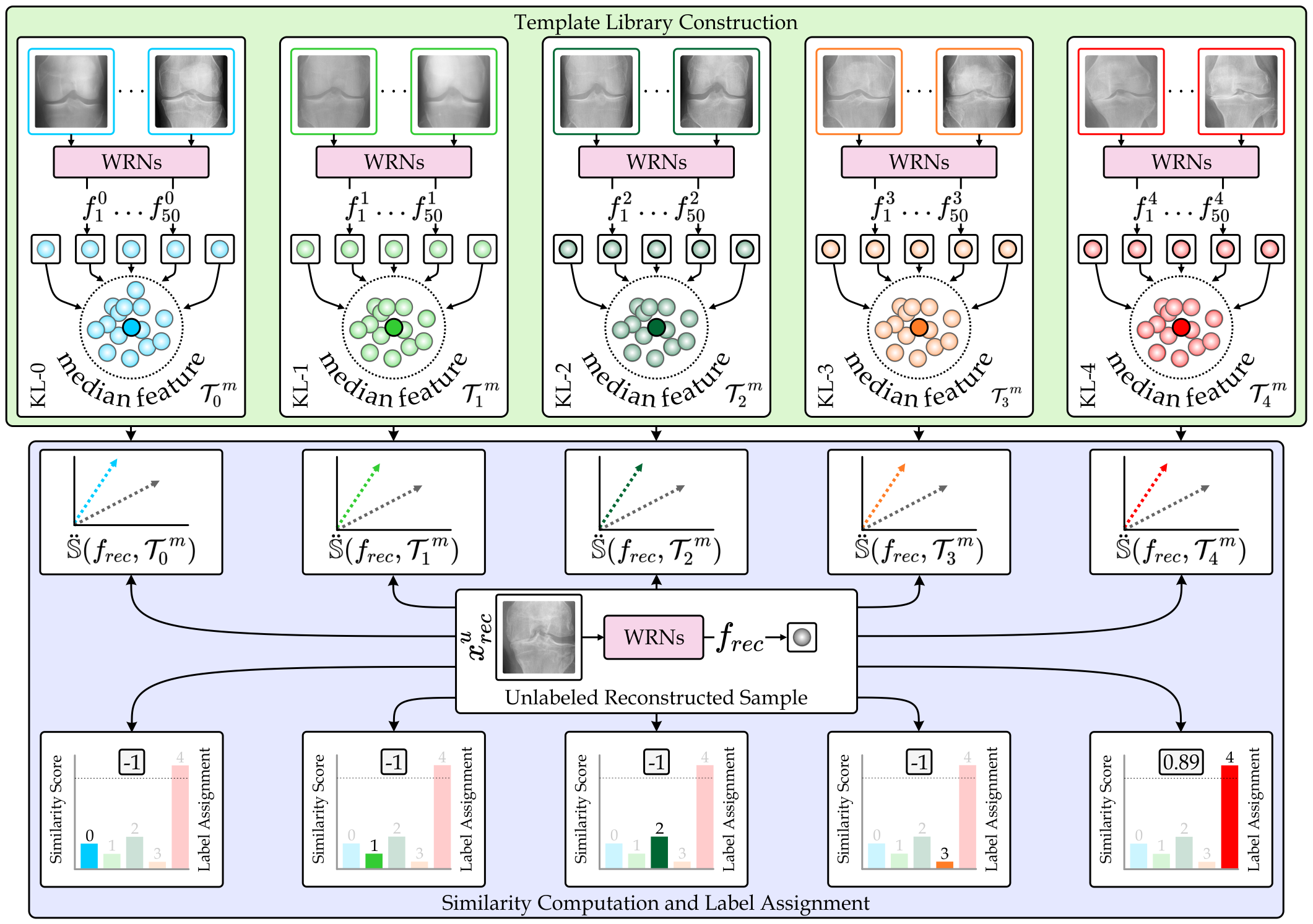}
    \caption{Overview of the Similarity-aware Reconstructed Image Labeler (SiRL) module for labeling unlabeled reconstructed KXR samples. The template library (top) is constructed using Wide Residual Networks (WRNs) to derive median feature vectors $\mathcal{T}_i^m$ for each KL grade. The similarity computation (bottom) assigns labels based on the highest similarity score.}
    \label{fig:sirl}
\end{figure*}

The Similarity-aware Reconstructed Image Labeler (SiRL) assigns proxy labels to unlabeled reconstructed samples by measuring their similarity to predefined templates. SiRL performs three steps: (a) template library construction by extracting and storing features, (b) similarity computation between a sample and the median feature vector of each template, and (c) similarity ranking and labeling by selecting the template with the highest similarity score. An overview is shown in~\cref{fig:sirl}.

\subsubsection{Template Library Construction}
The process begins by constructing a template library $\mathcal{T}=\{\mathcal{T}_0,\ldots,\mathcal{T}_4\}$ for the five osteoarthritis severity classes. For each class, we randomly select 50 labeled samples and extract super-pixel level features $f_s^i$ using a pre-trained Wide Residual Network~\citep{Zagoruyko2016wresnet}. These features capture the essential visual patterns of each class. For each template, we compute a representative feature vector $\mathcal{T}_i^m$ by aggregating the extracted features from the selected samples, which serves as the reference for subsequent similarity matching, where $i \in \{0,\ldots,4\}$.

\begin{figure*}[!b]
    \centering
    \includegraphics[width=\linewidth]{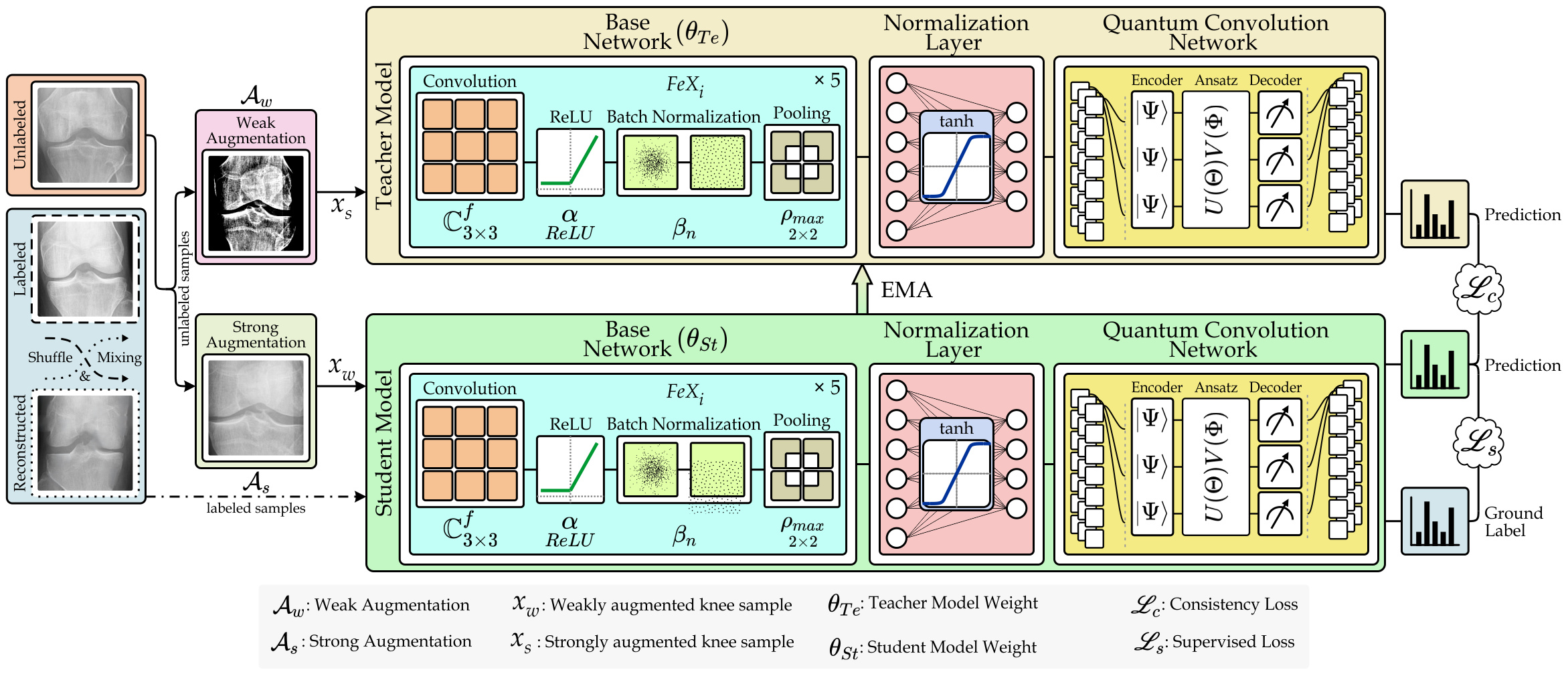}
    \caption{Overview of the Quantum-infused Teacher-Student (Q-TeSt) framework. Q-TeSt integrates convolutional networks with quantum convolutional networks in a teacher-student learning framework. The architecture leverages weakly and strongly augmented samples, transferring knowledge via EMA to optimize supervised and unsupervised learning objectives.}
    \label{fig:qtest}
\end{figure*}

\subsubsection{Similarity Computation and Label Assignment}
For each unlabeled reconstructed sample $x_{rec}^u$, SiRL compares its feature representation $f_{rec}$ with the median feature vectors in the template library. It computes similarity using cosine similarity and Euclidean distance, which are combined into a unified score $\ddot{\mathbb{S}}$ to provide a balanced similarity measure, as defined in~\cref{eq:simscore}:
\begin{equation}
    \label{eq:simscore}
    \ddot{\mathbb{S}} \left(f_{rec}, \mathcal{T}_i^m \right) = \alpha \cdot \underbrace{\frac{f_{rec} \cdot \mathcal{T}_i^m}
    {\Vert f_{rec} \Vert \cdot \Vert \mathcal{T}_i^m \Vert}}_{cosine-similarity} +
    (1 - \alpha) \cdot \underbrace{\frac{1}{\sqrt{(f_{rec} - \mathcal{T}_i^m)(f_{rec} - \mathcal{T}_i^m)^\text{T}}}}_{euclidean-distance}
\end{equation}
where $\alpha$ is a tunable parameter that balances the contributions of cosine similarity and Euclidean distance in the overall score. To ensure reliable template matching and avoid incorrect proxy label assignment, we apply the following labeling rules, formalized in~\cref{eq:proxylabelrule}:
\begin{equation}
    \label{eq:proxylabelrule}
    y_{rec} =  \left\{\begin{array}{rcl}
    {\mathcal{C}} & \mbox{if} & \arg \underset{i \in \mathcal{C}}{\max} \; \{\ddot{\mathbb{S}} \left(f_{rec}, \mathcal{T}_i^m \right) \geq \tau\}
    \\
    -1 & \mbox{if} & \arg \underset{i \in \mathcal{C}}{\max} \; \{\ddot{\mathbb{S}} \left(f_{rec}, \mathcal{T}_i^m \right) < \tau \}
    \end{array}\right.
\end{equation}
where $\mathcal{C}=\{KL_0,\ldots,KL_4\}$ denotes the osteoarthritis severity classes, $-1$ indicates an unassigned sample that is excluded from further processing, and $\tau$ is a threshold set to $0.80$ $(\tau_{80})$ to filter unreliable proxy labels. The operator $\arg \underset{i}{\max} \; {\ddot{\mathbb{S}}}_i (f_{rec}, \mathcal{T}_i)$ selects the template $\mathcal{T}_i$ with the highest similarity score and assigns the corresponding label $y_{rec}$ to the reconstructed sample. The labeled sample is then forwarded to the Hierarchical Quantum-infused Teacher–Student framework for final severity grading.

\subsection{Stage 3: Hierarchical Quantum-infused Teacher-Student Framework}
\label{sec:hqtest}
The proposed Hierarchical Quantum-infused Teacher–Student Framework (HQ-TeSt) is a semi-supervised teacher–student model that leverages self-supervised, reconstructed samples for grading the severity of osteoarthritis. HQ-TeSt enhances representation learning by integrating reconstructed samples into training, while the student model continuously updates the teacher via an exponential moving average. The framework comprises two stages:(i) semi-supervised training of the quantum-infused teacher–student (Q-TeSt) model and (ii) hierarchical multi classification of osteoarthritis severity. We first describe the Q-TeSt architecture, then present its consistency training strategy, and finally detail the hierarchical multi-classifier.

\subsubsection{Quantum-infused Teacher-Student Framework}
\label{sec:qtest}
The proposed Quantum-infused Teacher–Student (Q-TeSt) framework is inspired by the Mean-Teacher (MT) architecture \citep{Tarvainen2017mt}. Unlike MT, which utilizes a ResNet backbone, Q-TeSt employs a standard convolutional neural network infused with Quantum Convolutional Neural Networks, thereby forming a hybrid semi-supervised framework \citep{Mari2020qnntransf}. As shown in \cref{fig:qtest}, Q-TeSt extracts visual features from different augmented views of the same sample in both classical and Hilbert spaces, thereby capturing fine-grained spatial patterns and nonlinear relationships. The framework addresses data scarcity in osteoarthritis severity grading by learning from limited labeled data while leveraging unlabeled samples. Q-TeSt comprises three components: (i) a base network, (ii) a normalization layer, and (iii) a quantum convolution network (QCN).

\noindent\textit{(i) Base Network:}
The base network in Q-TeSt captures spatial information using a stack of five feature extraction blocks $FeX_i$, $i \in \{1,\ldots,5\}$. Each block applies a $3 \times 3$ Conv2D operation $\mathbb{C}_{3 \times 3}^f$ with ReLU activation, where the number of filters increases as $f \in \{32,64,128,256,512\}$. ReLU mitigates vanishing gradients with low computational cost, while batch normalization $\beta_n$ stabilizes training by normalizing mini-batch statistics. A $2 \times 2$ max-pooling layer $\rho_{max}^{2 \times 2}$ with stride 2 then downsamples the feature maps. After the five blocks, the output of $FeX_5$ is flattened and passed through two fully connected layers, $FC_{1024}$ and $FC_{512}$. The operations in the $i$-th block are defined in~\cref{eq:fex}:
\begin{equation}
\label{eq:fex}
    FeX_i(x) = \rho_{max}^{2 \times 2}\biggl( \beta_n\Bigl(\alpha_{ReLU} \bigl( \mathbb{C}^f_{3 \times 3}(x) \bigr) \Bigr) \biggr)
\end{equation}
where $x$ is the augmented input sample. Finally, global average pooling is applied to the $FC_{512}$ output to produce the base feature vector $\mathcal{F}_{base} \in \mathbb{R}^{512}$.

To address the limitations of current quantum systems, which support small qubit circuits (eight in our case), we first map classical features into a quantum-compatible space. We project the high-dimensional feature vector into a 256-dimensional representation $\mathcal{F}_{t} \in \mathbb{R}^{256}$, where $256=2^8$ corresponds to the number of qubits, as defined in~\cref{eq:transform}:
\begin{equation}    
\label{eq:transform}
    \mathcal{F}_{t} = \mathcal{W}^{(t)} \cdot \mathcal{F}_{base} + b^{(t)}
\end{equation}
where, $\mathcal{W}^{(t)} \in \mathbb{R}^{256 \times 512}$ and $b^{(t)}$ denote the projection weights and bias. The transformed feature $\mathcal{F}_t$ is then fed into the quantum convolutional network. This layer bridges the classical and quantum components of the Q-TeSt framework, enabling efficient quantum processing of osteoarthritis samples.

\vspace{5pt}
\noindent\textit{(ii) Normalization Layer:} 
The transformed feature vector $\mathcal{F}_t$ is fed into QCN, which requires normalized inputs within a bounded range for stable learning~\citep{Benedetti2019paramqml}. As $\mathcal{F}_t$ is not normalized, previous work uses $\tanh$ to restrict feature values to the $[-1,1]$ range~\citep{Mari2020qnntransf}. However, $\tanh$ saturates near $\pm1$, leading to vanishing gradients as $\nabla \tanh(x)=1-\tanh^2(x)$. Following \cite{Magallanes2022hycqcnn}, we adopt $L_2-\tanh$ normalization to improve gradient flow while preserving feature magnitude and nonlinearity. The normalized output $\mathcal{F}_{norm}$ is defined in~\cref{eq:l2tanh}:
\begin{equation}    
\label{eq:l2tanh}
    \mathcal{F}_{norm} = \tanh{\left(\omega \cdot \frac{\mathcal{F}_{t}}{\Vert\mathcal{F}_{t} \Vert_2}\right)}
\end{equation}
where $\mathcal{F}_{t}$ denotes the projected feature embedding and $\omega$ controls the saturation range, ensuring that the normalized features lie within a Bloch sphere of radius $\omega$.

\noindent\textit{(iii) Quantum Convolution Network:}
The quantum convolution network (QCN) operates directly in the quantum domain. First, the quantum encoder $\mathcal{U}_e(\mathcal{F}_{norm})$ maps the normalized feature vector $\mathcal{F}_{norm}$ to a quantum state. Next, the ansatz $\mathcal{U}_a(\Theta)$ transforms the encoded state using learnable parameters $\Theta$. Finally, the decoder applies quantum state measurement to project the qubit-encoded information back into the classical domain for further processing. During training, the decoder updates $\Theta$ by minimizing the cost function.

\begin{figure*}[!b]
    \centering
    \includegraphics[width=0.95\linewidth]{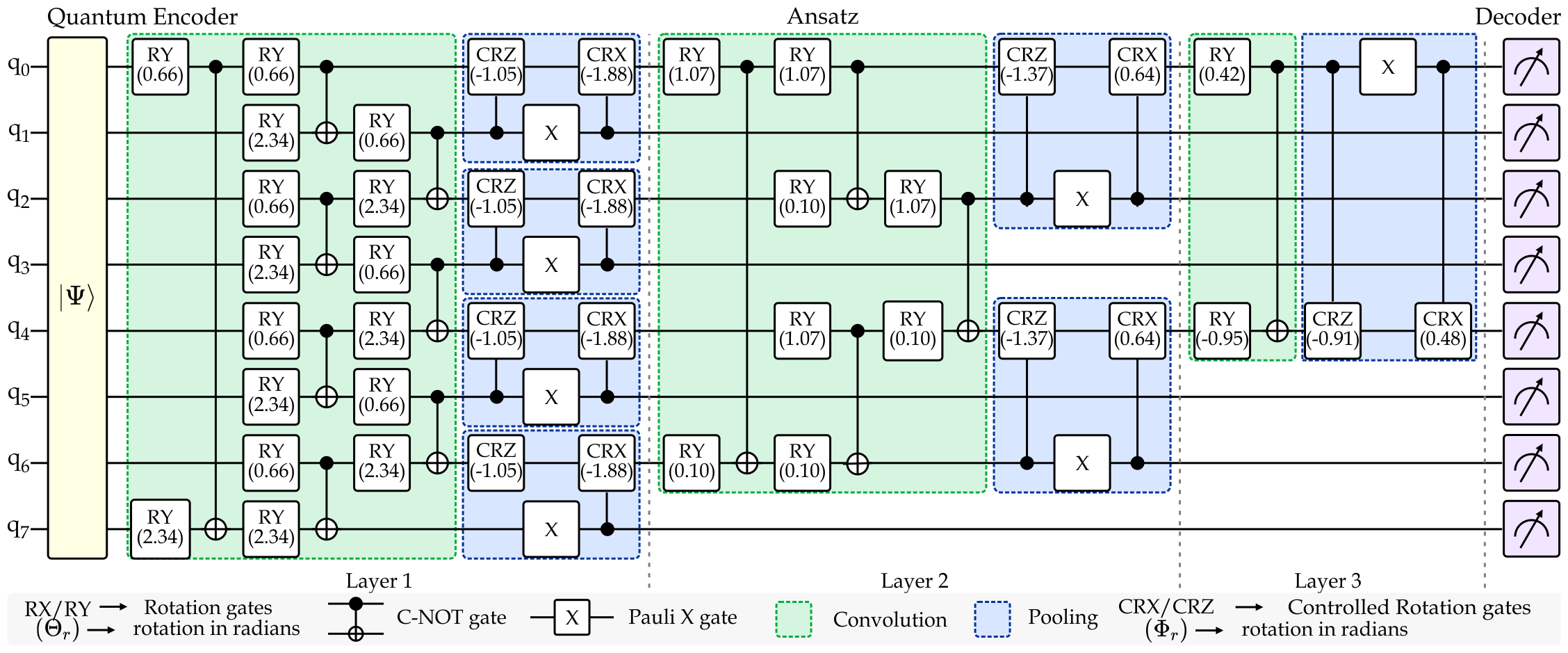}
    \caption{Architectural overview of the Quantum Convolutional Network (QCN), comprising a quantum encoder, ansatz, and decoder. The quantum encoder processes and encodes the input data for transformation in the ansatz. The ansatz applies quantum gates to capture complex transformations through entanglement, and the decoder extracts the final outputs.}
    \label{fig:qcn}
\end{figure*}

\vspace{2pt}
\noindent\textbf{Quantum Encoder:}
The quantum algorithm operates on quantum states, and normalized classical features cannot be used directly by the quantum convolution network. Therefore, the encoder maps the input features into a quantum state before ansatz processing. Among common encoding strategies, computation basis encoding converts inputs into binary states, whereas amplitude encoding maps data to quantum state amplitudes. We adopt amplitude encoding, which represents the normalized feature vector $\mathcal{F}_{norm}$ as the amplitudes of a quantum state, ensuring that the sum of squared amplitudes equals one. This encoding applies a unitary transformation, defined in~\cref{eq:amplien}:
\begin{equation}
\label{eq:amplien}
    \vert \Psi \rangle = \sum_{i=1}^N a_i \vert i \rangle        
\end{equation}
where, $\vert \Psi \rangle$ denotes the quantum state, $a_i$ are the components of $\mathcal{F}_{norm}$ satisfying $\sum_{i=1}^{N} |a_i|^2 = 1$, and $\vert i \rangle$ represents the computational basis states. Amplitude encoding is qubit-efficient, encoding $N$ features using $\lceil \log_2 N \rceil$ qubits, which limits circuit width and improves scalability.

\vspace{2pt}
\noindent\textbf{Ansatz:}
An ansatz is a parameterized quantum circuit composed of tunable quantum gates whose parameters are optimized to approximate a target quantum state. By adjusting these parameters, the ansatz explores a broad space of quantum states, capturing correlations and hidden patterns that enable efficient learning of complex relationships. In this work, the QCN circuit comprises three layers, as shown in~\autoref{fig:qcn}, where each layer applies unitary operations analogous to classical convolution and pooling.

The convolutional layers apply unitary transformations $U(\Theta)$ implemented as parameterized quantum gates acting on qubits. In our setting, each transformation consists of rotation gates $R_y(\Theta_0)$ and $R_y(\Theta_1)$ followed by a controlled-NOT (CNOT) gate, as defined in~\cref{eq:uniconv}:
\begin{equation}
\label{eq:uniconv}
    U(\Theta) = CNOT \cdot R_y (\Theta_0) \otimes R_y (\Theta_1)
\end{equation}
where the CNOT gate flips the target qubit when the control qubit is in the $\vert1\rangle$ state, and $R_y(\Theta_q)$ rotates qubit $q\in{0,1}$ around the Y-axis of the Bloch sphere by $\Theta_q$. The tensor product $\otimes$ combines the single-qubit rotations into a two-qubit operation. The pooling operation $V(\Phi)$ reduces quantum state dimensionality using Controlled Rotation-Z $(CR_z(\Phi_q))$, Pauli-X $(X)$, and Controlled Rotation-X $(CR_x(\Phi_q))$ gates to selectively reduce entanglement between qubit pairs. This downsampling limits circuit complexity while preserving salient information, playing a role analogous to pooling in classical CNNs while exploiting quantum entanglement and superposition for feature extraction.

\vspace{2pt}
\noindent\textbf{Decoder:}
The decoder maps the quantum state produced by the ansatz to classical features through quantum measurement. It performs computational basis measurements on the processed quantum state and estimates the associated outcome probabilities, which form the classical output vector. In our implementation, measurement is applied to wire 4, yielding the probability distribution over computational basis states. For a single qubit, the probability of observing state $\vert0\rangle$ or $\vert1\rangle$ is defined in~\cref{eq:measurement}:
\begin{equation}
\label{eq:measurement}
    p(i) = \vert\langle q_i \vert \ddot{\Psi} \rangle \vert^2
\end{equation}
where $\vert q_i \rangle$ denotes the computational basis state with $q_i \in \{0,1\}$, and $\ddot{\Psi}$ is the final quantum state after the convolution and pooling operations. The ansatz parameters are optimized using a hybrid quantum–classical approach. The quantum circuit produces measurement outcomes for current parameters and then used to evaluate an objective function. We employ cross-entropy loss $(\mathscr{L}_{CE})$ to iteratively update the parameters and minimize this objective. 

\subsubsection{Semi-Supervised Training}
\label{sec:semisupervised}
Semi-supervised training forms the first stage of the Hierarchical Quantum-infused Teacher–Student (HQ-TeSt) framework, leveraging labeled, reconstructed labeled, and unlabeled samples. The objective is to minimize supervised loss on labeled data while reducing consistency loss on unlabeled samples. During training, the teacher predicts on strongly augmented unlabeled samples, and the student enforces consistency on weakly augmented views of the same inputs. This process promotes invariant feature learning and enhances robustness to perturbations, thereby improving generalization. The semi-supervised training of Q-TeSt proceeds in three steps: (a) model initialization, (b) supervised training on labeled data, and (c) consistency training on unlabeled data.

\vspace{2pt}
\noindent\textit{(a) Model Initialization:}
The quantum-infused teacher–student (Q-TeSt) framework consists of a teacher and a student model with identical architectures, each comprising a convolutional network followed by quantum convolution (see~\autoref{sec:qtest}). The convolutional weights of the student and teacher are initialized independently and updated using EMA, while the quantum convolution weights are kept identical across both models. This design stabilizes the classical feature extraction stage, yielding more consistent inputs for the QCN. We denote the convolutional weights of the teacher and student by $\theta_{Te}$ and $\theta_{St}$, respectively, and the shared quantum weights by $\theta_{QCN}$.

\vspace{2pt}
\noindent\textit{(b) Supervised Training on Labeled data:}
In supervised training, we use labeled samples to train the student model. Given the scarcity of labeled data, the student alone may not learn sufficiently discriminative representations. To address this limitation, we incorporate self-supervised learning (see~\autoref{sec:mirec}) to generate reconstructed samples, which serves a dual purpose: (i) increasing the volume of labeled data strengthens the student model’s ability to capture local patterns within the samples, and (ii) training the student on enriched samples subsequently improves the teacher model’s performance when processing unlabeled data. The student model $(\mathcal{M}{St})$ is optimized using supervised loss $(\mathscr{L}{sup})$, implemented as cross-entropy to align predictions with ground-truth labels. The supervised objective is defined in~\cref{eq:suploss}.
\begin{equation}
\label{eq:suploss}
    \mathscr{L}_{sup} = \mathsf{CE} \left(\mathcal{M}_{St} (x_i; \theta_{St}, y_i) \right)
\end{equation}
where $x_i$ and $y_i$ denote the labeled input and its ground-truth, respectively, and $\mathsf{CE}$ represents the cross-entropy loss.

\vspace{2pt}
\noindent\textit{(c) Consistency Training on Unlabeled data:}
Consistency training uses a teacher–student framework to improve predictions on unlabeled data. The student model $\mathcal{M}_{St}$ is first trained on labeled samples in a supervised manner, and the teacher model $\mathcal{M}_{Te}$ is updated via EMA of the student’s weights. The approach encourages both models to produce consistent predictions under different augmentations of the same unlabeled sample. We apply weak augmentations $\mathcal{A}_w$, such as flipping and translation, to generate inputs for $\mathcal{M}_{Te}$, and strong augmentations $\mathcal{A}_s$ using RandAugment~\citep{Cubuk2020randaug}, including invert, shear, and scale operations, for $\mathcal{M}_{St}$ (see~\cref{sec:aug}). For an unlabeled sample $x_u$, the teacher predicts a pseudo-label from the weakly augmented input $x_w=\mathcal{A}_w(x_u)$, while the student predicts from the strongly augmented counterpart $x_s=\mathcal{A}_s(x_u)$. The corresponding predictions are given by $y_{Te}=\mathcal{M}_{Te}(x_w;\theta_{Te})$ and $y_{St}=\mathcal{M}_{St}(x_s;\theta_{St})$. The unsupervised loss $\mathscr{L}_{unsup}$ enforces prediction consistency between the teacher and student models under different augmentations of the same input, as defined in~\cref{eq:unsuploss}. Here, $\mathcal{M}(\cdot)$ denotes the Q-TeSt classification framework, while $\theta_{Te}$ and $\theta_{St}$ represent the teacher and student model parameters, respectively.
\begin{equation}
\label{eq:unsuploss}
    \mathscr{L}_{unsup} = \mathcal{M} (\cdot) (\theta_{Te}, \mathcal{A}_w) (\theta_{St}, \mathcal{A}_s)
\end{equation}

An individual consistency mechanism \citep{Tarvainen2017mt} is employed in our framework to ensure consistency between the teacher and student model outputs. As defined in \cref{eq:consistency}, the loss penalizes per-sample prediction discrepancies under weak and strong augmentations:
\begin{equation} 
\label{eq:consistency}
    \mathscr{L}_{con} = \sum_{i=1}^{n+m} \mathbb{E}_{\mathcal{A}_w, \mathcal{A}_s}
    \Vert\mathcal{M}_{Te} (x_i,\theta_{Te},\mathcal{A}_w) - \mathcal{M}_{St} (x_i,\theta_{St},\mathcal{A}_s) \Vert_2^2
\end{equation}
where $\mathcal{A}_w$ and $\mathcal{A}_s$ denote weak and strong augmentations; $n$ and $m,(m\gg n)$ are labeled and unlabeled samples; $x_i$ is a training sample; and $\theta_{Te},\theta_{St}$ are model parameters. Following Mean Teacher framework~\citep{Tarvainen2017mt}, we update the teacher via EMA applied only to the student’s base network, while keeping the QCN fixed to stabilize classical feature extraction and provide consistent inputs to the quantum module, as defined in~\cref{eq:ema}.
\begin{equation}
\label{eq:ema}
    \theta_{Te}^{(t+1)} \leftarrow \mu \cdot \theta_{Te}^{(t)} + (1-\mu) \cdot \theta_{St}^{(t)}
\end{equation}
where $t$ represents the training iteration and $\mu$ is the EMA smoothing coefficient, set to $0.99$ following~\cite{Tarvainen2017mt}, which controls the update rate of the teacher parameters. This consistency strategy gradually transfers knowledge from the student to the teacher, aligns their predictions, and improves performance on both labeled and unlabeled data.

\subsubsection{Hierarchical Multi-classifier}
\label{sec:him}
The Hierarchical Multi-classifier (HiM) classifies knee X-rays into five osteoarthritis severity grades using hierarchical binary classification. Motivated by \cite{Ning2024combat}, HiM mitigates class imbalance by decomposing the multi-class task into a sequence of binary decisions. The key distinction lies at each binary decision node, where reconstructed samples are incorporated and the classifier is trained in a semi-supervised manner using the Q-TeSt framework (refer to~\cref{sec:qtest}).

\vspace{2pt}
\noindent\textit{(i) Hierarchical Decomposition:}
In the hierarchical decomposition stage, we convert the multiclass osteoarthritis grading problem $\mathcal{C}={KL_0,\ldots,KL_4}$ into a sequence of binary classification tasks using a dual rule-based strategy. At the root node, classes are grouped by severity into normal $\mathcal{C}_n=\{KL_0,KL_1\}$ and osteoarthritis $\mathcal{C}_o=\{KL_2,KL_3,KL_4\}$. Subsequent levels leverage majority and minority class distributions to form balanced binary splits. This process builds a hierarchical binary tree, where each decision node is implemented with the Q-TeSt framework $\text{Q-TeSt}_i$ and leaf nodes correspond to specific severity grades. Hierarchical decomposition improves classification accuracy and mitigates multi-class imbalance.

\begin{figure}[!ht]
    \centering
    \includegraphics[width=\linewidth]{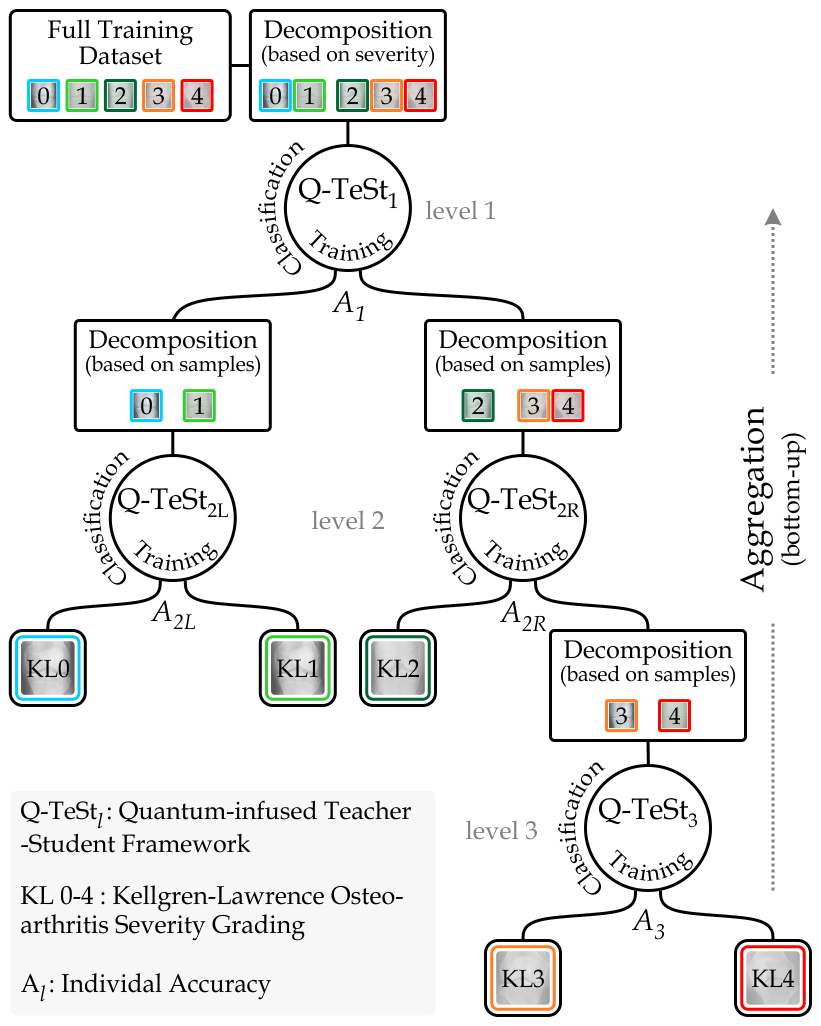}
    \caption{Overview of the Hierarchical Multi-classifier (HiM) module for knee osteoarthritis severity grading. The HiM module operates in three stages. First, during the decomposition stage, the entire training dataset is divided based on the severity at the root node and sample distributions at subsequent nodes. Second, in the classification stage, Q-TeSt models are trained at each level to grade osteoarthritis severity. Finally, in the aggregation stage, the framework combines classification outputs in a bottom-up manner using a weighted scheme.}
    \label{fig:him}
\end{figure}

\vspace{2pt}
\noindent\textit{(ii) Classification:}
At each decision node $(\text{Q-TeSt}_i)$, we train a dedicated Q-TeSt framework in a semi-supervised manner and incorporate reconstructed samples to mitigate class imbalance. As shown in~\cref{fig:him}, the hierarchy performs binary decisions at successive levels. The root node $\text{Q-TeSt}_1$ separates combined normal $(\text{KL}_0,\text{KL}_1)$ from osteoarthritis $(\text{KL}_2,\text{KL}_3,\text{KL}4)$. The left branch $\text{Q-TeSt}{2L}$ further distinguishes $\text{KL}_0$ from $\text{KL}_1$, while the right branch $\text{Q-TeSt}_{2R}$ splits $\text{KL}_2$ from $(\text{KL}_3,\text{KL}_4)$. At the final level, $\text{Q-TeSt}_3$ discriminates between $\text{KL}_3$ and $\text{KL}_4$. After training all nodes, we aggregate their outputs to compute the final classification accuracy, providing a structured solution to the multi-class grading task.

\vspace{2pt}
\noindent\textit{(iii) Aggregation:}
In the aggregation stage, we compute the final classification performance by combining the outputs of all Q-TeSt frameworks using a weighted scheme. Each framework $\text{Q-TeSt}_i$ is assigned a weight $w_i$ proportional to its depth $l_i$ in the hierarchy, defined as $w_i=\ln(l_i+1)$. The weighting scheme assigns greater importance to classifiers at deeper levels of the hierarchy, which capture finer-grained distinctions, than to higher-level classifiers that model coarse separations. The aggregated accuracy $A_{\text{agg}}$ is computed as the normalized weighted sum of individual accuracies $A_i$, as shown in~\cref{eq:wtacc}.
\begin{equation}
\label{eq:wtacc}
    A_{agg} = \frac{\sum_{i} w_i \cdot A_i}{\sum_{i} w_i}
\end{equation}
where $i \in \{\text{Q-TeSt}_{1},\text{Q-TeSt}_{2L},\text{Q-TeSt}_{2R},\text{Q-TeSt}_{3}\}$. This normalization keeps $A_{\text{agg}}$ within $[0,1]$ and preserves the hierarchical structure, thereby providing a balanced overall performance measure.

\section{Dataset and Experimental Settings}
\label{sec:dexp}
\subsection{Benchmark Datasets}
\label{sec:dataset}
\begin{figure*}[!ht]
    \centering
    \includegraphics[width=\linewidth]{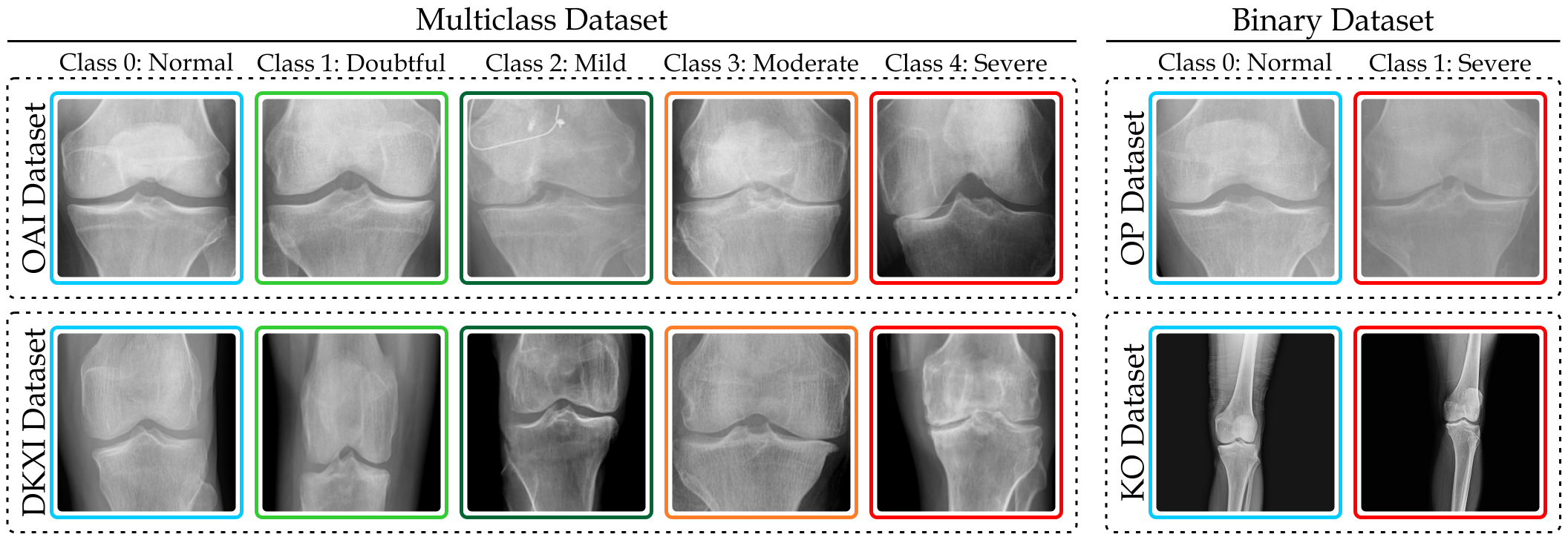}
    \caption{Sample dataset used for evaluation of the proposed framework. The Multiclass dataset (left) includes five severity levels (KL-0 to KL-4) from the OAI and DKXI datasets. The Binary dataset (right) comprises normal (Class 0) and severe (Class 1) cases from the OP and KO datasets. Each severity level is color-coded for clarity.}
    \label{fig:dataset}
\end{figure*}

We evaluate H-SemiS on two multiclass benchmark datasets and assess generalizability on two publicly available binary datasets. \cref{tab:dataset} summarizes the original and reconstructed samples used for training, and \cref{fig:dataset} illustrates representative knee osteoarthritis examples from each dataset.

\begin{table*}[!ht] \footnotesize
\centering
\begin{tabular*}{\linewidth}{@{\extracolsep{\fill}}l cccc cccc}
\toprule
\midrule
\textbf{Class} & & \textbf{Original} & \textbf{Reconstructed} & \textbf{Total} & &\textbf{Original} & \textbf{Reconstructed} & \textbf{Total} \\ 
& & \textbf{Samples} & \textbf{Samples} & \textbf{Samples} & &\textbf{Samples} & \textbf{Samples} & \textbf{Samples} \\ 
\midrule
\textbf{Multi-class Dataset} & & \multicolumn{3}{c}{\textbf{OAI Dataset }} & & \multicolumn{3}{c}{\textbf{DKXI Dataset}} \\ 
\cmidrule{1-1} \cmidrule{3-5} \cmidrule{7-9}
Class 0 && 3857 & 0    & 3857 &  & 576 & 0   & 576 \\ 
Class 1 && 1770 & 1982 & 3752 &  & 514 & 0   & 514 \\
Class 2 && 2578 & 1243 & 3821 &  & 273 & 243 & 516 \\
Class 3 && 1286 & 608  & 1894 &  & 267 & 0   & 267 \\
Class 4 && 295  & 1567 & 1862 &  & 206 & 54  & 260 \\
\noalign{\vskip 0.5ex}
\cdashline{1-9}[0.5pt/2pt]\noalign{\vskip 0.5ex}
\textbf{Binary Dataset} & & \multicolumn{3}{c}{\textbf{OP Dataset}} & & \multicolumn{3}{c}{\textbf{KO Dataset}} \\ 
\cmidrule{1-1} \cmidrule{3-5} \cmidrule{7-9}
Class 0 && 1651 & 478   & 2129 &  & 1434 & 2089 & 3523 \\
Class 1 && 2155 & 0 & 2155 && 3554 & 0 & 3554 \\  
\midrule
\bottomrule
\end{tabular*}
\caption{Overview of the dataset, highlighting the count of original and reconstructed samples in multi-class and binary-class datasets for KOA detection.}
\label{tab:dataset}
\end{table*}

\subsubsection{Multiclass Dataset}
\label{sec:multidata}
\vspace{2pt}
\noindent\textit{(i) Osteoarthritis Initiative Dataset:}
The Osteoarthritis Initiative (OAI)~\citep{ChenOAI} is a publicly available, multi-center dataset designed to study the development and progression of knee osteoarthritis. It contains 9,786 single-knee X-ray images from 4,796 subjects aged 45 to 79, each with a resolution of $299 \times 299$ pixels. To address multiclass imbalance in a hierarchical binary setting, we augment the dataset with 5,400 reconstructed KXR samples. Images are annotated using SiRL (see \cref{sec:sirl}), which follows the five-grade KL system: KL-0 (normal), KL-1 (doubtful), KL-2 (mild), KL-3 (moderate), and KL-4 (severe).

\vspace{2pt}
\noindent\textit{(ii) Digital Knee X-ray Images:}
The Digital Knee X-ray Images (DKXI) dataset~\citep{GornaleDKXI} comprises 8-bit grayscale images acquired using a PROTEC PRS 500E X-ray system from multiple hospitals and diagnostic centers. Two medical experts annotated all samples using the KL grading system. The dataset includes both single-knee images and scans containing both knees; we crop the latter into individual knees for consistency, resulting in 1,836 knee X-ray samples. All images are provided in PNG format, with resolutions ranging from $300 \times 162$ to $640 \times 481$ pixels. To mitigate class imbalance, we further augment the dataset with 297 reconstructed KXR samples.

\subsubsection{Binary Dataset}
\label{sec:binarydata}
\noindent\textit{(i) Osteoarthritis Prediction Dataset:}
The Osteoarthritis Prediction (OP) dataset~\citep{TaoOP}, released by the University of Florida and the OAI organization, contains anteroposterior, lateral, and oblique knee X-rays from diverse patients. In this study, we use only anteroposterior and lateral views. After cropping images that contain both knees into single-knee samples, the dataset includes 1,651 normal (class 0) and 2,155 severe (class 1) X-rays. To mitigate class imbalance, we augment the dataset with 478 reconstructed normal KXR samples.

\vspace{2pt}
\noindent\textit{(ii) Knee Osteoarthritis Dataset:}
The Knee Osteoarthritis (KO) dataset~\citep{ShawKO} contains X-rays with and without KL grading across multiple joints, including knees, wrists, elbows, necks, and hips. We use knee X-rays without KL grading to evaluate the generalizability of the proposed framework in a binary classification setting. Samples are grouped into normal (class 0) and severe (class 1). For consistency, we crop images containing both knees into single-knee views, yielding 4,988 knee X-ray samples. To address class imbalance, we augment the dataset with 2,089 reconstructed normal KXR samples.

\subsection{Evaluation Metrics}
We evaluate H-SemiS using standard metrics: Accuracy (Acc), Precision (Pre), Recall (Rec), and F1-score (F1) for grading KOA severity. For multiclass evaluation, Precision, Recall, and F1-score are macro-averaged across classes. \cref{tab:metrics} summarizes the mathematical definitions of these metrics. Here, $\text{TP}_{\text{KOA}}$ denotes KOA samples correctly classified as KOA, $\text{TN}_{\text{Nor}}$ denotes normal samples correctly classified as normal, $\text{FP}_{\text{KOA}}$ denotes normal samples misclassified as KOA, and $\text{FN}_{\text{Nor}}$ denotes KOA samples misclassified as normal.

\begin{table}[!ht] \footnotesize
\centering
    \caption{Evaluation metrics for H-SemiS framework for osteoarthritis severity grading}
    \begin{tabular*}{\linewidth}{@{\extracolsep{\fill}} l c}
        \toprule
        \midrule
        \textbf{Evaluation Metrics} & \textbf{Mathematical Formulation} \\
        \midrule
        Accuracy (Acc) & $ \mathlarger{\frac{\text{TP}_\text{KOA}+T\!N_{nor}}{\text{TP}_\text{KOA}+\text{FP}_\text{KOA}+T\!N_{nor}+\text{FN}_\text{Nor}}}$  \\
        & \\
        Precision (Pre) & $ \mathlarger{\frac{\text{TP}_\text{KOA}}{\text{TP}_\text{KOA} +\text{FP}_\text{KOA}}}$  \\
        & \\
        Recall (Rec) & $ \mathlarger{\frac{\text{TP}_\text{KOA}}{\text{TP}_\text{KOA}+\text{FN}_\text{Nor}}}$  \\
        & \\
        F1-Score (F1) & $ \mathlarger{2 \times \frac{Precision \times Recall}{Precision + Recall}}$ \\
        \midrule
        \bottomrule
    \end{tabular*}
\label{tab:metrics} 
\end{table}

\subsection{Implementation Details}
\label{sec:implementation}
\noindent\textbf{Data Preprocessing:}
The dataset comprises single KXR scans containing both knees with varying resolutions. For consistency, we detect individual knee joints using YOLOv2~\citep{Redmon2017yolo} and crop them into single-knee KXR samples, which are then resized to $224 \times 224$. To enhance low-level features and overall image quality, we apply Contrast Limited Adaptive Histogram Equalization (CLAHE)~\citep{Raghaw2024xccnet}. We further utilize BoneFinder~\citep{Lindner2013bonefinder} to detect key knee points and preserve joint space narrowing (JSN), which is crucial for capturing complex knee anatomy.

\vspace{2pt}
\noindent\textbf{Augmentations:}
\label{sec:aug}
We adopt two augmentation strategies for consistency training in the teacher–student framework (see \cref{sec:semisupervised}): weak $(\mathcal{A}_w)$ and strong $(\mathcal{A}_s)$. For weak augmentation $(\mathcal{A}_w)$, applied to the teacher, we flip KXR samples vertically and/or horizontally with probability 0.5 and randomly translate them by up to 0.2 of the image dimensions along each axis. For strong augmentation $(\mathcal{A}_s)$, applied to the student, we use RandAugment~\citep{Cubuk2020randaug}, which applies a sequence of transformations including pixel inversion, X/Y shearing in [-0.3, 0.3], scaling in [0.51, 0.60], horizontal and vertical translation in [-0.3, 0.3], brightness adjustment in [0.05, 0.95], rotation in [-30$^\circ$, 30$^\circ$], HSV ratio modification in [-1, 1], and Gaussian blurring with kernel sizes in [3,5].

\vspace{2pt}
\noindent\textbf{Training Setting:}
We train and evaluate H-SemiS and all baselines using TensorFlow and PyTorch on a Linux server equipped with an Intel Xeon Silver S-4316 CPU (2.30 GHz), 512 GB DDR4 RAM, and an NVIDIA A16 GPU (64 GB). We optimize networks with Adam (weight decay 3e-4, initial learning rate 3e-4) using a batch size of 8 (4 labeled, 4 unlabeled) for up to 100 epochs with early stopping (patience = 10). Datasets are split into 80\% training and 20\% testing, with the training set containing 20\% labeled (original + reconstructed) and 80\% unlabeled samples. All metrics are reported on the held-out test set. We further evaluate performance under varying labeled ratios $(1\%, 5\%, 10\%, 20\%)$ (see \cref{sec:labelratios}). The Quantum Convolution Network (QCN) is simulated using PennyLane\footnote{\url{https://pennylane.ai}} with Adam (learning rate 1e-3), batch size 64, and cross-entropy loss with logits for binary classification.

\section{Experimental Results and Analysis}
\label{sec:results}
\subsection{Comparison with Competing Baselines}
We evaluate H-SemiS against thirteen baselines on two multi-class datasets, OAI~\citep{ChenOAI} and DKXI~\citep{GornaleDKXI}. The comparison includes four supervised: \cite{Nguyen2024mat, Lo2023dlseptic, Teoh2023dhl, Hu2022enn}, four self-supervised: \cite{Azizi2023culp, Cai2023, Wu2023, Azizi2021}, and five semi-supervised: \cite{Berrimi2024mri, Farooq2023dcaae, Huo2022dcmt, Nguyen2020semix, Burton2020} methods. We reproduce all baseline results under identical experimental settings to ensure a fair comparison.

\subsubsection{Quantitative Analysis}
\label{sec:quantitative}
We report quantitative results across three training settings: supervised, self-supervised, and semi-supervised. In the supervised setting, models are trained using the full labeled dataset. In the self-supervised setting, models learn representations from unlabeled data without manual annotations. In the semi-supervised setting, all baselines are trained with 20\% labeled samples and no additional data, matching the H-SemiS protocol.

\begin{table*}[!ht]
\footnotesize
\centering
\caption{Quantitative evaluation of H-SemiS against existing baselines on the OAI and DKXI datasets. \textcolor{horg}{\textbf{Orange}} and \textcolor{sblue}{\textbf{blue}} denote the best and second-best results. $|\Downarrow_d|$ indicates the absolute performance drop relative to H-SemiS.}
\begin{threeparttable}
\begin{NiceTabular*}{\linewidth}{@{\extracolsep{\fill}}l@{}l c@{}c@{}c@{}c@{}c@{}c@{}c@{}c@{} c@{} c@{}c@{}c@{}c@{}c@{}c@{}c@{}c}
\toprule
\midrule

 & \textbf{Methods} & \multicolumn{8}{c}{\textbf{OAI Dataset}} & & \multicolumn{8}{c}{\textbf{DKXI Dataset}} \\ 
\cmidrule{3-10} \cmidrule{12-19}

 &  & \textbf{Acc} & $|\Downarrow_d|$ & \textbf{Pre} & $|\Downarrow_d|$ & \textbf{Rec} & $|\Downarrow_d|$ & \textbf{F1} & $|\Downarrow_d|$ 
 &  & \textbf{Acc} & $|\Downarrow_d|$ & \textbf{Pre} & $|\Downarrow_d|$ & \textbf{Rec} & $|\Downarrow_d|$ & \textbf{F1} & $|\Downarrow_d|$ \\  

\midrule
 I
 & \cite{Nguyen2024mat}   
 & 69.1 & 15.8 & 68.8 & 17.3 & 71.4 & 12.0 & 70.1 & 14.6 
 &  & 74.6 & 12.2 & 75.8 & 9.5 & 73.1 & 10.5 & 74.4 & 10.0 \\

 & \cite{Lo2023dlseptic}  
 & 80.9 & 4.0 & 81.6 & 4.5 & \textcolor{horg}{\textbf{83.5}} & 0.1 & 82.5 & 2.2  
 &  & 77.8 & 9.0 & 78.3 & 7.0 & 77.2 & 6.4 & 77.7 & 6.7 \\

 & \cite{Teoh2023dhl}     
 & \textcolor{sblue}{\textbf{83.1}} & 1.8 & 82.4 & 3.7 & 79.3 & 4.1 & 80.8 & 3.9  
 &  & 81.4 & 5.4 & \textcolor{sblue}{\textbf{84.4}} & 0.9 & 79.5 & 4.1 & 81.9 & 2.5 \\

 & \cite{Hu2022enn}       
 & 73.9 & 11.0 & 74.5 & 11.6 & 72.8 & 10.6 & 73.6 & 11.1  
 &  & 66.2 & 20.6 & 67.4 & 17.9 & 63.6 & 20.0 & 65.4 & 19.0 \\

\cdashline{1-19}[0.5pt/2pt]\noalign{\vskip 0.5ex}

 II
 & \cite{Wu2023}          
 & 59.4 & 25.5 & 58.1 & 28.0 & 57.5 & 25.9 & 57.8 & 26.9  
 &  & 69.8 & 17.0 & 70.6 & 14.7 & 67.3 & 16.3 & 68.9 & 15.5 \\

 & \cite{Azizi2023culp}   
 & 81.2 & 3.7 & 82.3 & 3.8 & 78.6 & 4.8 & 80.4 & 4.3  
 &  & 82.1 & 4.7 & 83.2 & 2.1 & 81.3 & 2.3 & \textcolor{sblue}{\textbf{82.2}} & 2.2 \\

 & \cite{Cai2023}         
 & 78.6 & 6.3 & 78.8 & 7.3 & 76.9 & 6.5 & 77.8 & 6.9  
 &  & \textcolor{sblue}{\textbf{85.1}} & 1.7 & 82.2 & 3.1 & 81.6 & 2.0 & 81.9 & 2.5 \\    

 & \cite{Azizi2021}       
 & 79.8 & 5.1 & 80.1 & 6.0 & 81.2 & 2.2 & 80.6 & 4.1  
 &  & 78.2 & 8.6 & 76.3 & 9.0 & 79.1 & 4.5 & 77.7 & 6.7 \\

\cdashline{1-19}[0.5pt/2pt]\noalign{\vskip 0.5ex}

 III
 & \cite{Berrimi2024mri}  
 & 77.1 & 7.8 & 78.4 & 7.7 & 80.1 & 3.3 & 79.2 & 5.5  
 &  & 82.8 & 4.0 & 81.9 & 3.4 & 82.4 & 1.2 & 82.1 & 2.3 \\

 & \cite{Farooq2023dcaae} 
 & 76.7 & 8.2 & 75.1 & 11.0 & 79.4 & 4.0 & 77.2 & 7.5  
 &  & 75.5 & 11.3 & 76.0 & 9.3 & 74.2 & 9.4 & 75.1 & 9.3 \\

 & \cite{Huo2022dcmt}     
 & 81.8 & 3.1 & 82.4 & 3.7 & 83.1 & 0.3 & \textcolor{sblue}{\textbf{82.7}} & 2.0  
 &  & 69.8 & 17.0 & 70.6 & 14.7 & 71.3 & 12.3 & 70.9 & 13.5 \\

 & \cite{Burton2020}      
 & 82.9 & 2.0 & \textcolor{sblue}{\textbf{82.9}} & 3.2 & 80.5 & 2.9 & 81.7 & 3.0  
 &  & 84.5 & 2.3 & 81.3 & 4.0 & \textcolor{sblue}{\textbf{82.8}} & 0.8 & 82.0 & 2.4 \\ 

 & \cite{Nguyen2020semix} 
 & 71.3 & 13.6 & 70.3 & 15.8 & 72.1 & 11.3 & 71.2 & 13.5  
 &  & 76.2 & 10.6 & 74.5 & 10.8 & 76.8 & 6.8 & 75.6 & 8.8 \\    

\cdashline{1-19}[0.5pt/2pt]\noalign{\vskip 0.5ex}

\rowcolor{lightgray}
  \textbf{Ours}
 & \textbf{H-SemiS}         
 & \textcolor{horg}{\textbf{84.9}} & 0.0 & \textcolor{horg}{\textbf{86.1}} & 0.0 & \textcolor{sblue}{\textbf{83.4}} & 0.0 & \textcolor{horg}{\textbf{84.7}} & 0.0  
 &  & \textcolor{horg}{\textbf{86.8}} & 0.0 & \textcolor{horg}{\textbf{85.3}} & 0.0 & \textcolor{horg}{\textbf{83.6}} & 0.0 & \textcolor{horg}{\textbf{84.4}} & 0.0 \\

\midrule
\bottomrule
\end{NiceTabular*}
\begin{tablenotes}
\scriptsize
\centering
\item I: Supervised-based Techniques; II: Self-Supervised-based Techniques; III: Semi-Supervised-based Techniques
\end{tablenotes}
\end{threeparttable}
\label{tab:quantitative}
\end{table*}

\textbf{(i) Analysis on OAI Dataset:}
We report quantitative results in \cref{tab:quantitative}, showing that H-SemiS demonstrates superior performance across nearly all metrics. Among supervised methods, H-SemiS achieves the strongest overall performance. Only \cite{Lo2023dlseptic} achieve a marginally higher recall by 0.1, while the best competing supervised approach, \cite{Teoh2023dhl}, still lags by 1.8. These gaps indicate that H-SemiS exhibits stronger feature learning under limited supervision. H-SemiS also surpasses all self-supervised methods, with \cite{Wu2023} yielding the lowest accuracy, highlighting the limitations of learning from unlabeled data alone. Among semi-supervised approaches, \cite{Huo2022dcmt} achieves the closest performance with an F1-score of 82.7, whereas H-SemiS maintains a clear advantage with an F1-score of 84.7. Although \cite{Burton2020} attains the highest precision among existing methods, it still falls short of H-SemiS by 3.2. Overall, these results demonstrate the effectiveness of the hierarchical design and self-supervised components in enhancing feature learning and maintaining balanced performance, thereby establishing H-SemiS as a strong framework for grading osteoarthritis severity with limited labeled data.

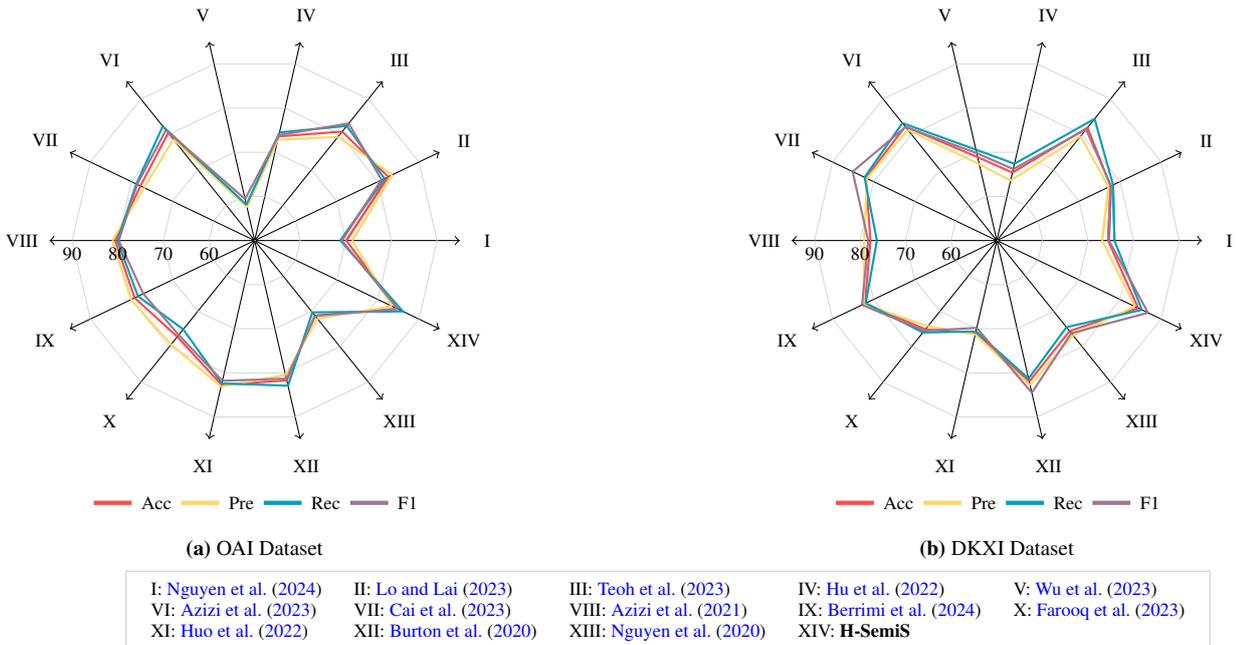
\begin{figure*}[!b]
\captionsetup[subfigure]{justification=centering}
    \centering
    \hspace{-12mm}
    \begin{subfigure}[t]{0.49\textwidth}\centering
    \begin{tikzpicture}
    \tkzKiviatDiagramFromFile
        [scale=.6,
        label distance=.5cm,
        gap     = 1,
        label space=1.1,  
        lattice = 4,
        radial style ->]{oai.dat}
    \tkzKiviatLineFromFile
        [thick,
        color      = cc1,
        ball color = blue,
        ]{oai.dat}{4}
    \tkzKiviatLineFromFile
        [thick,
        color      = cc2,
        ball color = blue,
        ]{oai.dat}{3}
    \tkzKiviatLineFromFile
        [thick,
        color      = cc3,
        ball color = blue,
        ]{oai.dat}{2}
    \tkzKiviatLineFromFile
        [thick,
        color      = cc4,
        ball color = red,
        ]{oai.dat}{1} 
    \node[anchor=south] at (-1,-0.65) {\scriptsize 60}; 
    \node[anchor=south] at (-2,-0.65) {\scriptsize 70}; 
    \node[anchor=south] at (-3,-0.65) {\scriptsize 80}; 
    \node[anchor=south] at (-4,-0.65) {\scriptsize 90}; 

    \node[anchor=north, draw=none] at (0, -5.25) {
    \vspace{0.5cm} 
    \begin{tikzpicture}[baseline=(current bounding box.center)] 
        \matrix[anchor=north, column sep=1 mm] {
            \draw[ultra thick, color=cc1] (0,0) -- (0.5,0); & 
            \node[anchor=mid, inner sep=0] {\scriptsize Acc}; &
            \draw[ultra thick, color=cc2] (0,0) -- (0.5,0); & 
            \node[anchor=mid, inner sep=0] {\scriptsize Pre}; &
            \draw[ultra thick, color=cc3] (0,0) -- (0.5,0); & 
            \node[anchor=mid, inner sep=0] {\scriptsize Rec}; &
            \draw[ultra thick, color=cc4] (0,0) -- (0.5,0); & 
            \node[anchor=mid, inner sep=0] {\scriptsize F1}; \\
        };
    \end{tikzpicture}
    \vspace{0.5cm} 
    };
    \node[anchor=north, draw=none, yshift=-20pt] at (0, -5.25) {
        \footnotesize
        \textbf{(a)} OAI Dataset
    };
    \end{tikzpicture}
    \end{subfigure}
    \hspace{6mm}
    \begin{subfigure}[t]{0.49\textwidth}\centering    
    \begin{tikzpicture}
        \tkzKiviatDiagramFromFile
            [scale=.6,
            label distance=.5cm,
            gap     = 1,
            label space=1.1,  
            lattice = 4,
            radial style ->]{dkxi.dat}
        \tkzKiviatLineFromFile
            [thick,
            color      = cc1,
            ball color = blue,
            ]{dkxi.dat}{4}
        \tkzKiviatLineFromFile
            [thick,
            color      = cc2,
            ball color = blue,
            ]{dkxi.dat}{3}
        \tkzKiviatLineFromFile
            [thick,
            color      = cc3,
            ball color = blue,
            ]{dkxi.dat}{2}
        \tkzKiviatLineFromFile
            [thick,
            color      = cc4,
            ball color = red,
            ]{dkxi.dat}{1} 
        \node[anchor=south] at (-1,-0.65) {\scriptsize 60}; 
        \node[anchor=south] at (-2,-0.65) {\scriptsize 70}; 
        \node[anchor=south] at (-3,-0.65) {\scriptsize 80}; 
        \node[anchor=south] at (-4,-0.65) {\scriptsize 90}; 
    
        \node[anchor=north, draw=none] at (0, -5.25) {
        \begin{tikzpicture}[baseline=(current bounding box.center)] 
            \matrix[anchor=north, column sep=1 mm] {
                \draw[ultra thick, color=cc1] (0,0) -- (0.5,0); & 
                \node[anchor=mid, inner sep=0] {\scriptsize Acc}; &
                \draw[ultra thick, color=cc2] (0,0) -- (0.5,0); & 
                \node[anchor=mid, inner sep=0] {\scriptsize Pre}; &
                \draw[ultra thick, color=cc3] (0,0) -- (0.5,0); & 
                \node[anchor=mid, inner sep=0] {\scriptsize Rec}; &
                \draw[ultra thick, color=cc4] (0,0) -- (0.5,0); & 
                \node[anchor=mid, inner sep=0] {\scriptsize F1}; \\
            };
        \end{tikzpicture}
        };
        \node[anchor=north, draw=none, yshift=-20pt] at (0, -5.25) {
            \footnotesize
            \textbf{(b)} DKXI Dataset
        };
    \end{tikzpicture}    
   
    \end{subfigure}
    \begin{tikzpicture}
    \node[draw=black!20] (box) {
        \scriptsize        
        \begin{tabular}{lllll}
        I: \cite{Nguyen2024mat} & II: \cite{Lo2023dlseptic} & III: \cite{Teoh2023dhl} & IV:  \cite{Hu2022enn} & V: \cite{Wu2023} \\
        VI: \cite{Azizi2023culp} & VII: \cite{Cai2023} & VIII: \cite{Azizi2021} & IX: \cite{Berrimi2024mri} & X: \cite{Farooq2023dcaae} \\
        XI: \cite{Huo2022dcmt} & XII: \cite{Burton2020} & XIII: \cite{Nguyen2020semix} & XIV: \textbf{H-SemiS} & \\
        \end{tabular}
    };
    \end{tikzpicture}
    \caption{Quantitative evaluation of the proposed H-SemiS framework (XIV) against existing baselines (I-XIII) for KOA severity grading. The radar plots compare the Acc, Pre, Rec, F1-score against different competing baselines on (a) OAI dataset~\citep{ChenOAI} and (b) DKXI dataset~\citep{GornaleDKXI}.}
    \label{fig:quant-eval}
\end{figure*}

\vspace{2pt}
\textbf{(ii) Analysis on DKXI Dataset:}
For the DKXI dataset, H-SemiS outperforms all competing baselines across every evaluation metric, as shown in \cref{tab:quantitative}, confirming its effectiveness in grading osteoarthritis severity. Among supervised methods, \cite{Teoh2023dhl} achieves the closest performance with an F1-score of 81.9. Although a self-supervised method attains the second-highest accuracy, H-SemiS remains superior by 1.7, reflecting stronger discriminative feature learning. \cite{Cai2023} reports the highest accuracy among self-supervised approaches, followed by H-SemiS. \cite{Azizi2023culp} achieves the best self-supervised F1-score of 82.2 but still trails H-SemiS by 2.2. This gap reflects the advantage of combining self-supervision with hierarchical learning. Among semi-supervised methods, \cite{Burton2020} and \cite{Berrimi2024mri} show competitive results. However, despite achieving 84.5 accuracy and 82.8 recall, \cite{Burton2020} remains below H-SemiS. Overall, these results show that H-SemiS effectively addresses limitations in prior work. The framework leverages quantum principles to capture complex patterns in knee X-ray data. In addition, the use of proxy labels filtered from noisy and ambiguous samples supports learning more discriminative features for KOA severity grading.

We utilize radar plots to visualize the quantitative results, as shown in \cref{fig:quant-eval}. These plots provide a comparative view of Acc, Pre, Rec, and F1-score achieved by the proposed framework and competing baselines on the OAI~\citep{ChenOAI} and DKXI~\citep{GornaleDKXI} datasets. We further report confusion matrices for both datasets in \cref{fig:cm} to illustrate class-wise prediction behavior.

\begin{figure*}[!ht]
\captionsetup[subfigure]{justification=centering}
    \centering
    \begin{subfigure}[t]{0.48\linewidth}\centering
    \def\myConfMat{{%
        {676, 67, 28, 0, 0},%
        {0, 653, 97, 0, 0},%
        {1, 34, 667, 62, 0},%
        {1, 7, 58, 294, 19},%
        {44, 11, 29, 0, 288}%
    }}

    \begin{tikzpicture}[scale=\myScale, font={\scriptsize}]
    \tikzset{vertical label/.style={rotate=90,anchor=east}}
    \tikzset{diagonal label/.style={rotate=45,anchor=north east}}

    \foreach \y in {1,...,\numClasses}
    {
        \node [anchor=east] at (0.4,-\y) {\pgfmathparse{\classNames[\y-1]}\pgfmathresult};

        \foreach \x in {1,...,\numClasses}
        {
            \def\totSamples{0}
            \foreach \ll in {1,...,\numClasses}
            {
                \pgfmathparse{\myConfMat[\ll-1][\x-1]}
                \xdef\totSamples{\totSamples+\pgfmathresult}
            }
            \pgfmathparse{\totSamples} \xdef\totSamples{\pgfmathresult}

            \begin{scope}[shift={(\x,-\y)}]
                \def\mVal{\myConfMat[\y-1][\x-1]}
                \pgfmathtruncatemacro{\r}{\mVal}

                \ifthenelse{\r=0}{
                    \def\fillcol{myDarkBlue!3}
                    \def\txtcol{black}
                }{
                    \pgfmathparse{\totSamples>0 ? round(\r/\totSamples*100) : 0}
                    \pgfmathtruncatemacro{\rawP}{\pgfmathresult}
                    
                    \pgfmathtruncatemacro{\p}{max(20, \rawP)}
                    
                    \def\fillcol{myDarkBlue!\p}
                    
                    \ifthenelse{\p<50}{\def\txtcol{black}}{\def\txtcol{white}}
                }

                \coordinate (C) at (0,0);
                \node[
                    text=\txtcol,
                    align=center,
                    fill=\fillcol,
                    minimum size=\myScale*10mm,
                    inner sep=0,
                    ] (C) {\r};

                \ifthenelse{\y=\numClasses}{
                \node [] at ($(C)-(0,0.75)$)
                {\pgfmathparse{\classNames[\x-1]}\pgfmathresult};}{}
            \end{scope}
        }
    }
    \coordinate (yaxis) at (-0.9,0.5-\numClasses/2);
    \coordinate (xaxis) at (0.5+\numClasses/2, -\numClasses-1.2);
    \node [vertical label] at (yaxis) {Actual Class};
    \node [] at (xaxis) {Predicted Class};
    \end{tikzpicture}
    \caption{OAI Dataset Evaluation}
    \label{fig:cm-oai}
    \end{subfigure}\hfill
    \begin{subfigure}[t]{0.48\linewidth}\centering
    \def\myConfMat{{%
        {106, 3, 4, 2, 0},%
        {5, 94, 0, 1, 3},%
        {2, 3, 95, 2, 1},%
        {0, 1, 6, 38, 8},%
        {5, 8, 0, 2, 37}%
    }}

    \begin{tikzpicture}[scale=\myScale, font={\scriptsize}]
    \tikzset{vertical label/.style={rotate=90,anchor=east}}
    \tikzset{diagonal label/.style={rotate=45,anchor=north east}}

    \foreach \y in {1,...,\numClasses}
    {
        \node [anchor=east] at (0.4,-\y) {\pgfmathparse{\classNames[\y-1]}\pgfmathresult};
        \foreach \x in {1,...,\numClasses}
        {
            \def\totSamples{0}
            \foreach \ll in {1,...,\numClasses}
            {
                \pgfmathparse{\myConfMat[\ll-1][\x-1]}
                \xdef\totSamples{\totSamples+\pgfmathresult}
            }
            \pgfmathparse{\totSamples} \xdef\totSamples{\pgfmathresult}

            \begin{scope}[shift={(\x,-\y)}]
                \def\mVal{\myConfMat[\y-1][\x-1]}
                \pgfmathtruncatemacro{\r}{\mVal}

                \ifthenelse{\r=0}{
                    \def\fillcol{myDarkBlue!3} 
                    \def\txtcol{black}
                }{
                    \pgfmathparse{\totSamples>0 ? round(\r/\totSamples*100) : 0}
                    \pgfmathtruncatemacro{\rawP}{\pgfmathresult}
                    
                    \pgfmathtruncatemacro{\p}{max(20, \rawP)}
                    
                    \def\fillcol{myDarkBlue!\p}
                    \ifthenelse{\p<50}{\def\txtcol{black}}{\def\txtcol{white}}
                }

                \coordinate (C) at (0,0);
                \node[
                    text=\txtcol,
                    align=center,
                    fill=\fillcol,
                    minimum size=\myScale*10mm,
                    inner sep=0,
                    ] (C) {\r};

                \ifthenelse{\y=\numClasses}{
                \node [] at ($(C)-(0,0.75)$)
                {\pgfmathparse{\classNames[\x-1]}\pgfmathresult};}{}
            \end{scope}
        }
    }
    \coordinate (yaxis) at (-0.9,0.5-\numClasses/2);
    \coordinate (xaxis) at (0.5+\numClasses/2, -\numClasses-1.2);
    \node [vertical label] at (yaxis) {Actual Class};
    \node [] at (xaxis) {Predicted Class};
    \end{tikzpicture}
    \caption{DKXI Dataset Evaluation}
    \label{fig:cm-dkxi}
    \end{subfigure}

    \caption{Confusion matrices displaying the classification performance on the OAI and DKXI datasets.}
    \label{fig:cm}
\end{figure*}

\begin{figure*}[!b]
    \centering
    \includegraphics[width=\linewidth]{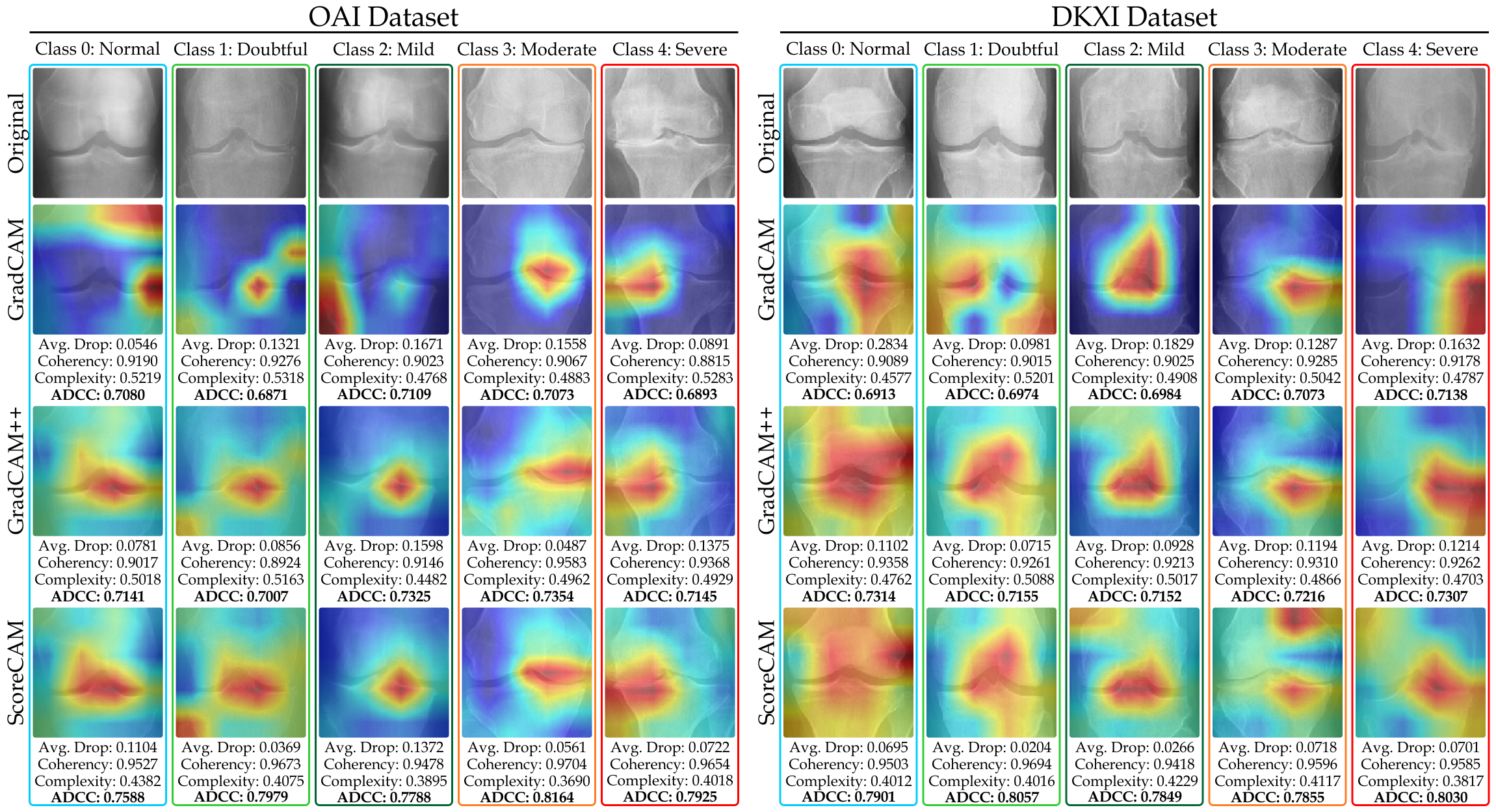}
    \caption{Qualitative evaluation of the H-SemiS framework on the OAI and DKXI datasets across various KOA severity levels. The consistently high ADCC scores suggest that the framework maintains confidence while accurately highlighting the relevant features for prediction.}
    \label{fig:qualitative}
\end{figure*}

\subsubsection{Qualitative Analysis}
\label{sec:qualitative}
We evaluate the qualitative performance of H-SemiS using Explainable Artificial Intelligence (XAI) methods, including Grad-CAM~\citep{Selvaraju2020gradcam}, Grad-CAM++~\citep{Chattopadhay2018gradcam++}, and Score-CAM~\citep{Wang2020scorecam}. To assess the reliability and effectiveness of the resulting explanation maps, we use the ADCC score~\citep{Poppi2021adcc}, which identifies critical regions for KOA severity. The ADCC score $(\chi_{\text{ADCC}})$ is defined as the harmonic mean of Average Drop $(\chi_{\text{AvD}})$, Coherency $(\chi_{\text{Co}})$, and Complexity $(\chi_{\text{Cx}})$, and is computed as follows:
\begin{equation}
\label{eq:adcc}
    \chi_{ADCC} = 3 \left(\frac{1}{1-\chi_{AvD}} + \frac{1}{\chi_{Co}} + \frac{1}{1-\chi_{Cx}} \right)^{-1}
\end{equation}

\cref{fig:qualitative} visualizes class activation maps with ADCC scores, demonstrating the ability of H-SemiS to localize clinically relevant features. Higher ADCC values indicate more reliable and interpretable explanations aligned with the framework’s predictions. A lower $\chi_{\text{AvD}}$ reflects reduced confidence when highlighted regions are removed, while $\chi_{\text{Co}}$ measures the consistency of activation maps with expert interpretation. The $\chi_{\text{Cx}}$ term ensures that highlighted regions remain compact and meaningful. Together, these components provide quantitative validation of the interpretability of H-SemiS and its clinical relevance for assessing KOA.

\begin{table*}[!ht]
\footnotesize
\centering
\caption{Comparative performance analysis of H-SemiS against state-of-the-art models on the OAI and DKXI datasets. \textcolor{horg}{\textbf{Orange}} and \textcolor{sblue}{\textbf{blue}} denote the best and second-best results. $|\Downarrow_d|$ indicates the absolute performance drop relative to H-SemiS.}    
\begin{NiceTabular*}{\linewidth}{@{\extracolsep{\fill}}l c@{}c@{}c@{}c@{}c@{}c@{}c@{}c@{} c@{} c@{}c@{}c@{}c@{}c@{}c@{}c@{}c}
    \toprule
    \midrule

    \textbf{Methods} & \multicolumn{8}{c}{\textbf{OAI Dataset}} & & \multicolumn{8}{c}{\textbf{DKXI Dataset}} \\ 
    \cmidrule{2-9} \cmidrule{11-18}
    
    & \textbf{Acc} & $|\Downarrow_d|$ & \textbf{Pre} & $|\Downarrow_d|$ & \textbf{Rec} & $|\Downarrow_d|$ & \textbf{F1} & $|\Downarrow_x|$ & & \textbf{Acc} & $|\Downarrow_d|$ & \textbf{Pre} & $|\Downarrow_d|$ & \textbf{Rec} & $|\Downarrow_d|$ & \textbf{F1} & $|\Downarrow_d|$ \\                  
    \midrule
    ResNet50~\citep{He_2016ResNet} & 72.8 & 12.1 & 71.9 & 14.2 & 73.0 & 10.4 & 72.4 & 12.3 &  & 75.9 & 10.9 & 73.8 & 11.5 & 75.8 & 7.8  & 74.8 & 9.60  \\
    DenseNet201~\citep{Huang_2017DenseNet} & 74.8 & 10.1 & 75.3 & 10.8 & 75.6 & 7.8  & 75.4 & 9.3  &  & 78.6 & 8.2  & 77.3 & 8.0  & 76.9 & 6.7  & 77.1 & 7.26  \\
    ViT-B/32~\citep{Dosovitskiy2021vit} & 70.4 & 14.5 & 72.6 & 13.5 & 73.1 & 10.3 & 72.8 & 11.9 &  & 73.8 & 13.0 & 73.6 & 11.7 & 75.1 & 8.5  & 74.3 & 10.05 \\
    MedViT~\citep{Manzari_2023MedVit} & 76.9 & 8.0  & 77.2 & 8.9  & 77.3 & 6.1  & 77.2 & 7.5  &  & 77.0 & 9.8  & 79.5 & 5.8  & 80.6 & 3.0  & 80.0 & 4.35  \\
     CMT~\citep{Guo2022cmt} & 73.2 & 11.7 & 74.8 & 11.3 & 72.5 & 10.9 & 73.6 & 11.1 &  & 75.7 & 11.1 & 76.8 & 8.5  & 73.0 & 10.6 & 74.9 & 9.53  \\
    $\pi$-Model~\citep{Laine2017pi} & 79.5 & 5.4  & \textcolor{sblue}{\textbf{82.2}} & 3.9  & 78.6 & 4.8  & 80.4 & 4.3  &  & 80.2 & 6.6  & \textcolor{horg}{\textbf{85.8}} & 0.5  & 80.7 & 2.9  & 83.2 & 1.23  \\
     Mean-Teacher~\citep{Tarvainen2017mt} & \textcolor{sblue}{\textbf{80.3}} & 4.6  & 81.5 & 4.6  & \textcolor{sblue}{\textbf{82.3}} & 1.1  & \textcolor{sblue}{\textbf{81.9}} & 2.8  &  & \textcolor{sblue}{\textbf{85.5}} & 1.3  & 84.1 & 1.2  & \textcolor{sblue}{\textbf{82.8}} & 0.8  & \textcolor{sblue}{\textbf{83.5}} & 0.93  \\

    \cdashline{1-18}[0.5pt/2pt]\noalign{\vskip 0.5ex}
    \rowcolor{lightgray}
    \textbf{H-SemiS} & \textcolor{horg}{\textbf{84.9}} & 0.0  & \textcolor{horg}{\textbf{86.1}} & 0.0  & \textcolor{horg}{\textbf{83.4}} & 0.0  & \textcolor{horg}{\textbf{84.7}} & 0.0  &  & \textcolor{horg}{\textbf{86.8}} & 0.0  & \textcolor{sblue}{\textbf{85.3}} & 0.0  & \textcolor{horg}{\textbf{83.6}} & 0.0  & \textcolor{horg}{\textbf{84.4}} & 0.0 \\
    
    \midrule
    \bottomrule    
\end{NiceTabular*}
\label{tab:sota}
\end{table*}

\subsection{Comparison with State-of-the-art models}
\label{sec:sota}
We compare H-SemiS with state-of-the-art (SOTA) models to demonstrate its effectiveness in KOA severity grading, as summarized in \cref{tab:sota}. Evaluations are conducted on the OAI \citep{ChenOAI} and DKXI \citep{GornaleDKXI} datasets against a diverse set of SOTA methods, including ResNet50 \citep{He_2016ResNet}, DenseNet201 \citep{Huang_2017DenseNet}, ViT-B/32 \citep{Dosovitskiy2021vit}, MedViT \citep{Manzari_2023MedVit}, CMT \citep{Guo2022cmt}, $\pi$-Model \citep{Laine2017pi}, and Mean-Teacher \citep{Tarvainen2017mt}. Across both datasets, H-SemiS achieves the highest accuracy, at 84.9 on OAI and 86.8 on DKXI, surpassing Mean-Teacher at 80.3 and 85.5, respectively. This improvement reflects stronger feature learning under limited supervision, a strength also observed in the Mean-Teacher model. While Mean-Teacher and the $\pi$-Model benefit from effective pseudo-labeling, H-SemiS further improves performance by leveraging filtered proxy labels and quantum-inspired feature learning to capture fine-grained knee structures and nonlinear patterns. In contrast, transfer learning models such as ResNet50 and DenseNet201, as well as transformer-based architectures including ViT-B/32, MedViT, and CMT, exhibit limited adaptability to KOA grading, resulting in weaker performance. Overall, these findings establish H-SemiS as a robust and superior framework for osteoarthritis severity grading under limited supervision.

\subsection{Ablative Experiments}
\label{sec:ablation}
We conduct an ablation study on the OAI~\citep{ChenOAI} dataset to assess the contribution of individual components in the H-SemiS framework. By varying one component at a time while keeping all other settings fixed, we quantify their impact on overall performance. The following subsections summarize the results.

\begin{figure}[!ht]
        \centering
        \includegraphics[width=\linewidth]{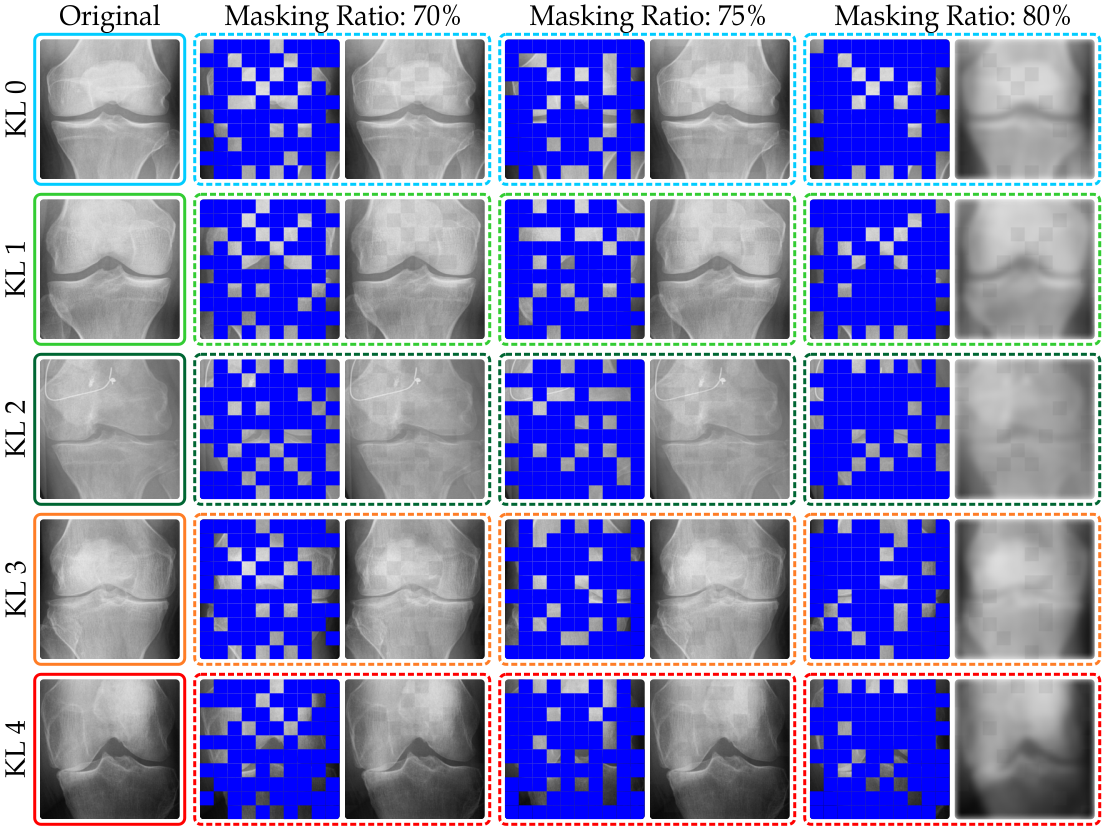}
        \caption{KXR reconstruction examples across different KL grades. The first column represents the original images, while the subsequent columns display masked and reconstructed images with varying masking ratios (70\%, 75\%, and 80\%).}
        \label{fig:abl-imrec}
\end{figure}

\begin{figure}[!ht]
        \centering    
        \begin{tikzpicture}
        \begin{axis}[
            width=\linewidth,        
            height=4cm,
            xlabel={Mask Ratio (\%)},
            ylabel={Accuracy},
            xmin=55, xmax=95,
            ymin=82, ymax=85,
            xtick={60,65,70,75,80,85,90},
            ytick={82.5,83.0,83.5,84.0,84.5},
            grid=both,
            grid style={dotted, gray},
            major grid style={line width=0.5pt},
            minor tick num=1,
            tick label style={font=\footnotesize},
            xlabel style={font=\footnotesize}, 
            ylabel style={font=\footnotesize}, 
            legend pos=south east
        ]
            \addplot+[
                color=cyan,
                mark=star, 
                mark options={solid},
                mark size=3.5pt, 
                line width=1.25pt, 
                nodes near coords, 
                nodes near coords style={
                    font=\footnotesize, 
                    anchor=south west, 
                    yshift=-8pt, 
                    xshift=-1pt 
                },
            ] coordinates {
                (60,82.8)
                (65,83.2)
                (70,83.8)
                (75,84.9)
                (80,83.1)
                (85,82.7)
                (90,82.3)
            };
        \end{axis}
        \end{tikzpicture}
        \caption{Mask ratio ablation study highlighting the impact of masking on H-SemiS performance.}
        \label{fig:plot-imrec}
\end{figure}
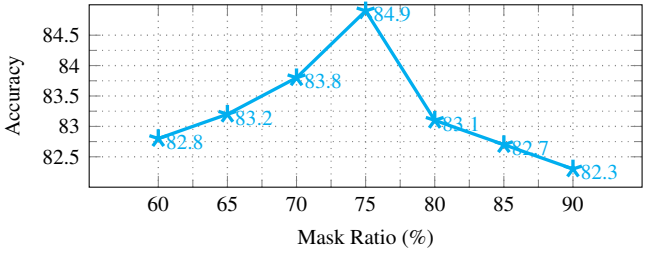

\vspace{2pt}
\noindent\textit{(i) Effect of Image Reconstruction:}
We examine the contribution of Masked Image Reconstruction (MI-Rec) (see \cref{sec:mirec}) by analyzing the effects of the masking ratio and the choice of reconstruction technique. \cref{tab:abl-imrec} reports performance across masking ratios $(N_m)$, showing optimal results at 75\%. Performance drops by 1.1 and 1.8 when the masking ratio is set to 70\% and 80\%, respectively. \cref{fig:abl-imrec} visualizes reconstructed samples, where excessive masking (80\%) yields blurry reconstructions with poorly defined edges that degrade performance. \cref{fig:plot-imrec} further shows a peak accuracy of 84.9 at 75\%, with accuracy decreasing as the masking ratio moves away from this value. This trend indicates that MI-Rec performs best at 75\%, where missing patches are reconstructed uniformly without bias toward specific regions.

\begin{table}[!ht]  \footnotesize
\setlength{\tabcolsep}{2pt}
\centering
\caption{Effect of masking ratio and reconstruction techniques on H-SemiS performance using the OAI dataset. $|\Downarrow_d|$ represents the absolute performance drop relative to H-SemiS.}    
    \begin{threeparttable}            
    \begin{NiceTabular*}{0.5\textwidth}{@{\extracolsep{\fill}}ll cc cc cc cc }
    \toprule
    \midrule
    & \textbf{Methods} & \textbf{Acc} & $|\Downarrow_d|$ & \textbf{Pre} & $|\Downarrow_d|$ & \textbf{Rec} & $|\Downarrow_d|$ & \textbf{F1} & $|\Downarrow_d|$\\                  
    \midrule
    I
    & N$_{m} = 0.70$ & 83.8 & 1.1 & 82.5 & 3.6 & 83.1 & 0.3 & 82.8 & 1.9 \\
    & N$_{m} = 0.80$ & 83.1 & 1.8 & 84.3 & 1.8 & 82.6 & 0.8 & 83.4 & 1.3 \\

    \cdashline{1-10}[0.5pt/2pt]\noalign{\vskip 0.5ex}
    II
    & MAE & 79.5 & 5.4 & 80.1 & 6.0 & 77.2 & 6.2 & 78.6 & 6.1 \\
    & CMAE & 80.4 & 4.5 & 82.6 & 3.5 & 81.9 & 1.5 & 82.2 & 2.5 \\
    & ViT-AE++ & 78.5 & 6.4 & 79.7 & 6.4 & 82.6 & 0.8 & 81.1 & 3.6 \\
    & GAN-MAE & 83.1 & 1.8 & 80.4 & 5.7 & 83.2 & 0.2 & 81.8 & 2.9 \\
    \cdashline{1-10}[0.5pt/2pt]\noalign{\vskip 0.5ex}
    \rowcolor{lightgray}
    \textbf{Ours}
    & \textbf{N}$\mathbf{_{m} {=} 0.75}$\textbf{+ MI-Rec} & \textbf{84.9} & \textbf{0.0} & \textbf{86.1} & \textbf{0.0} & \textbf{83.4} & \textbf{0.0} & \textbf{84.7} & \textbf{0.0} \\
    
    \midrule
    \bottomrule    
    \end{NiceTabular*}
    \begin{tablenotes}
    \centering    
      \scriptsize
      \item I: Effect of Masking Ratio; II: Effect of Reconstruction Technique
    \end{tablenotes}
    \end{threeparttable}
    \label{tab:abl-imrec}
\end{table}

We next compare reconstruction techniques at a fixed masking ratio of 75\%. As summarized in \cref{tab:abl-imrec}, we evaluate MAE \citep{He2022mae}, CMAE \citep{Huang2024cmae}, ViT-AE++ \citep{Prabhakar2024vitae}, and GAN-MAE \citep{Fei2023maegan} against the proposed MI-Rec. GAN-MAE achieves the second-best performance, while MI-Rec consistently outperforms all alternatives across metrics. These results suggest that MI-Rec better preserves structural consistency and key KOA markers, including cartilage thickness and joint space narrowing. The combined choice of an optimal masking ratio and reconstruction strategy prevents spatial feature loss and enables high-fidelity reconstructions, leading to more accurate KOA classification.

\begin{figure}[!ht]
    \centering
    \includegraphics[width=\linewidth]{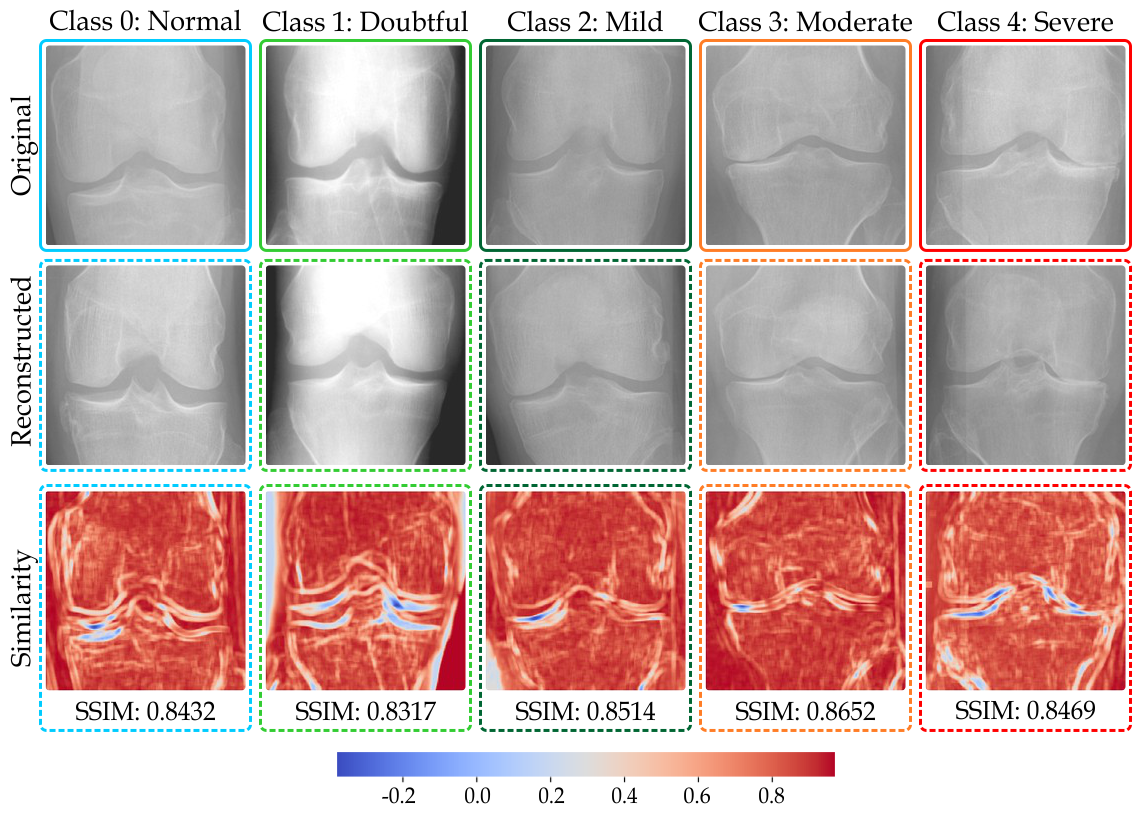}
    \caption{Structural Similarity Index (SSIM) values assess the structural similarity between original and reconstructed KXR samples. Higher values indicate greater structural resemblance, while the blue regions highlight the differences.}
    \label{fig:abl-sitm}
\end{figure}

\vspace{2pt}
\noindent\textit{(ii) Effect of Template Matching:}
\cref{tab:abl-sirl} analyzes the role of the Similarity-aware Reconstructed Image Labeler (SiRL), described in \cref{sec:sirl}, within the H-SemiS framework. SiRL assigns proxy labels to reconstructed unlabeled samples based on similarity to predefined templates. We evaluate thresholds from $\tau_{60}$ to $\tau_{90}$ and identify $\tau_{80}$ as optimal, achieving the best performance across all metrics. Lower thresholds ($\tau_{60}$, $\tau_{70}$) admit more reconstructed samples but introduce noise, reducing performance by 6.3\% and 1.1\%, respectively. A higher threshold ($\tau_{90}$) admits too few samples, limiting class imbalance mitigation and causing a 5\% drop. The strong performance at $\tau_{80}$ indicates effective selection of reliable reconstructed samples. We further assess reliability using Structural Similarity Index Measure (SSIM) between randomly selected original and reconstructed images. As shown in \cref{fig:abl-sitm}, $\tau_{80}$ yields high SSIM scores, confirming that SiRL assigns accurate proxy labels while MI-Rec preserves critical structural details.

\begin{table}[!ht]  
\footnotesize
\centering
\setlength{\tabcolsep}{2pt}
\caption{Effect of the threshold on H-SemiS performance for labeling reconstructed samples in the OAI dataset. $|\Downarrow_d|$ represents the absolute performance drop relative to H-SemiS.}
    \begin{NiceTabular*}{0.5\textwidth}{@{\extracolsep{\fill}}l cc cc cc cc }
    \toprule
    \midrule
    \textbf{Methods} & \textbf{Acc} & $|\Downarrow_d|$ & \textbf{Pre} & $|\Downarrow_d|$ & \textbf{Rec} & $|\Downarrow_d|$ & \textbf{F1} & $|\Downarrow_d|$\\                  
    \midrule       
    H-SemiS $(\tau_{60})$ & 78.6 & 6.3 & 75.2 & 10.9 & 76.7 & 6.7 & 75.9 & 8.8 \\
    H-SemiS $(\tau_{70})$ & 83.8 & 1.1 & 84.1 & 2.0  & 83.6 & 0.2 & 83.8 & 0.9 \\
    H-SemiS $(\tau_{90})$ & 79.9 & 5.0 & 81.2 & 4.9  & 80.3 & 3.1 & 80.7 & 4.0 \\
    
    \cdashline{1-9}[0.5pt/2pt]\noalign{\vskip 0.5ex}
    \rowcolor{lightgray}
    \textbf{H-SemiS} $\mathbf{(\tau_{80})}$ & \textbf{84.9} & \textbf{0.0} & \textbf{86.1} & \textbf{0.0} & \textbf{83.4} & \textbf{0.0} & \textbf{84.7} & \textbf{0.0} \\
    
    \midrule
    \bottomrule
    \end{NiceTabular*}
    \label{tab:abl-sirl}
\end{table}

\begin{figure}[!ht]
    \centering
    \begin{tikzpicture}[scale=0.75]
    \drawaxes[diagram angle=30, draw axes]{}; 
    \foreach \r/\text in {0/60, 1/65, 2/70, 3/75, 4/80, 5/85} {
        \node[draw, rectangle, fill=white, inner sep=2pt] at (90:\r cm) {\footnotesize\text};
    }
    \sectorlist[sector angle=30,
    draw sector list={
    4.85cm/1g,  
    5.18cm/1h,  
    5.37cm/1i,  
    5.60cm/1j,  
    5.95cm/1k,  
    6.26cm/black!90,
    1.61cm/1a,  
    1.94cm/1b,  
    2.09cm/1c,  
    2.59cm/1d,  
    2.87cm/1e,  
    3.29cm/1f
    }]{};
    \foreach \i/\angle in {1/15, 2/45, 3/75, 4/105, 5/135, 6/165, 7/195, 8/225, 9/255, 10/285, 11/315, 12/345} {
    \node[anchor=center] at (\angle:5.5cm) { 
        \ifnum\i=12 \footnotesize\textbf{\uppercase\expandafter{\romannumeral\i}}
        \else \footnotesize\uppercase\expandafter{\romannumeral\i} 
        \fi};
    }    

    \scriptsize
    \begin{customlegend}[
        legend entries={
            {I: $1\%$: BN}, {II: $1\%$: BN+NL}, {III: $1\%$: BN+NL+QCN},
            {IV: $5\%$: BN}, {V: $5\%$: BN+NL}, {VI: $5\%$: BN+NL+QCN},
            {VII: $10\%$: BN}, {VIII: $10\%$: BN+NL}, {IX: $10\%$: BN+NL+QCN},
            {X: $20\%$: BN}, {XI: $20\%$: BN+NL}, {\colorbox{lightgray}{\textbf{XII:} $\mathbf{20\%}$: \textbf{BN+NL+QCN}}}
        },
        legend style={
            at={(0,-5.8)}, 
            anchor=north, 
            draw=black!20, 
            column sep=2pt, 
            row sep=0pt 
        },
        legend columns=3 
        ]
        \legendlistcolors{1a,1b,1c,1d,1e,1f,1g,1h,1i,1j,1k,black!90}
        \node[below=1.8cm, align=center, text width=\linewidth] at (current bounding box.south) 
    {\scriptsize BN: Base Network; NL: Normalization Layer; QCN: Quantum Convolution Network};
    \end{customlegend}
    
    \end{tikzpicture}
    \caption{Impact of labeled data percentage on Teacher-Student framework accuracy. Each quadrant shows performance at 1\%, 5\%, 10\%, and 20\% labeled data for three architectural configurations: Base Network (BN), BN with Quantum Convolution Network (QCN), and BN with Normalization Layer (NL) and QCN. Quadrant XII (20\% BN+NL+QCN) demonstrates the best performance.}
    \label{fig:abl-test}
\end{figure}

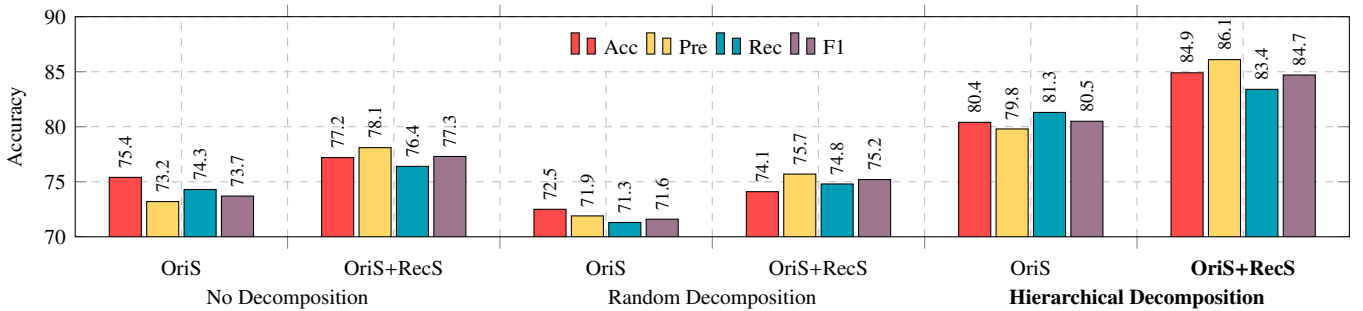
\begin{figure*}[!b]
    \centering
    \pgfplotstableread{
0   75.4	73.2	74.3	73.7
1   77.2	78.1	76.4	77.3
2   72.5	71.9	71.3	71.6
3   74.1	75.7	74.8	75.2
4   80.4	79.8	81.3	80.5
5   84.9	86.1	83.4	84.7
}\dataset
    \begin{tikzpicture}
    \begin{axis}[
        grid=both,
        grid style=dashed,
        ybar,
        width=\textwidth,
        height=4.5cm,
        ymin=70,
        ymax=90,  
        font=\footnotesize,
        ylabel={Accuracy},
        xtick=data,
        xticklabels = {
            \footnotesize\strut OriS,
            \footnotesize\strut OriS+RecS,
            \footnotesize\strut OriS,
            \footnotesize\strut OriS+RecS,
            \footnotesize\strut OriS,
            \footnotesize\strut \textbf{OriS+RecS},
        },
        major x tick style = {opacity=0},
        minor x tick num = 1,
        minor tick length=1ex,
        extra x ticks={0.5, 2.5, 4.5},
        extra x tick labels={No Decomposition, Random Decomposition, \textbf{Hierarchical Decomposition}},
        extra x tick style={
            grid=none,
            tick label style={yshift=-1.2em, font=\footnotesize}
        },
        bar width=12pt,
        enlarge x limits=0.1,
        legend entries={Acc, Pre, Rec, F1},
        legend columns=4,
        legend style={at={(0.5,0.75)}, anchor=south, draw=none, nodes={inner sep=3pt}, font=\footnotesize},
        nodes near coords,
        nodes near coords align={vertical},
        every node near coord/.append style={font=\scriptsize, rotate=90, anchor=west},
    ]
    \addplot[draw=black, fill=acc, nodes near coords] table[x index=0, y index=1] \dataset; 
    \addplot[draw=black, fill=pre, nodes near coords] table[x index=0, y index=2] \dataset; 
    \addplot[draw=black, fill=rec, nodes near coords] table[x index=0, y index=3] \dataset; 
    \addplot[draw=black, fill=f1, nodes near coords] table[x index=0, y index=4] \dataset; 
    \end{axis}
    \end{tikzpicture}
    \caption{Performance comparison of H-SemiS under different decomposition techniques (No Decomposition, Random Decomposition, and Hierarchical Decomposition) using original samples (OriS) and a combination of original and reconstructed samples (OriS+RecS).}
    \label{fig:abl-him}
\end{figure*}

\begin{table}[!ht]  \footnotesize
\setlength{\tabcolsep}{2pt}
\centering
\caption{Effect of the percentage of labeled samples on H-SemiS performance on OAI dataset. BN, NL, and QCN represent the Base Network, Normalization Layer, and Quantum Convolutional Neural Network. $|\Downarrow_d|$ represents the absolute performance drop relative to H-SemiS.}    
    \begin{NiceTabular*}{\linewidth}{@{\extracolsep{\fill}}l c@{}c@{}c cc cc cc cc }
    \toprule
    \midrule
    \% &BN &NL &QCN &\textbf{Acc} & $|\Downarrow_d|$ & \textbf{Pre} & $|\Downarrow_d|$ & \textbf{Rec} & $|\Downarrow_d|$ & \textbf{F1} & $|\Downarrow_d|$\\                 
    \midrule       
    1\% &$\checkmark$ &$\times$ &$\times$ & 66.4 & 18.5 & 65.5 & 20.6 & 63.2 & 20.2 & 64.3 & 20.4 \\
    &$\checkmark$ &$\times$ &$\checkmark$ & 67.7 & 17.2 & 67.2 & 18.9 & 67.1 & 16.3 & 67.1 & 17.6 \\
    &$\checkmark$ &$\checkmark$ &$\checkmark$ & 68.3 & 16.6 & 69.4 & 16.7 & 69.4 & 14.0 & 69.4 & 15.3 \\
    
    \cdashline{1-12}[0.5pt/2pt]\noalign{\vskip 0.5ex}

    5\% &$\checkmark$ &$\times$ &$\times$ & 70.3 & 14.6 & 71.3 & 14.8 & 70.6 & 12.8 & 70.9 & 13.8 \\
    &$\checkmark$ &$\times$ &$\checkmark$ & 71.4 & 13.5 & 71.7 & 14.4 & 72.8 & 10.6 & 72.2 & 12.5 \\
    &$\checkmark$ &$\checkmark$ &$\checkmark$ & 73.1 & 11.8 & 74.7 & 11.4 & 73.9 & 9.5  & 74.3 & 10.4 \\

    \cdashline{1-12}[0.5pt/2pt]\noalign{\vskip 0.5ex}

    10\% &$\checkmark$ &$\times$ &$\times$ & 79.3 & 5.6  & 80.2 & 5.9  & 78.3 & 5.1  & 79.2 & 5.5  \\
    &$\checkmark$ &$\times$ &$\checkmark$ & 80.6 & 4.3  & 81.8 & 4.3  & 78.9 & 4.5  & 80.3 & 4.4  \\
    &$\checkmark$ &$\checkmark$ &$\checkmark$ & 81.4 & 3.5  & 82.2 & 3.9  & 79.4 & 4.0  & 80.8 & 3.9  \\

    \cdashline{1-12}[0.5pt/2pt]\noalign{\vskip 0.5ex}

    20\% &$\checkmark$ &$\times$ &$\times$ & 82.3 & 2.6  & 83.5 & 2.6  & 80.7 & 2.7  & 82.1 & 2.6  \\
    &$\checkmark$ &$\times$ &$\checkmark$ & 83.7 & 1.2  & 84.9 & 1.2  & 81.6 & 1.8  & 83.2 & 1.5  \\
    
    \rowcolor{lightgray}    
    &$\pmb{\checkmark}$ &$\pmb{\checkmark}$ &$\pmb{\checkmark}$ & \textbf{84.9} & \textbf{0.0}  & \textbf{86.1} & \textbf{0.0}  & \textbf{83.4} & \textbf{0.0}  & \textbf{84.7} & \textbf{0.0}  \\
    
    \midrule
    \bottomrule
    \end{NiceTabular*}
    \label{tab:abl-test}
\end{table}

\vspace{2pt}
\noindent\textit{(iii) Effect of Labeling in Teacher-Student Framework:}
\label{sec:labelratios}
We evaluate H-SemiS under varying labeled data ratios (1\%, 5\%, 10\%, and 20\%). For each ratio, we examine different architectural configurations of the quantum-infused teacher–student framework (Q-TeSt) (see \cref{sec:qtest}), including the Base Network (BN), Normalization Layer (NL), and Quantum Convolutional Neural Network (QCN). As reported in \cref{tab:abl-test}, with only 1\% labeled data, the base configuration achieves competitive performance, highlighting robustness under severe data scarcity. As shown in \cref{fig:abl-test}, increasing the labeled ratio from 1\% to 5\% consistently improves performance across all configurations, highlighting the contribution of labeled supervision. This trend persists at 10\% and 20\%, where Q-TeSt further enhances feature extraction by learning complex knee patterns in quantum space. Incorporating QCN strengthens the capture of nonlinear relationships within normalized spatial features. The best performance is achieved at 20\% labeled data using the combined BN+NL+QCN configuration, demonstrating consistent gains across all supervision levels.

\vspace{2pt}
\noindent\textit{(iv) Effect of Decomposition:}
We analyze the impact of decomposition strategies in the hierarchical multi-classifier on H-SemiS performance for KOA severity grading. As shown in \cref{fig:abl-him}, we compare three strategies: no decomposition, random decomposition, and hierarchical decomposition, and examine the use of original samples (OriS) and reconstructed samples (RecS). Without decomposition, the framework performs direct multiclass classification and attains a lower accuracy of 75.4\% using OriS. Random decomposition, which splits classes into binary tasks without accounting for class imbalance, yields inferior results, with performance drops of 12.4\% using OriS and 10.8\% using OriS+RecS relative to hierarchical decomposition. Incorporating reconstructed samples consistently improves performance across all decomposition strategies, demonstrating their effectiveness in mitigating multiclass imbalance. Hierarchical decomposition achieves the best performance, reaching an accuracy of 84.9\% with OriS+RecS, corresponding to a 4.5\% gain over using OriS alone. These results show that hierarchical decomposition effectively integrates information from both sample types. In addition, the proposed hierarchical strategy employs a dual rule-based design (see \cref{sec:him}), supporting balanced learning under multiclass imbalance.

\subsection{Generalizability Evaluation}
\label{sec:generalizability}
\begin{figure*}[!ht]
    \captionsetup[subfigure]{justification=centering}
    \centering
    
    \begin{subfigure}[t]{0.24\linewidth}\centering
        \includegraphics[width=\linewidth]{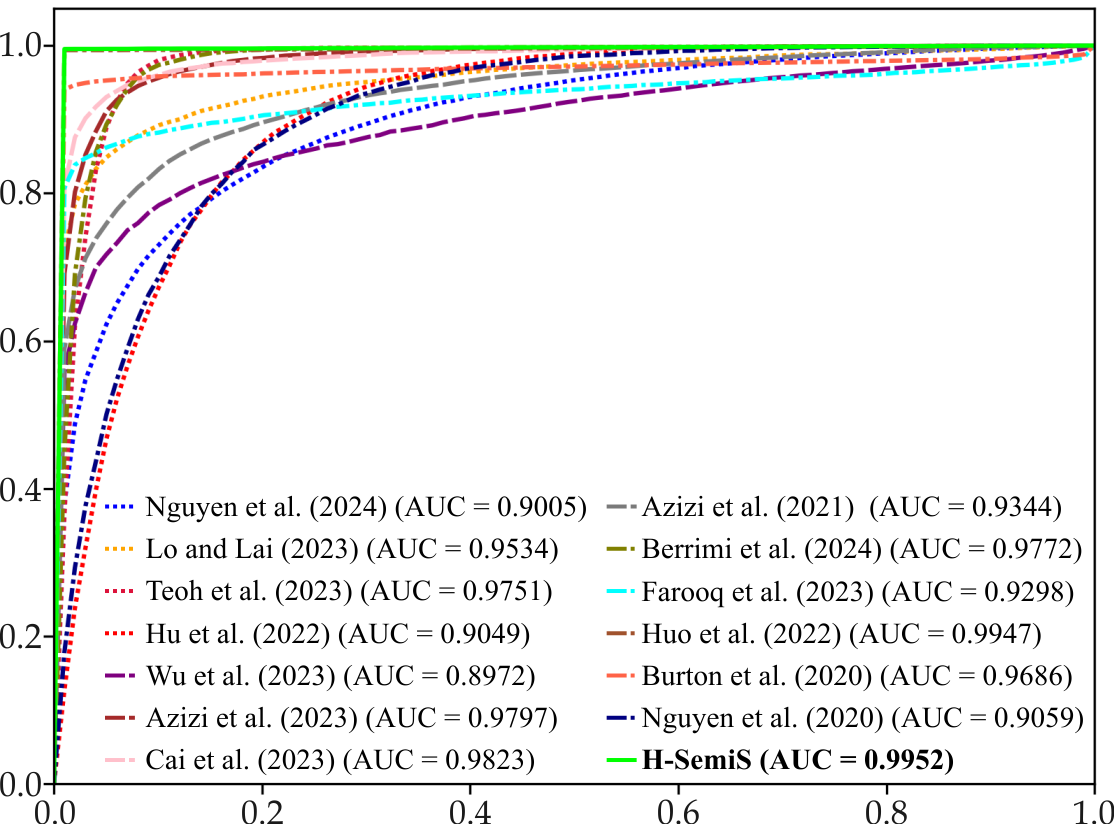}
        \caption{ROC Curve on OP Dataset}
    \end{subfigure}\hfill
    \begin{subfigure}[t]{0.24\linewidth}\centering
        \includegraphics[width=\linewidth]{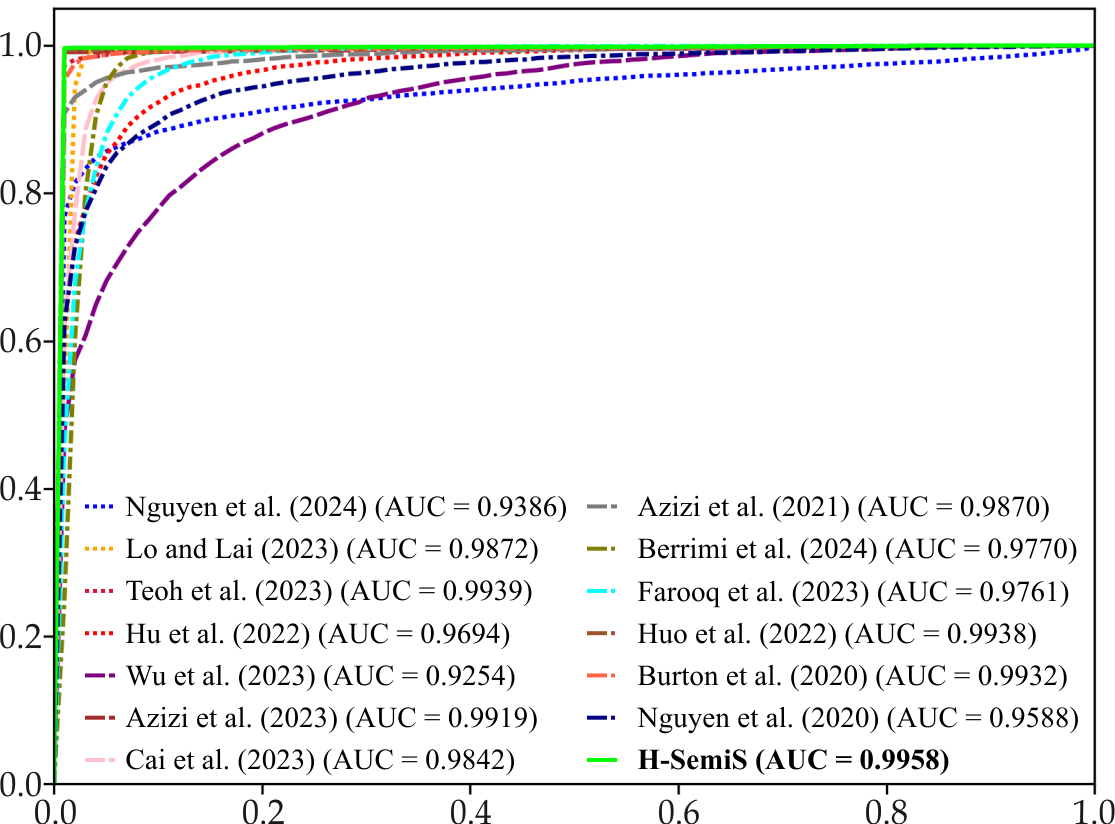}
        \caption{ROC Curve on KO Dataset}
    \end{subfigure}\hfill
    \begin{subfigure}[t]{0.24\linewidth}\centering
        \includegraphics[width=\linewidth]{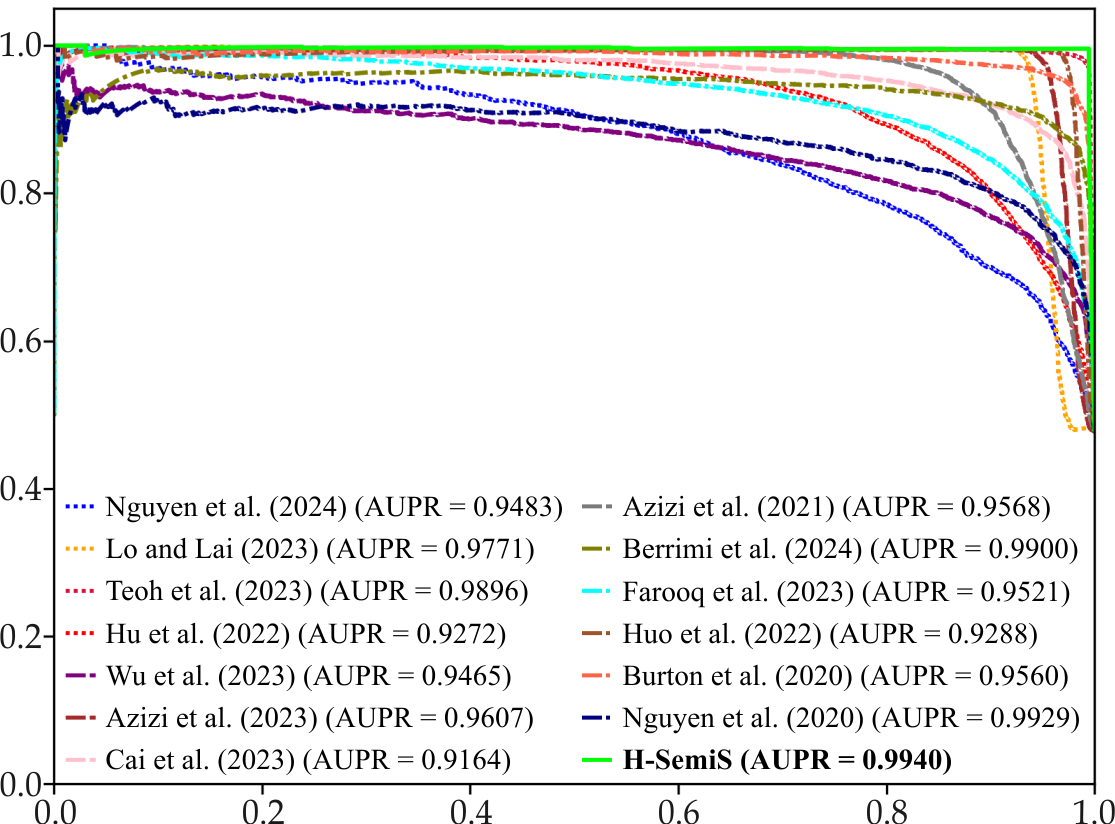}
        \caption{PR Curve on OP Dataset}
    \end{subfigure}\hfill
    \begin{subfigure}[t]{0.24\linewidth}\centering
        \includegraphics[width=\linewidth]{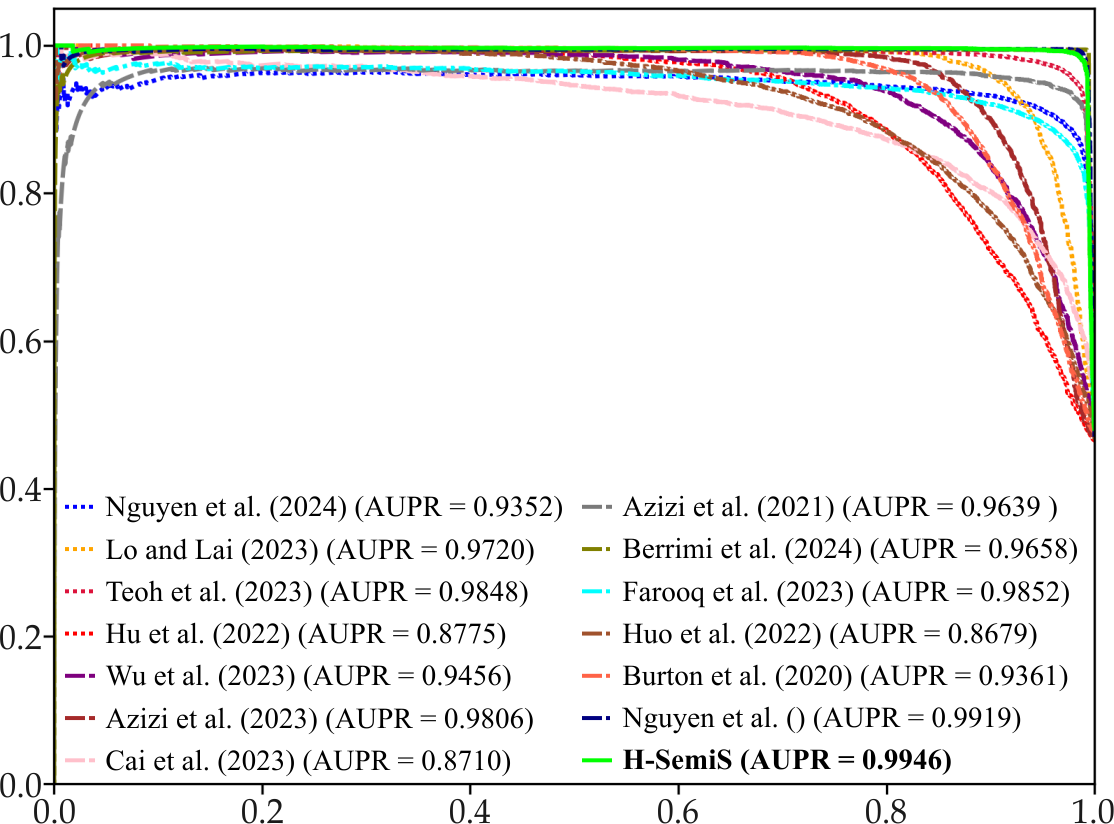}
        \caption{PR Curve on KO Dataset}
    \end{subfigure}
    
    \caption{Performance comparison: ROC and PR curves for OP and KO datasets.}
    \label{fig:roc-pr}
\end{figure*}

We assess the generalizability of H-SemiS through cross-dataset evaluation by training the framework on one dataset and testing it on another, including concatenated datasets. To strengthen this analysis, we also consider additional binary data-sets (see \cref{sec:binarydata}) alongside the multi-class datasets. \cref{table:gen-eval} reports performance across different training and testing distributions for both multi-class and binary settings. H-SemiS performs best when trained and evaluated on the same dataset, while performance decreases when models trained on smaller datasets are tested on larger or combined datasets. For multi-class evaluation, training and testing on OAI~\citep{ChenOAI} yields the highest accuracy of 84.9, which drops by 7.7 and 3.7 when tested on DKXI~\citep{GornaleDKXI} and the combined OAI+DKXI dataset, respectively. A similar trend is observed in the binary setting, where H-SemiS achieves accuracies of 90.8, 92.3, and 91.7 when trained and tested on OP, KO, and OP+KO datasets, respectively. In contrast, training on OP~\citep{TaoOP} and testing on KO~\citep{ShawKO} and OP+KO leads to accuracy reductions of 11.3 and 18.9. We further report ROC and PR curves with AUC and AUPR for the binary datasets in \cref{fig:roc-pr}, demonstrating the robustness of H-SemiS under cross-dataset shifts. Overall, the cross-dataset results confirm that H-SemiS generalizes reliably across both multi-class and binary settings and remains competitive on unseen datasets.

\begin{table*}[!ht]  \footnotesize
\centering
\caption{Evaluation of H-SemiS generalizability across varied training and testing samples.}
\begin{tabular*}{\linewidth}{@{\extracolsep{\fill}}l cccc cccc cccc }
    \toprule
    \midrule
    
    \textbf{{Training}}
    &\multicolumn{12}{c}{\textbf{Testing}} \\
    \midrule
    &\multicolumn{12}{c}{\textbf{Category-1: Multi-class Dataset}} \\
    \cmidrule(l){2-13}
        
    &\multicolumn{4}{c}{\textbf{OAI}}
    &\multicolumn{4}{c}{\textbf{DKXI}}
    &\multicolumn{4}{c}{\textbf{OAI + DKXI}}\\ 
    
    \cmidrule(l){2-5} \cmidrule(l){6-9} \cmidrule(l){10-13}
    & \textbf{Acc} & \textbf{ Pre} & \textbf{ Rec} & \textbf{ F1}
    & \textbf{Acc} & \textbf{ Pre} & \textbf{ Rec} & \textbf{ F1}
    & \textbf{Acc} & \textbf{ Pre} & \textbf{ Rec} & \textbf{ F1}\\

    \cdashline{1-13}[0.5pt/2pt]\noalign{\vskip 0.5ex}
    OAI      & 84.9 & 86.1 & 83.4 & 84.7 & 77.3 & 73.6 & 79.1 & 76.3 & 81.2  & 79.3  & 80.8  & 80.0 \\
    DKXI     & 68.2 & 67.9 & 67.3 & 67.6 & 86.8 & 85.3 & 83.6 & 84.4 & 73.6  & 69.8  & 72.1  & 70.9 \\
    OAI+DKXI & 83.8 & 84.5 & 83.1 & 83.8 & 80.5 & 79.6 & 80.3 & 79.9 & 82.1  & 80.5  & 81.8  & 81.1 \\
    \midrule
    
    &\multicolumn{12}{c}{\textbf{Category-2: Binary Dataset}} \\
    \cmidrule(l){2-13}
    &\multicolumn{4}{c}{\textbf{OP}}
    &\multicolumn{4}{c}{\textbf{KO}}
    &\multicolumn{4}{c}{\textbf{OP + KO}}\\ 
    
    \cmidrule(l){2-5} \cmidrule(l){6-9} \cmidrule(l){10-13}
    & \textbf{Acc} & \textbf{ Pre} & \textbf{ Rec} & \textbf{ F1}
    & \textbf{Acc} & \textbf{ Pre} & \textbf{ Rec} & \textbf{ F1}
    & \textbf{Acc} & \textbf{ Pre} & \textbf{ Rec} & \textbf{ F1}\\
    
    \cdashline{1-13}[0.5pt/2pt]\noalign{\vskip 0.5ex}
    OP       & 90.8 & 92.5 & 91.7 & 92.1 & 79.5 & 77.1 & 78.2 & 77.6 & 71.9  & 70.3  & 72.8  & 71.5 \\
    KO       & 88.2 & 89.3 & 90.2 & 89.7 & 92.3 & 94.6 & 92.8 & 93.7 & 85.3  & 83.2  & 84.9  & 84.0 \\
    OP+KO    & 87.6 & 85.1 & 86.3 & 85.7 & 89.5 & 92.7 & 91.3 & 92.0 & 91.7  & 90.8  & 89.4  & 90.1 \\
    \midrule
    \bottomrule
\end{tabular*}
\label{table:gen-eval}
\end{table*}

\subsection{Statistical analysis}
We statistically analyze H-SemiS's performance to determine its significance relative to a null distribution. The null hypothesis postulates that a random process generates the observed results. We define the significance level as $0.05$ $(\alpha = 0.05)$, the threshold for rejecting the null hypothesis. We then compute \textit{p}-values for all training and testing combinations across the multi-class and binary-class datasets and find that all values are below $10^{-10}$. We reject the null hypothesis since every \textit{p}-value is significantly lower than $\alpha$. We conclude that H-SemiS's performance significantly surpasses that of random chance.

\subsection{Computation Cost Analysis}
\label{sec:computaion}
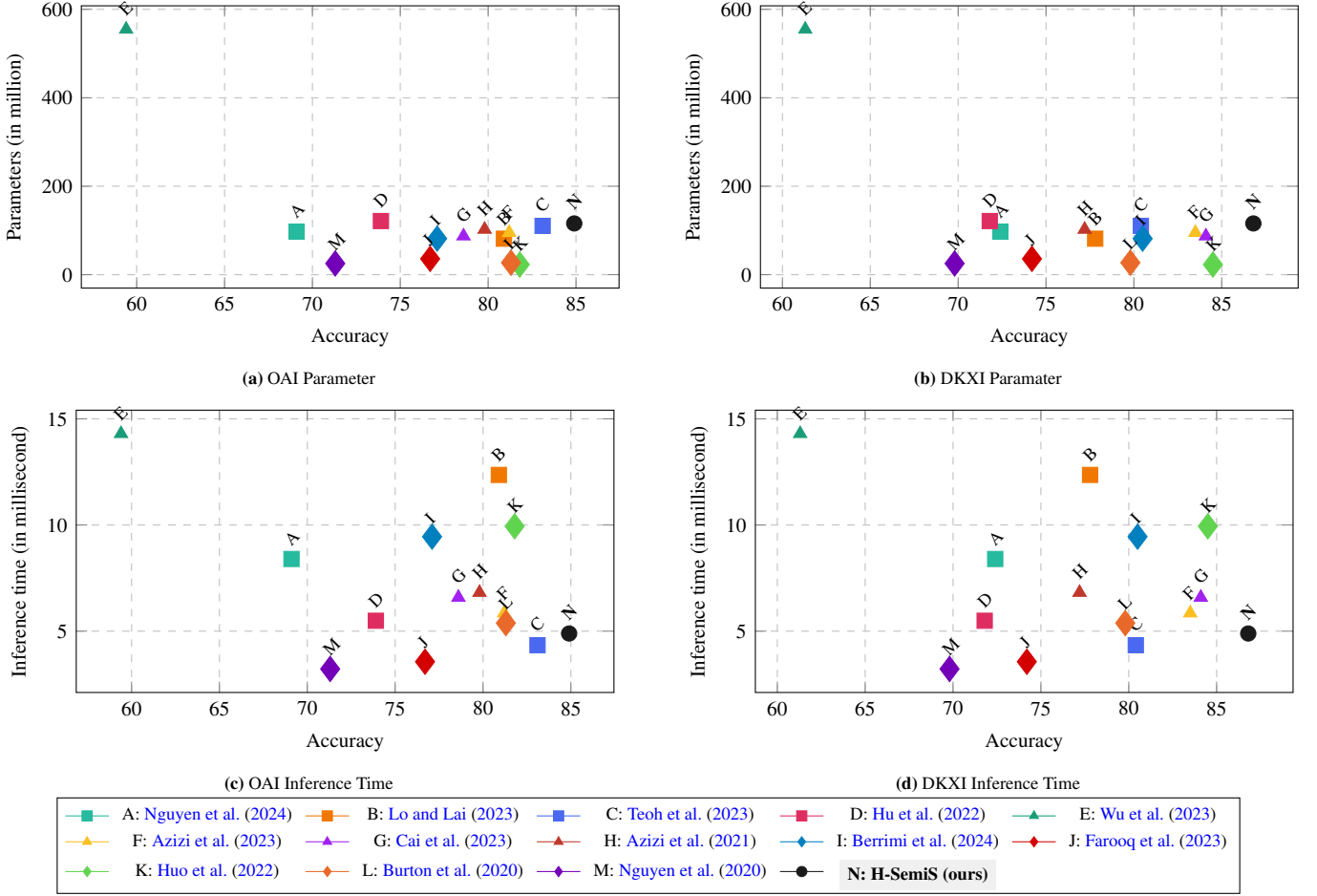
\begin{figure*}[!ht]
\captionsetup[subfigure]{justification=centering}
    \centering
\begin{subfigure}[t]{0.49\linewidth}\centering
    \begin{tikzpicture}
    \begin{axis}[
        width=\textwidth,
        height=5.5cm,
        grid=both,
        grid style=dashed,
        xlabel={Accuracy}, 
        ylabel={Parameters (in million)},
        font=\footnotesize,
        legend style={
            font=\footnotesize,
            at={(0.5,-0.25)},
            anchor=north,
            legend columns=3,
            column sep=1em
        },
        scatter/classes={
            A={mark=square*,l1},
            B={mark=square*,l2}, 
            C={mark=square*,l3}, 
            D={mark=square*,l4}, 
            E={mark=triangle*,l5}, 
            F={mark=triangle*,l6}, 
            G={mark=triangle*,l7}, 
            H={mark=triangle*,l8}, 
            I={mark=diamond*,l9},
            J={mark=diamond*,l10}, 
            K={mark=diamond*,l11}, 
            L={mark=diamond*,l12}, 
            M={mark=diamond*,l13}, 
            N={mark=*, black!90}
        },
    ] 
    \addplot[
        scatter, 
        only marks,
        scatter src=explicit symbolic,
        visualization depends on={value \thisrow{size} \as \perpointmarksize},
        scatter/@pre marker code/.append style={
            /tikz/mark size=\perpointmarksize
        },
        visualization depends on={value \thisrow{annotation} \as \annotvalue},
        nodes near coords*={\annotvalue},
        node near coord style={rotate=45, anchor=south west, font=\scriptsize},
    ]
    table[meta=label] {
        y	x	label	annotation	size
        69.1	97.40	A	A	3
        80.9	81.36	B	B	3
        83.1	110.40	C	C	3
        73.9	121.35	D	D	3
        59.4	554.75	E	E	3
        81.2	95.15	F	F	3
        78.6	87.32	G	G	3
        79.8	102.35	H	H	3
        77.1	81.58	I	I	5
        76.7	35.87	J	J	5
        81.8	22.89	K	K	5
        81.3	27.11	L	L	5
        71.3	25.35	M	M	5
        84.9	115.89	N	\textbf{N}	3
        };
    \end{axis}
    \end{tikzpicture}
    \caption{OAI Parameter}\label{fig:oai-param}
    \end{subfigure}\hfill
\begin{subfigure}[t]{0.49\linewidth}\centering
    \begin{tikzpicture}
    \begin{axis}[
        width=\textwidth,
        height=5.5cm,
        grid=both,
        grid style=dashed,
        xlabel={Accuracy}, 
        ylabel={Parameters (in million)},
        font=\footnotesize,
        legend style={
            font=\footnotesize,
            at={(0.5,-0.25)},
            anchor=north,
            legend columns=3,
            column sep=1em
        },
        scatter/classes={
            A={mark=square*,l1},
            B={mark=square*,l2}, 
            C={mark=square*,l3}, 
            D={mark=square*,l4}, 
            E={mark=triangle*,l5}, 
            F={mark=triangle*,l6}, 
            G={mark=triangle*,l7}, 
            H={mark=triangle*,l8}, 
            I={mark=diamond*,l9},
            J={mark=diamond*,l10}, 
            K={mark=diamond*,l11}, 
            L={mark=diamond*,l12}, 
            M={mark=diamond*,l13}, 
            N={mark=*, black!90}
        },
    ] 
    \addplot[
            scatter, 
            only marks,
            scatter src=explicit symbolic,
            visualization depends on={value \thisrow{size} \as \perpointmarksize},
            scatter/@pre marker code/.append style={
                /tikz/mark size=\perpointmarksize
            },
            visualization depends on={value \thisrow{annotation} \as \annotvalue},
            nodes near coords*={\annotvalue},
            node near coord style={rotate=45, anchor=south west, font=\scriptsize},
        ]
        table[meta=label] {
    	y	x	label	annotation	size
            72.4	97.40	A	A	3
            77.8	81.36	B	B	3
            80.4	110.40	C	C	3
            71.8	121.35	D	D	3
            61.3	554.75	E	E	3
            83.5	95.15	F	F	3
            84.1	87.32	G	G	3
            77.2	102.35	H	H	3
            80.5	81.58	I	I	5
            74.2	35.87	J	J	5
            84.5	22.89	K	K	5
            79.8	27.11	L	L	5
            69.8	25.35	M	M	5
            86.8	115.89	N	\textbf{N}	3
        };
    \end{axis}
    \end{tikzpicture}
    \caption{DKXI Paramater}\label{fig:kxi-param}
    \end{subfigure}\hfill
\\ 
\begin{subfigure}[t]{0.49\linewidth}\centering
\begin{tikzpicture}
    \begin{axis}[
        width=\textwidth,
        height=5.5cm,
        grid=both,
        grid style=dashed,
        xlabel={Accuracy}, 
        ylabel={Inference time (in millisecond)},
        font=\footnotesize,
        legend style={
            font=\footnotesize,
            at={(0.5,-0.25)},
            anchor=north,
            legend columns=3,
            column sep=1em
        },
        scatter/classes={
            A={mark=square*,l1},
            B={mark=square*,l2}, 
            C={mark=square*,l3}, 
            D={mark=square*,l4}, 
            E={mark=triangle*,l5}, 
            F={mark=triangle*,l6}, 
            G={mark=triangle*,l7}, 
            H={mark=triangle*,l8}, 
            I={mark=diamond*,l9},
            J={mark=diamond*,l10}, 
            K={mark=diamond*,l11}, 
            L={mark=diamond*,l12}, 
            M={mark=diamond*,l13}, 
            N={mark=*, black!90}
        },
    ] 
    \addplot[
            scatter, 
            only marks,
            scatter src=explicit symbolic,
            visualization depends on={value \thisrow{size} \as \perpointmarksize},
            scatter/@pre marker code/.append style={
                /tikz/mark size=\perpointmarksize
            },
            visualization depends on={value \thisrow{annotation} \as \annotvalue},
            nodes near coords*={\annotvalue},
            node near coord style={rotate=45, anchor=south west, font=\scriptsize},
        ]
        table[meta=label] {
            y	x	label	annotation	size
            69.1	8.40	A	A	3
            80.9	12.36	B	B	3
            83.1	4.34	C	C	3
            73.9	5.49	D	D	3
            59.4	14.30	E	E	3
            81.2	5.85	F	F	3
            78.6	6.59	G	G	3
            79.8	6.82	H	H	3
            77.1	9.45	I	I	5
            76.7	3.56	J	J	5
            81.8	9.94	K	K	5
            81.3	5.38	L	L	5
            71.3	3.22	M	M	5
            84.9	4.89	N	\textbf{N}	3
        };
    \end{axis}
    \end{tikzpicture}
\caption{OAI Inference Time}\label{fig:oai-inftime}
\end{subfigure}\hfill
\begin{subfigure}[t]{0.49\linewidth}\centering
    \begin{tikzpicture}
    \begin{axis}[
        width=\textwidth,
        height=5.5cm,
        grid=both,
        grid style=dashed,
        xlabel={Accuracy}, 
        ylabel={Inference time (in millisecond)},
        font=\footnotesize,
        legend style={
            font=\footnotesize,
            at={(0.5,-0.25)},
            anchor=north,
            legend columns=3,
            column sep=1em
        },
        scatter/classes={
            A={mark=square*,l1},
            B={mark=square*,l2}, 
            C={mark=square*,l3}, 
            D={mark=square*,l4}, 
            E={mark=triangle*,l5}, 
            F={mark=triangle*,l6}, 
            G={mark=triangle*,l7}, 
            H={mark=triangle*,l8}, 
            I={mark=diamond*,l9},
            J={mark=diamond*,l10}, 
            K={mark=diamond*,l11}, 
            L={mark=diamond*,l12}, 
            M={mark=diamond*,l13}, 
            N={mark=*, black!90}
        },
    ]    
    \addplot[
            scatter, 
            only marks,
            scatter src=explicit symbolic,
            visualization depends on={value \thisrow{size} \as \perpointmarksize},
            scatter/@pre marker code/.append style={
                /tikz/mark size=\perpointmarksize
            },
            visualization depends on={value \thisrow{annotation} \as \annotvalue},
            nodes near coords*={\annotvalue},
            node near coord style={rotate=45, anchor=south west, font=\scriptsize},
        ]
        table[meta=label] {
            y	x	label	annotation	size
            72.4	8.40	A	A	3
            77.8	12.36	B	B	3
            80.4	4.34	C	C	3
            71.8	5.49	D	D	3
            61.3	14.30	E	E	3
            83.5	5.85	F	F	3
            84.1	6.59	G	G	3
            77.2	6.82	H	H	3
            80.5	9.45	I	I	5
            74.2	3.56	J	J	5
            84.5	9.94	K	K	5
            79.8	5.38	L	L	5
            69.8	3.22	M	M	5
            86.8	4.89	N	\textbf{N}	3
        };
    \end{axis}
    \end{tikzpicture}
    \caption{DKXI Inference Time}\label{fig:dkxi-inftime}
    \end{subfigure}\hfill
\\
\begin{tikzpicture}
\begin{axis}[
    hide axis, 
    xmin=0, xmax=1,
    ymin=0, ymax=1,
    legend style={
        font=\scriptsize,
        at={(0.5,-0.25)},
        anchor=north,
        legend columns=5,
        column sep=3pt
    }
]
\addlegendimage{mark=square*, l1}
\addlegendentry{A: \cite{Nguyen2024mat}}
\addlegendimage{mark=square*, l2}
\addlegendentry{B: \cite{Lo2023dlseptic}}
\addlegendimage{mark=square*, l3}
\addlegendentry{C: \cite{Teoh2023dhl}}
\addlegendimage{mark=square*, l4}
\addlegendentry{D: \cite{Hu2022enn}}
\addlegendimage{mark=triangle*, l5}
\addlegendentry{E: \cite{Wu2023}}
\addlegendimage{mark=triangle*, l6}
\addlegendentry{F: \cite{Azizi2023culp}}
\addlegendimage{mark=triangle*, l7}
\addlegendentry{G: \cite{Cai2023}}
\addlegendimage{mark=triangle*, l8}
\addlegendentry{H: \cite{Azizi2021}}
\addlegendimage{mark=diamond*, l9}
\addlegendentry{I: \cite{Berrimi2024mri}}
\addlegendimage{mark=diamond*, l10}
\addlegendentry{J: \cite{Farooq2023dcaae}}
\addlegendimage{mark=diamond*, l11}
\addlegendentry{K: \cite{Huo2022dcmt}}
\addlegendimage{mark=diamond*, l12}
\addlegendentry{L: \cite{Burton2020}}
\addlegendimage{mark=diamond*, l13}
\addlegendentry{M: \cite{Nguyen2020semix}}
\addlegendimage{mark=*, black!90}
\addlegendentry{\colorbox{lightgray}{\textbf{N: H-SemiS (ours)}}}
    \end{axis}
    \end{tikzpicture}
          
\caption{Comparative analysis of model parameters, inference time, and accuracy for the H-SemiS framework against existing baselines on benchmark datasets. The first row illustrates the comparison between accuracy (x-axis) and the number of parameters (y-axis) on the OAI and DKXI datasets. The second row depicts the comparison between accuracy (x-axis) and inference time (y-axis) on the same datasets.}
\label{fig:computation-analysis}
\end{figure*}

We evaluate the computational cost of H-SemiS by comparing trainable parameters and inference time against supervised, self-supervised, and semi-supervised baselines. Among supervised methods, \cite{Teoh2023dhl} reports the lowest inference time at 4.34 4 milliseconds (ms) with 110.40  million (M) parameters, whereas \cite{Hu2022enn} has the largest model size at 121.35M. In the self-supervised setting, \cite{Wu2023} exhibits the highest complexity with 554.75M parameters and a 14.30 ms inference time, while \cite{Azizi2023culp} is more efficient with 95.15M parameters and 5.85 ms inference. Among semi-supervised approaches, \cite{Nguyen2020semix} achieves the fastest inference at 3.22 ms with 25.35M parameters, whereas \cite{Huo2022dcmt} shows higher latency at 9.94 ms despite fewer parameters. In comparison, H-SemiS integrates hierarchical semi-supervised learning with self-supervision, maintaining a parameter count of 115.89M with an inference time of 4.09 ms, which is among the lowest for high-performing models. As summarized in \cref{tab:computation} and visualized in \cref{fig:computation-analysis}, H-SemiS achieves state-of-the-art performance while maintaining a balance between accuracy, model size, and inference efficiency.

\begin{table}[!ht]  \footnotesize
\setlength{\tabcolsep}{2pt}
\centering
\caption{Comparative analysis of parameters and inference times for the H-SemiS framework against existing baselines}
    \begin{threeparttable}    
    \begin{NiceTabular*}{\linewidth}{@{\extracolsep{\fill}}ll cc }
    \toprule
    \midrule
    & \textbf{Methods} & \textbf{Parameters (M)} & \textbf{Inference time (ms)} \\
    
    \midrule
    I
    & \cite{Nguyen2024mat}   & 97.40  & 8.40  \\
    & \cite{Lo2023dlseptic}  & 81.36  & 12.36 \\
    & \cite{Teoh2023dhl}     & 110.40 & 4.34  \\
    & \cite{Hu2022enn}       & 121.35 & 5.49  \\

    \cdashline{1-4}[0.5pt/2pt]\noalign{\vskip 0.5ex}
    II
    & \cite{Wu2023}          & 554.75 & 14.30 \\
    & \cite{Azizi2023culp}   & 95.15  & 5.85  \\
    & \cite{Cai2023}         & 87.32  & 6.59  \\
    & \cite{Azizi2021}       & 102.35 & 6.82  \\

    \cdashline{1-4}[0.5pt/2pt]\noalign{\vskip 0.5ex}
    III
    & \cite{Berrimi2024mri}  & 81.58  & 9.45  \\
    & \cite{Farooq2023dcaae} & 35.87  & 3.56  \\
    & \cite{Huo2022dcmt}     & 22.89  & 9.94  \\
    & \cite{Burton2020}      & 27.11  & 5.38  \\
    & \cite{Nguyen2020semix} & 25.35  & 3.22  \\

    \cdashline{1-4}[0.5pt/2pt]\noalign{\vskip 0.5ex}

    \rowcolor{lightgray}
    \textbf{Ours} & \textbf{H-SemiS}  & \textbf{115.89} & \textbf{4.09} \\
    
    \midrule
    \bottomrule
    \end{NiceTabular*}
    \begin{tablenotes}
    \scriptsize
    \centering
    \item I: Supervised-based Techniques; II: Self-Supervised-based Techniques; 
    \item III: Semi-Supervised-based Techniques; \textbf{M}: million; \textbf{ms}: millisecond
    \end{tablenotes}
    \end{threeparttable}
    
    \label{tab:computation}
\end{table}

\section{Discussion}
\label{sec:discussion}
In this study, we propose a Hierarchical Semi-Supervised Framework with Self-Supervision (H-SemiS) for grading knee osteoarthritis across five severity levels (KL0–KL4). H-SemiS integrates self-supervised and semi-supervised learning to improve severity grading under limited annotation. The framework operates in three stages. First, an adversarial-inspired self-supervised module generates reconstructed samples (\cref{sec:mirec}). Second, these samples receive proxy labels through a similarity-based labeling strategy (\cref{sec:sirl}). Third, the imbalanced multi-class problem is decomposed into a hierarchy of binary classifiers using a rule-based scheme (\cref{sec:him}).

We evaluate H-SemiS on two multi-class datasets (\cref{sec:dataset}) and show consistent improvements over competing baselines (\cref{sec:quantitative}) and state-of-the-art methods (\cref{sec:sota}). To interpret model decisions, we apply XAI techniques, including Grad-CAM, Grad-CAM++, and Score-CAM (\cref{sec:qualitative}). We further conduct comprehensive ablation studies to quantify the contribution of each component (\cref{sec:ablation}) and assess cross-dataset generalizability (\cref{sec:generalizability}). Finally, we analyze the computational characteristics of H-SemiS relative to existing approaches (\cref{sec:computaion}).

\subsection{Feature visualization using t-SNE}
\begin{figure*}[!ht]
\captionsetup[subfigure]{justification=centering}
    \centering
    \begin{subfigure}[t]{0.49\linewidth}\centering
    \begin{tikzpicture}
    \begin{axis}[
        width=\linewidth,
        height=5.8cm,
        grid=both,
        grid style=dashed,
        font=\footnotesize,
        legend style={
            font=\footnotesize,
            at={(0.5,-0.1)}, 
            anchor=north,     
            legend cell align={left},
            legend columns=-1, 
            column sep=2pt,
            draw=black!20,        
            fill=none
        },
        scatter/classes={
            0={mark=o, txt},        
            1={mark=star, red},     
            2={mark=square, blue},  
            3={mark=triangle, orange}, 
            4={mark=oplus, cyan}    
        }
    ]   

    \addplot[
        scatter, 
        only marks,
        scatter src=explicit 
    ]
    table[x index=0, y index=1, meta index=2] {pre.dat};

    \legend{\scriptsize Class 0, Class 1, Class 2, Class 3, Class 4}

    \end{axis}          
    \end{tikzpicture}
         \caption{Without quantum enhancement}
         \label{fig:pre-tsne}
    \end{subfigure}\hfill
    \begin{subfigure}[t]{0.49\linewidth}\centering
    \begin{tikzpicture}
    \begin{axis}[
        width=\linewidth,
        height=5.8cm,
        grid=both,
        grid style=dashed,
        font=\footnotesize,
        legend style={
            font=\footnotesize,
            at={(0.5,-0.1)}, 
            anchor=north, 
            legend cell align={left},
            legend columns=-1, 
            column sep=2pt,
            draw=black!20,    
            fill=none
        },
        scatter/classes={
            0={mark=o, txt},        
            1={mark=star, red},     
            2={mark=square, blue},  
            3={mark=triangle, orange},
            4={mark=oplus, cyan}    
        }
    ]   

    \addplot[
        scatter, 
        only marks,
        scatter src=explicit 
    ]
    table[x index=0, y index=1, meta index=2] {post.dat};

    \legend{\scriptsize Class 0, Class 1, Class 2, Class 3, Class 4}

    \end{axis}          
    \end{tikzpicture}
         \caption{With quantum enhancement}
         \label{fig:post-tsne}
     \end{subfigure}
          
    \caption{The t-SNE visualization comparing features (a) without quantum enhancement and (b) with quantum enhancement. The plot of quantum-enhanced features effectively visualizes the multi-class classification, showing distinct and well-separated clusters for each class.}
    \label{fig:tsne}
\end{figure*}
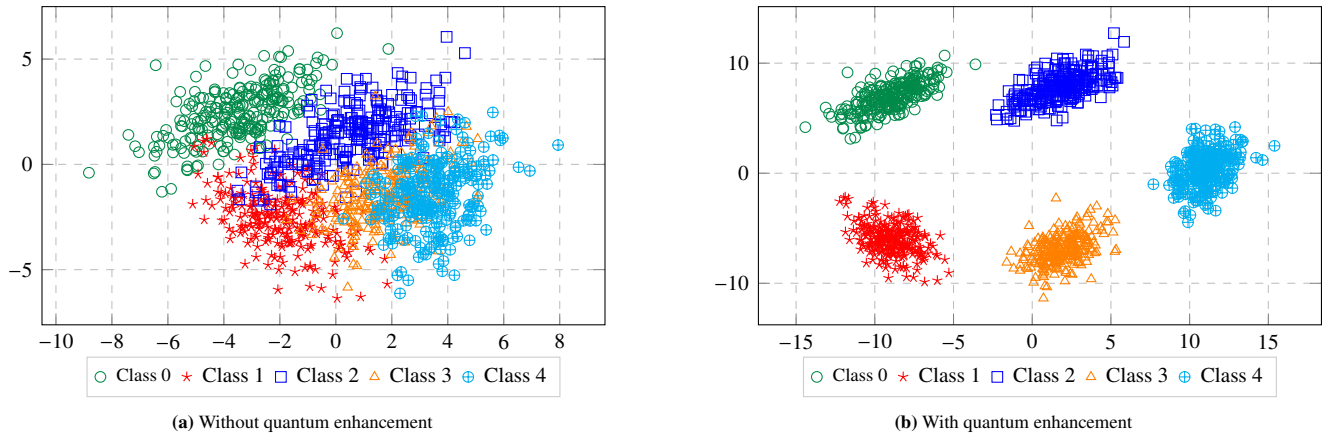

We examine the feature distribution learned by the teacher framework (see \cref{sec:qtest}) using t-distributed Stochastic Neighbor Embedding (t-SNE)~\citep{Vandermaaten08tsne}. \cref{fig:tsne} shows the visualization for multi-class classification. Without quantum enhancement, features from different classes exhibit substantial overlap. With quantum enhancement, the embeddings form compact class-wise clusters with clearer separation across all classes. This behavior indicates that the quantum-enhanced teacher captures class-specific structure more effectively and supports reliable discrimination among multiple KOA severity levels.

\subsection{Limitation and Future Directions}
Although H-SemiS achieves strong performance for KOA severity grading on KXR samples, several directions remain for future work. First, the current study focuses on KXR-based feature learning, and extending the framework to other imaging modalities such as MRI and CT could further enhance diagnostic capability. Second, we use only anteroposterior views; incorporating lateral and oblique views may improve scalability and robustness. Third, reliance on 2D scans limits the capture of internal structural variations, and integrating 3D imaging could provide richer anatomical information and improve grading accuracy. Despite these limitations, H-SemiS delivers reliable KOA severity grading with limited annotations, highlighting its practical potential.

\section{Conclusion}
\label{sec:conclusion}
In this work, we propose H-SemiS, a hierarchical semi-supervised framework with self-supervision for grading knee osteoarthritis (KOA) severity from knee X-ray (KXR) images. The framework addresses data scarcity and multiclass imbalance by combining limited labeled data with reconstructed samples. To handle imbalance, H-SemiS adopts a dual strategy. First, it decomposes the multiclass problem into a hierarchy of binary tasks using a rule-based design to promote balanced learning across severity grades. Second, it introduces an adversarial inspired self-supervised module to generate reconstructed samples, increasing feature diversity and improving data distribution. We further integrate quantum learning within a teacher student framework to capture complex nonlinear anatomical relationships and enhance fine-grained feature discrimination. Extensive experiments on multiclass imbalanced datasets demonstrate that H-SemiS consistently outperforms competing methods across key metrics. Ablation studies confirm the contribution of each component. These results establish H-SemiS as a reliable and effective approach for grading KOA severity, particularly in resource-constrained clinical settings.

\footnotesize
\setlength{\bibsep}{1pt}
\bibliographystyle{elsarticle-harv} 
\bibliography{refs}

\begin{thebibliography}{55}
\expandafter\ifx\csname natexlab\endcsname\relax\def\natexlab#1{#1}\fi
\providecommand{\url}[1]{\texttt{#1}}
\providecommand{\href}[2]{#2}
\providecommand{\path}[1]{#1}
\providecommand{\DOIprefix}{doi:}
\providecommand{\ArXivprefix}{arXiv:}
\providecommand{\URLprefix}{URL: }
\providecommand{\Pubmedprefix}{pmid:}
\providecommand{\doi}[1]{\href{http://dx.doi.org/#1}{\path{#1}}}
\providecommand{\Pubmed}[1]{\href{pmid:#1}{\path{#1}}}
\providecommand{\bibinfo}[2]{#2}
\ifx\xfnm\relax \def\xfnm[#1]{\unskip,\space#1}\fi
\bibitem[{Azizi et~al.(2023)Azizi, Culp, Freyberg, Mustafa, Baur, Kornblith, Chen, Tomasev, Mitrović, Strachan, Mahdavi, Wulczyn, Babenko, Walker, Loh, Chen, Liu, Bavishi, McKinney, Winkens, Roy, Beaver, Ryan, Krogue, Etemadi, Telang, Liu, Peng, Corrado, Webster, Fleet, Hinton, Houlsby, Karthikesalingam, Norouzi and Natarajan}]{Azizi2023culp}
\bibinfo{author}{Azizi, S.}, \bibinfo{author}{Culp, L.}, \bibinfo{author}{Freyberg, J.}, \bibinfo{author}{Mustafa, B.}, \bibinfo{author}{Baur, S.}, \bibinfo{author}{Kornblith, S.}, \bibinfo{author}{Chen, T.}, \bibinfo{author}{Tomasev, N.}, \bibinfo{author}{Mitrović, J.}, \bibinfo{author}{Strachan, P.}, \bibinfo{author}{Mahdavi, S.S.}, \bibinfo{author}{Wulczyn, E.}, \bibinfo{author}{Babenko, B.}, \bibinfo{author}{Walker, M.}, \bibinfo{author}{Loh, A.}, \bibinfo{author}{Chen, P.H.C.}, \bibinfo{author}{Liu, Y.}, \bibinfo{author}{Bavishi, P.}, \bibinfo{author}{McKinney, S.M.}, \bibinfo{author}{Winkens, J.}, \bibinfo{author}{Roy, A.G.}, \bibinfo{author}{Beaver, Z.}, \bibinfo{author}{Ryan, F.}, \bibinfo{author}{Krogue, J.}, \bibinfo{author}{Etemadi, M.}, \bibinfo{author}{Telang, U.}, \bibinfo{author}{Liu, Y.}, \bibinfo{author}{Peng, L.}, \bibinfo{author}{Corrado, G.S.}, \bibinfo{author}{Webster, D.R.}, \bibinfo{author}{Fleet, D.}, \bibinfo{author}{Hinton, G.}, \bibinfo{author}{Houlsby, N.},
  \bibinfo{author}{Karthikesalingam, A.}, \bibinfo{author}{Norouzi, M.}, \bibinfo{author}{Natarajan, V.}, \bibinfo{year}{2023}.
\newblock \bibinfo{title}{Robust and data-efficient generalization of self-supervised machine learning for diagnostic imaging}.
\newblock \bibinfo{journal}{Nature Biomedical Engineering} \bibinfo{volume}{7}, \bibinfo{pages}{756–779}.
\newblock \DOIprefix\doi{10.1038/s41551-023-01049-7}.
\bibitem[{Azizi et~al.(2021)Azizi, Mustafa, Ryan, Beaver, Freyberg, Deaton, Loh, Karthikesalingam, Kornblith, Chen, Natarajan and Norouzi}]{Azizi2021}
\bibinfo{author}{Azizi, S.}, \bibinfo{author}{Mustafa, B.}, \bibinfo{author}{Ryan, F.}, \bibinfo{author}{Beaver, Z.}, \bibinfo{author}{Freyberg, J.}, \bibinfo{author}{Deaton, J.}, \bibinfo{author}{Loh, A.}, \bibinfo{author}{Karthikesalingam, A.}, \bibinfo{author}{Kornblith, S.}, \bibinfo{author}{Chen, T.}, \bibinfo{author}{Natarajan, V.}, \bibinfo{author}{Norouzi, M.}, \bibinfo{year}{2021}.
\newblock \bibinfo{title}{Big self-supervised models advance medical image classification}, in: \bibinfo{booktitle}{2021 IEEE/CVF International Conference on Computer Vision (ICCV)}, \bibinfo{publisher}{IEEE}. p. \bibinfo{pages}{3458–3468}.
\newblock \DOIprefix\doi{10.1109/iccv48922.2021.00346}.
\bibitem[{Benedetti et~al.(2019)Benedetti, Lloyd, Sack and Fiorentini}]{Benedetti2019paramqml}
\bibinfo{author}{Benedetti, M.}, \bibinfo{author}{Lloyd, E.}, \bibinfo{author}{Sack, S.}, \bibinfo{author}{Fiorentini, M.}, \bibinfo{year}{2019}.
\newblock \bibinfo{title}{Parameterized quantum circuits as machine learning models}.
\newblock \bibinfo{journal}{Quantum Science and Technology} \bibinfo{volume}{4}, \bibinfo{pages}{043001}.
\newblock \DOIprefix\doi{10.1088/2058-9565/ab4eb5}.
\bibitem[{Berrimi et~al.(2024)Berrimi, Hans and Jennane}]{Berrimi2024mri}
\bibinfo{author}{Berrimi, M.}, \bibinfo{author}{Hans, D.}, \bibinfo{author}{Jennane, R.}, \bibinfo{year}{2024}.
\newblock \bibinfo{title}{A semi-supervised multiview-mri network for the detection of knee osteoarthritis}.
\newblock \bibinfo{journal}{Computerized Medical Imaging and Graphics} \bibinfo{volume}{114}, \bibinfo{pages}{102371}.
\newblock \DOIprefix\doi{10.1016/j.compmedimag.2024.102371}.
\bibitem[{Burton et~al.(2020)Burton, Myers and Rullkoetter}]{Burton2020}
\bibinfo{author}{Burton, W.}, \bibinfo{author}{Myers, C.}, \bibinfo{author}{Rullkoetter, P.}, \bibinfo{year}{2020}.
\newblock \bibinfo{title}{Semi-supervised learning for automatic segmentation of the knee from mri with convolutional neural networks}.
\newblock \bibinfo{journal}{Computer Methods and Programs in Biomedicine} \bibinfo{volume}{189}, \bibinfo{pages}{105328}.
\newblock \DOIprefix\doi{10.1016/j.cmpb.2020.105328}.
\bibitem[{Cai et~al.(2023)Cai, Chen, Yang, Zhou and Cheng}]{Cai2023}
\bibinfo{author}{Cai, Y.}, \bibinfo{author}{Chen, H.}, \bibinfo{author}{Yang, X.}, \bibinfo{author}{Zhou, Y.}, \bibinfo{author}{Cheng, K.T.}, \bibinfo{year}{2023}.
\newblock \bibinfo{title}{Dual-distribution discrepancy with self-supervised refinement for anomaly detection in medical images}.
\newblock \bibinfo{journal}{Medical Image Analysis} \bibinfo{volume}{86}, \bibinfo{pages}{102794}.
\newblock \DOIprefix\doi{10.1016/j.media.2023.102794}.
\bibitem[{Chattopadhay et~al.(2018)Chattopadhay, Sarkar, Howlader and Balasubramanian}]{Chattopadhay2018gradcam++}
\bibinfo{author}{Chattopadhay, A.}, \bibinfo{author}{Sarkar, A.}, \bibinfo{author}{Howlader, P.}, \bibinfo{author}{Balasubramanian, V.N.}, \bibinfo{year}{2018}.
\newblock \bibinfo{title}{Grad-cam++: Generalized gradient-based visual explanations for deep convolutional networks}, in: \bibinfo{booktitle}{2018 IEEE Winter Conference on Applications of Computer Vision (WACV)}, \bibinfo{publisher}{IEEE}. pp. \bibinfo{pages}{839--847}.
\newblock \DOIprefix\doi{10.1109/wacv.2018.00097}.
\bibitem[{Chen(2018)}]{ChenOAI}
\bibinfo{author}{Chen, P.}, \bibinfo{year}{2018}.
\newblock \bibinfo{title}{Knee osteoarthritis severity grading dataset}.
\newblock \DOIprefix\doi{0.17632/56rmx5bjcr.1}.
\bibitem[{Chen et~al.(2020)Chen, Kornblith, Norouzi and Hinton}]{Chen2020SimCLR}
\bibinfo{author}{Chen, T.}, \bibinfo{author}{Kornblith, S.}, \bibinfo{author}{Norouzi, M.}, \bibinfo{author}{Hinton, G.}, \bibinfo{year}{2020}.
\newblock \bibinfo{title}{A simple framework for contrastive learning of visual representations}, in: \bibinfo{booktitle}{37th International Conference on Machine Learning, ICML 2020}, pp. \bibinfo{pages}{1597--1607}.
\newblock \URLprefix \url{https://proceedings.mlr.press/v119/chen20j/chen20j.pdf}. \bibinfo{note}{accessed 8 March 2025}.
\bibitem[{Cubuk et~al.(2020)Cubuk, Zoph, Shlens and Le}]{Cubuk2020randaug}
\bibinfo{author}{Cubuk, E.D.}, \bibinfo{author}{Zoph, B.}, \bibinfo{author}{Shlens, J.}, \bibinfo{author}{Le, Q.V.}, \bibinfo{year}{2020}.
\newblock \bibinfo{title}{Randaugment: Practical automated data augmentation with a reduced search space}, in: \bibinfo{booktitle}{2020 IEEE/CVF Conference on Computer Vision and Pattern Recognition Workshops (CVPRW)}, \bibinfo{publisher}{IEEE}. pp. \bibinfo{pages}{702--703}.
\newblock \DOIprefix\doi{10.1109/cvprw50498.2020.00359}.
\bibitem[{Dosovitskiy et~al.(2021)Dosovitskiy, Beyer, Kolesnikov, Weissenborn, Zhai, Unterthiner, Dehghani, Minderer, Heigold, Gelly, Uszkoreit and Houlsby}]{Dosovitskiy2021vit}
\bibinfo{author}{Dosovitskiy, A.}, \bibinfo{author}{Beyer, L.}, \bibinfo{author}{Kolesnikov, A.}, \bibinfo{author}{Weissenborn, D.}, \bibinfo{author}{Zhai, X.}, \bibinfo{author}{Unterthiner, T.}, \bibinfo{author}{Dehghani, M.}, \bibinfo{author}{Minderer, M.}, \bibinfo{author}{Heigold, G.}, \bibinfo{author}{Gelly, S.}, \bibinfo{author}{Uszkoreit, J.}, \bibinfo{author}{Houlsby, N.}, \bibinfo{year}{2021}.
\newblock \bibinfo{title}{An image is worth 16x16 words: Transformers for image recognition at scale}, in: \bibinfo{booktitle}{ICLR 2021 - 9th International Conference on Learning Representations}, pp. \bibinfo{pages}{1--22}.
\newblock \URLprefix \url{https://openreview.net/forum?id=YicbFdNTTy}. \bibinfo{note}{accessed 8 March 2025}.
\bibitem[{Du et~al.(2018)Du, Almajalid, Shan and Zhang}]{Du2018ml}
\bibinfo{author}{Du, Y.}, \bibinfo{author}{Almajalid, R.}, \bibinfo{author}{Shan, J.}, \bibinfo{author}{Zhang, M.}, \bibinfo{year}{2018}.
\newblock \bibinfo{title}{A novel method to predict knee osteoarthritis progression on mri using machine learning methods}.
\newblock \bibinfo{journal}{IEEE Transactions on NanoBioscience} \bibinfo{volume}{17}, \bibinfo{pages}{228–236}.
\newblock \DOIprefix\doi{10.1109/tnb.2018.2840082}.
\bibitem[{Farooq et~al.(2023)Farooq, Ullah, Khan and Gwak}]{Farooq2023dcaae}
\bibinfo{author}{Farooq, M.U.}, \bibinfo{author}{Ullah, Z.}, \bibinfo{author}{Khan, A.}, \bibinfo{author}{Gwak, J.}, \bibinfo{year}{2023}.
\newblock \bibinfo{title}{Dc-aae: Dual channel adversarial autoencoder with multitask learning for kl-grade classification in knee radiographs}.
\newblock \bibinfo{journal}{Computers in Biology and Medicine} \bibinfo{volume}{167}, \bibinfo{pages}{107570}.
\newblock \DOIprefix\doi{10.1016/j.compbiomed.2023.107570}.
\bibitem[{Fei et~al.(2023)Fei, Fan, Zhu, Huang, Wei and Wei}]{Fei2023maegan}
\bibinfo{author}{Fei, Z.}, \bibinfo{author}{Fan, M.}, \bibinfo{author}{Zhu, L.}, \bibinfo{author}{Huang, J.}, \bibinfo{author}{Wei, X.}, \bibinfo{author}{Wei, X.}, \bibinfo{year}{2023}.
\newblock \bibinfo{title}{Masked auto-encoders meet generative adversarial networks and beyond}, in: \bibinfo{booktitle}{2023 IEEE/CVF Conference on Computer Vision and Pattern Recognition (CVPR)}, \bibinfo{publisher}{IEEE}. p. \bibinfo{pages}{24449–24459}.
\newblock \DOIprefix\doi{10.1109/cvpr52729.2023.02342}.
\bibitem[{Ghosh et~al.(2024)Ghosh, Kumar, Kumar and Verma}]{Ghosh2024cregssl}
\bibinfo{author}{Ghosh, S.}, \bibinfo{author}{Kumar, S.}, \bibinfo{author}{Kumar, A.}, \bibinfo{author}{Verma, J.}, \bibinfo{year}{2024}.
\newblock \bibinfo{title}{A closer look at consistency regularization for semi-supervised learning}, in: \bibinfo{booktitle}{Proceedings of the 7th Joint International Conference on Data Science; Management of Data (11th ACM IKDD CODS and 29th COMAD)}, \bibinfo{publisher}{ACM}. p. \bibinfo{pages}{10–17}.
\newblock \DOIprefix\doi{10.1145/3632410.3632453}.
\bibitem[{Goodfellow et~al.(2020)Goodfellow, Pouget-Abadie, Mirza, Xu, Warde-Farley, Ozair, Courville and Bengio}]{Goodfellow2020gan}
\bibinfo{author}{Goodfellow, I.}, \bibinfo{author}{Pouget-Abadie, J.}, \bibinfo{author}{Mirza, M.}, \bibinfo{author}{Xu, B.}, \bibinfo{author}{Warde-Farley, D.}, \bibinfo{author}{Ozair, S.}, \bibinfo{author}{Courville, A.}, \bibinfo{author}{Bengio, Y.}, \bibinfo{year}{2020}.
\newblock \bibinfo{title}{Generative adversarial networks}.
\newblock \bibinfo{journal}{Communications of the ACM} \bibinfo{volume}{63}, \bibinfo{pages}{139–144}.
\newblock \DOIprefix\doi{10.1145/3422622}.
\bibitem[{Gornale and Patravali(2020)}]{GornaleDKXI}
\bibinfo{author}{Gornale, S.}, \bibinfo{author}{Patravali, P.}, \bibinfo{year}{2020}.
\newblock \bibinfo{title}{Digital knee x-ray images}.
\newblock \DOIprefix\doi{10.17632/T9NDX37V5H.1}.
\bibitem[{Guo et~al.(2022)Guo, Han, Wu, Tang, Chen, Wang and Xu}]{Guo2022cmt}
\bibinfo{author}{Guo, J.}, \bibinfo{author}{Han, K.}, \bibinfo{author}{Wu, H.}, \bibinfo{author}{Tang, Y.}, \bibinfo{author}{Chen, X.}, \bibinfo{author}{Wang, Y.}, \bibinfo{author}{Xu, C.}, \bibinfo{year}{2022}.
\newblock \bibinfo{title}{Cmt: Convolutional neural networks meet vision transformers}, in: \bibinfo{booktitle}{2022 IEEE/CVF Conference on Computer Vision and Pattern Recognition (CVPR)}, \bibinfo{publisher}{IEEE}. p. \bibinfo{pages}{12165–12175}.
\newblock \DOIprefix\doi{10.1109/cvpr52688.2022.01186}.
\bibitem[{He et~al.(2022)He, Chen, Xie, Li, Dollar and Girshick}]{He2022mae}
\bibinfo{author}{He, K.}, \bibinfo{author}{Chen, X.}, \bibinfo{author}{Xie, S.}, \bibinfo{author}{Li, Y.}, \bibinfo{author}{Dollar, P.}, \bibinfo{author}{Girshick, R.}, \bibinfo{year}{2022}.
\newblock \bibinfo{title}{Masked autoencoders are scalable vision learners}, in: \bibinfo{booktitle}{2022 IEEE/CVF Conference on Computer Vision and Pattern Recognition (CVPR)}, \bibinfo{publisher}{IEEE}. pp. \bibinfo{pages}{16000--16009}.
\newblock \DOIprefix\doi{10.1109/cvpr52688.2022.01553}.
\bibitem[{He et~al.(2016)He, Zhang, Ren and Sun}]{He_2016ResNet}
\bibinfo{author}{He, K.}, \bibinfo{author}{Zhang, X.}, \bibinfo{author}{Ren, S.}, \bibinfo{author}{Sun, J.}, \bibinfo{year}{2016}.
\newblock \bibinfo{title}{Deep residual learning for image recognition}, in: \bibinfo{booktitle}{2016 IEEE Conference on Computer Vision and Pattern Recognition (CVPR)}, \bibinfo{publisher}{IEEE}. pp. \bibinfo{pages}{770--778}.
\newblock \DOIprefix\doi{10.1109/cvpr.2016.90}.
\bibitem[{Hu et~al.(2022)Hu, Wu, Li, Simic, Zomaya and Wang}]{Hu2022enn}
\bibinfo{author}{Hu, K.}, \bibinfo{author}{Wu, W.}, \bibinfo{author}{Li, W.}, \bibinfo{author}{Simic, M.}, \bibinfo{author}{Zomaya, A.}, \bibinfo{author}{Wang, Z.}, \bibinfo{year}{2022}.
\newblock \bibinfo{title}{Adversarial evolving neural network for longitudinal knee osteoarthritis prediction}.
\newblock \bibinfo{journal}{IEEE Transactions on Medical Imaging} \bibinfo{volume}{41}, \bibinfo{pages}{3207–3217}.
\newblock \DOIprefix\doi{10.1109/tmi.2022.3181060}.
\bibitem[{Huang et~al.(2017)Huang, Liu, Van Der~Maaten and Weinberger}]{Huang_2017DenseNet}
\bibinfo{author}{Huang, G.}, \bibinfo{author}{Liu, Z.}, \bibinfo{author}{Van Der~Maaten, L.}, \bibinfo{author}{Weinberger, K.Q.}, \bibinfo{year}{2017}.
\newblock \bibinfo{title}{Densely connected convolutional networks}, in: \bibinfo{booktitle}{2017 IEEE Conference on Computer Vision and Pattern Recognition (CVPR)}, \bibinfo{publisher}{IEEE}. pp. \bibinfo{pages}{4700--4708}.
\newblock \DOIprefix\doi{10.1109/cvpr.2017.243}.
\bibitem[{Huang et~al.(2023)Huang, Pareek, Jensen, Lungren, Yeung and Chaudhari}]{Huang2023}
\bibinfo{author}{Huang, S.C.}, \bibinfo{author}{Pareek, A.}, \bibinfo{author}{Jensen, M.}, \bibinfo{author}{Lungren, M.P.}, \bibinfo{author}{Yeung, S.}, \bibinfo{author}{Chaudhari, A.S.}, \bibinfo{year}{2023}.
\newblock \bibinfo{title}{Self-supervised learning for medical image classification: a systematic review and implementation guidelines}.
\newblock \bibinfo{journal}{npj Digital Medicine} \bibinfo{volume}{6}.
\newblock \DOIprefix\doi{10.1038/s41746-023-00811-0}.
\bibitem[{Huang et~al.(2024)Huang, Jin, Lu, Hou, Cheng, Fu, Shen and Feng}]{Huang2024cmae}
\bibinfo{author}{Huang, Z.}, \bibinfo{author}{Jin, X.}, \bibinfo{author}{Lu, C.}, \bibinfo{author}{Hou, Q.}, \bibinfo{author}{Cheng, M.M.}, \bibinfo{author}{Fu, D.}, \bibinfo{author}{Shen, X.}, \bibinfo{author}{Feng, J.}, \bibinfo{year}{2024}.
\newblock \bibinfo{title}{Contrastive masked autoencoders are stronger vision learners}.
\newblock \bibinfo{journal}{IEEE Transactions on Pattern Analysis and Machine Intelligence} \bibinfo{volume}{46}, \bibinfo{pages}{2506–2517}.
\newblock \DOIprefix\doi{10.1109/tpami.2023.3336525}.
\bibitem[{Huo et~al.(2022)Huo, Ouyang, Si, Xuan, Wang, Yao, Liu, Xu, Qian, Xue, Wang, Shen and Zhang}]{Huo2022dcmt}
\bibinfo{author}{Huo, J.}, \bibinfo{author}{Ouyang, X.}, \bibinfo{author}{Si, L.}, \bibinfo{author}{Xuan, K.}, \bibinfo{author}{Wang, S.}, \bibinfo{author}{Yao, W.}, \bibinfo{author}{Liu, Y.}, \bibinfo{author}{Xu, J.}, \bibinfo{author}{Qian, D.}, \bibinfo{author}{Xue, Z.}, \bibinfo{author}{Wang, Q.}, \bibinfo{author}{Shen, D.}, \bibinfo{author}{Zhang, L.}, \bibinfo{year}{2022}.
\newblock \bibinfo{title}{Automatic grading assessments for knee mri cartilage defects via self-ensembling semi-supervised learning with dual-consistency}.
\newblock \bibinfo{journal}{Medical Image Analysis} \bibinfo{volume}{80}, \bibinfo{pages}{102508}.
\newblock \DOIprefix\doi{10.1016/j.media.2022.102508}.
\bibitem[{Kellgren and Lawrence(1957)}]{Kellgren1957}
\bibinfo{author}{Kellgren, J.}, \bibinfo{author}{Lawrence, J.}, \bibinfo{year}{1957}.
\newblock \bibinfo{title}{Radiological assessment of osteo-arthrosis}.
\newblock \bibinfo{journal}{Annals of the Rheumatic Diseases} \bibinfo{volume}{16}, \bibinfo{pages}{494–502}.
\newblock \DOIprefix\doi{10.1136/ard.16.4.494}.
\bibitem[{Laine and Aila(2017)}]{Laine2017pi}
\bibinfo{author}{Laine, S.}, \bibinfo{author}{Aila, T.}, \bibinfo{year}{2017}.
\newblock \bibinfo{title}{Temporal ensembling for semi-supervised learning}, in: \bibinfo{booktitle}{5th International Conference on Learning Representations, ICLR 2017 - Conference Track Proceedings}, pp. \bibinfo{pages}{1--17}.
\newblock \URLprefix \url{https://openreview.net/forum?id=BJ6oOfqge}. \bibinfo{note}{accessed 8 March 2025}.
\bibitem[{Lindner et~al.(2013)Lindner, Thiagarajah, Wilkinson, Consortium, Wallis and Cootes}]{Lindner2013bonefinder}
\bibinfo{author}{Lindner, C.}, \bibinfo{author}{Thiagarajah, S.}, \bibinfo{author}{Wilkinson, J.}, \bibinfo{author}{Consortium, T.}, \bibinfo{author}{Wallis, G.}, \bibinfo{author}{Cootes, T.}, \bibinfo{year}{2013}.
\newblock \bibinfo{title}{Fully automatic segmentation of the proximal femur using random forest regression voting}.
\newblock \bibinfo{journal}{IEEE Transactions on Medical Imaging} \bibinfo{volume}{32}, \bibinfo{pages}{1462–1472}.
\newblock \DOIprefix\doi{10.1109/tmi.2013.2258030}.
\bibitem[{Lo and Lai(2023)}]{Lo2023dlseptic}
\bibinfo{author}{Lo, C.M.}, \bibinfo{author}{Lai, K.L.}, \bibinfo{year}{2023}.
\newblock \bibinfo{title}{Deep learning-based assessment of knee septic arthritis using transformer features in sonographic modalities}.
\newblock \bibinfo{journal}{Computer Methods and Programs in Biomedicine} \DOIprefix\doi{10.1016/j.cmpb.2023.107575}.
\bibitem[{van~der Maaten and Hinton(2008)}]{Vandermaaten08tsne}
\bibinfo{author}{van~der Maaten, L.}, \bibinfo{author}{Hinton, G.}, \bibinfo{year}{2008}.
\newblock \bibinfo{title}{Visualizing data using t-sne}.
\newblock \bibinfo{journal}{Journal of Machine Learning Research} \bibinfo{volume}{9}.
\newblock \URLprefix \url{https://jmlr.org/papers/volume9/vandermaaten08a/vandermaaten08a.pdf}. \bibinfo{note}{accessed 8 March 2025}.
\bibitem[{Manzari et~al.(2023)Manzari, Ahmadabadi, Kashiani, Shokouhi and Ayatollahi}]{Manzari_2023MedVit}
\bibinfo{author}{Manzari, O.N.}, \bibinfo{author}{Ahmadabadi, H.}, \bibinfo{author}{Kashiani, H.}, \bibinfo{author}{Shokouhi, S.B.}, \bibinfo{author}{Ayatollahi, A.}, \bibinfo{year}{2023}.
\newblock \bibinfo{title}{Medvit: A robust vision transformer for generalized medical image classification}.
\newblock \bibinfo{journal}{Computers in Biology and Medicine} \bibinfo{volume}{157}, \bibinfo{pages}{106791}.
\newblock \DOIprefix\doi{10.1016/j.compbiomed.2023.106791}.
\bibitem[{Mari et~al.(2020)Mari, Bromley, Izaac, Schuld and Killoran}]{Mari2020qnntransf}
\bibinfo{author}{Mari, A.}, \bibinfo{author}{Bromley, T.R.}, \bibinfo{author}{Izaac, J.}, \bibinfo{author}{Schuld, M.}, \bibinfo{author}{Killoran, N.}, \bibinfo{year}{2020}.
\newblock \bibinfo{title}{Transfer learning in hybrid classical-quantum neural networks}.
\newblock \bibinfo{journal}{Quantum} \bibinfo{volume}{4}, \bibinfo{pages}{340}.
\newblock \DOIprefix\doi{10.22331/q-2020-10-09-340}.
\bibitem[{Nguyen et~al.(2024)Nguyen, Blaschko, Saarakkala and Tiulpin}]{Nguyen2024mat}
\bibinfo{author}{Nguyen, H.H.}, \bibinfo{author}{Blaschko, M.B.}, \bibinfo{author}{Saarakkala, S.}, \bibinfo{author}{Tiulpin, A.}, \bibinfo{year}{2024}.
\newblock \bibinfo{title}{Clinically-inspired multi-agent transformers for disease trajectory forecasting from multimodal data}.
\newblock \bibinfo{journal}{IEEE Transactions on Medical Imaging} \bibinfo{volume}{43}, \bibinfo{pages}{529–541}.
\newblock \DOIprefix\doi{10.1109/tmi.2023.3312524}.
\bibitem[{Nguyen et~al.(2020)Nguyen, Saarakkala, Blaschko and Tiulpin}]{Nguyen2020semix}
\bibinfo{author}{Nguyen, H.H.}, \bibinfo{author}{Saarakkala, S.}, \bibinfo{author}{Blaschko, M.B.}, \bibinfo{author}{Tiulpin, A.}, \bibinfo{year}{2020}.
\newblock \bibinfo{title}{Semixup: In- and out-of-manifold regularization for deep semi-supervised knee osteoarthritis severity grading from plain radiographs}.
\newblock \bibinfo{journal}{IEEE Transactions on Medical Imaging} \bibinfo{volume}{39}, \bibinfo{pages}{4346–4356}.
\newblock \DOIprefix\doi{10.1109/tmi.2020.3017007}.
\bibitem[{Ning et~al.(2025)Ning, Jiang and Zhang}]{Ning2024combat}
\bibinfo{author}{Ning, Z.}, \bibinfo{author}{Jiang, Z.}, \bibinfo{author}{Zhang, D.}, \bibinfo{year}{2025}.
\newblock \bibinfo{title}{To combat multiclass imbalanced problems by aggregating evolutionary hierarchical classifiers}.
\newblock \bibinfo{journal}{IEEE Transactions on Neural Networks and Learning Systems} \bibinfo{volume}{36}, \bibinfo{pages}{5258–5272}.
\newblock \DOIprefix\doi{10.1109/tnnls.2024.3383672}.
\bibitem[{Ovalle-Magallanes et~al.(2022)Ovalle-Magallanes, Avina-Cervantes, Cruz-Aceves and Ruiz-Pinales}]{Magallanes2022hycqcnn}
\bibinfo{author}{Ovalle-Magallanes, E.}, \bibinfo{author}{Avina-Cervantes, J.G.}, \bibinfo{author}{Cruz-Aceves, I.}, \bibinfo{author}{Ruiz-Pinales, J.}, \bibinfo{year}{2022}.
\newblock \bibinfo{title}{Hybrid classical–quantum convolutional neural network for stenosis detection in x-ray coronary angiography}.
\newblock \bibinfo{journal}{Expert Systems with Applications} \bibinfo{volume}{189}, \bibinfo{pages}{116112}.
\newblock \DOIprefix\doi{10.1016/j.eswa.2021.116112}.
\bibitem[{Pan et~al.(2024)Pan, Wu, Tang, Sun, Li, Sun, Liu, Tian and Shen}]{Pan2024autok}
\bibinfo{author}{Pan, J.}, \bibinfo{author}{Wu, Y.}, \bibinfo{author}{Tang, Z.}, \bibinfo{author}{Sun, K.}, \bibinfo{author}{Li, M.}, \bibinfo{author}{Sun, J.}, \bibinfo{author}{Liu, J.}, \bibinfo{author}{Tian, J.}, \bibinfo{author}{Shen, B.}, \bibinfo{year}{2024}.
\newblock \bibinfo{title}{Automatic knee osteoarthritis severity grading based on x-ray images using a hierarchical classification method}.
\newblock \bibinfo{journal}{Arthritis Research \& Therapy} \bibinfo{volume}{26}.
\newblock \DOIprefix\doi{10.1186/s13075-024-03416-4}.
\bibitem[{Poppi et~al.(2021)Poppi, Cornia, Baraldi and Cucchiara}]{Poppi2021adcc}
\bibinfo{author}{Poppi, S.}, \bibinfo{author}{Cornia, M.}, \bibinfo{author}{Baraldi, L.}, \bibinfo{author}{Cucchiara, R.}, \bibinfo{year}{2021}.
\newblock \bibinfo{title}{Revisiting the evaluation of class activation mapping for explainability: A novel metric and experimental analysis}, in: \bibinfo{booktitle}{2021 IEEE/CVF Conference on Computer Vision and Pattern Recognition Workshops (CVPRW)}, \bibinfo{publisher}{IEEE}. pp. \bibinfo{pages}{2299--2304}.
\newblock \DOIprefix\doi{10.1109/cvprw53098.2021.00260}.
\bibitem[{Prabhakar et~al.(2024)Prabhakar, Li, Yang, Shit, Wiestler and Menze}]{Prabhakar2024vitae}
\bibinfo{author}{Prabhakar, C.}, \bibinfo{author}{Li, H.}, \bibinfo{author}{Yang, J.}, \bibinfo{author}{Shit, S.}, \bibinfo{author}{Wiestler, B.}, \bibinfo{author}{Menze, B.}, \bibinfo{year}{2024}.
\newblock \bibinfo{title}{Vit-ae++: improving vision transformer autoencoder for self-supervised medical image representations}, in: \bibinfo{booktitle}{Medical Imaging with Deep Learning}, \bibinfo{organization}{PMLR}. pp. \bibinfo{pages}{666--679}.
\newblock \URLprefix \url{https://proceedings.mlr.press/v227/prabhakar24b/prabhakar24b.pdf}. \bibinfo{note}{accessed 8 March 2025}.
\bibitem[{Raghaw et~al.(2024)Raghaw, Bhore, Rehman and Kumar}]{Raghaw2024xccnet}
\bibinfo{author}{Raghaw, C.S.}, \bibinfo{author}{Bhore, P.S.}, \bibinfo{author}{Rehman, M.Z.U.}, \bibinfo{author}{Kumar, N.}, \bibinfo{year}{2024}.
\newblock \bibinfo{title}{An explainable contrastive-based dilated convolutional network with transformer for pediatric pneumonia detection}.
\newblock \bibinfo{journal}{Applied Soft Computing} \bibinfo{volume}{167}, \bibinfo{pages}{112258}.
\newblock \DOIprefix\doi{10.1016/j.asoc.2024.112258}.
\bibitem[{Redmon and Farhadi(2017)}]{Redmon2017yolo}
\bibinfo{author}{Redmon, J.}, \bibinfo{author}{Farhadi, A.}, \bibinfo{year}{2017}.
\newblock \bibinfo{title}{Yolo9000: Better, faster, stronger}, in: \bibinfo{booktitle}{2017 IEEE Conference on Computer Vision and Pattern Recognition (CVPR)}, \bibinfo{publisher}{IEEE}. p. \bibinfo{pages}{6517–6525}.
\newblock \DOIprefix\doi{10.1109/cvpr.2017.690}.
\bibitem[{Sam Chandra~Bose et~al.(2024)Sam Chandra~Bose, Srinivasan and Immaculate~Joy}]{SCBose2024}
\bibinfo{author}{Sam Chandra~Bose, A.}, \bibinfo{author}{Srinivasan, C.}, \bibinfo{author}{Immaculate~Joy, S.}, \bibinfo{year}{2024}.
\newblock \bibinfo{title}{Optimized feature selection for enhanced accuracy in knee osteoarthritis detection and severity classification with machine learning}.
\newblock \bibinfo{journal}{Biomedical Signal Processing and Control} \bibinfo{volume}{97}, \bibinfo{pages}{106670}.
\newblock \DOIprefix\doi{10.1016/j.bspc.2024.106670}.
\bibitem[{Selvaraju et~al.(2019)Selvaraju, Cogswell, Das, Vedantam, Parikh and Batra}]{Selvaraju2020gradcam}
\bibinfo{author}{Selvaraju, R.R.}, \bibinfo{author}{Cogswell, M.}, \bibinfo{author}{Das, A.}, \bibinfo{author}{Vedantam, R.}, \bibinfo{author}{Parikh, D.}, \bibinfo{author}{Batra, D.}, \bibinfo{year}{2019}.
\newblock \bibinfo{title}{Grad-cam: Visual explanations from deep networks via gradient-based localization}.
\newblock \bibinfo{journal}{International Journal of Computer Vision} \bibinfo{volume}{128}, \bibinfo{pages}{336–359}.
\newblock \DOIprefix\doi{10.1007/s11263-019-01228-7}.
\bibitem[{Shaw(2024)}]{ShawKO}
\bibinfo{author}{Shaw, D.}, \bibinfo{year}{2024}.
\newblock \bibinfo{title}{Osteoarthritis}.
\newblock \DOIprefix\doi{10.21227/mszn-gr21}.
\bibitem[{Sohail et~al.(2025)Sohail, Azad and Kim}]{Sohail2025tl}
\bibinfo{author}{Sohail, M.}, \bibinfo{author}{Azad, M.M.}, \bibinfo{author}{Kim, H.S.}, \bibinfo{year}{2025}.
\newblock \bibinfo{title}{Knee osteoarthritis severity detection using deep inception transfer learning}.
\newblock \bibinfo{journal}{Computers in Biology and Medicine} \bibinfo{volume}{186}, \bibinfo{pages}{109641}.
\newblock \DOIprefix\doi{10.1016/j.compbiomed.2024.109641}.
\bibitem[{Song et~al.(2024)Song, Du, Yan, Li, Shou, Lai, Fan and Xu}]{Song2024unitpathssl}
\bibinfo{author}{Song, Z.}, \bibinfo{author}{Du, P.}, \bibinfo{author}{Yan, J.}, \bibinfo{author}{Li, K.}, \bibinfo{author}{Shou, J.}, \bibinfo{author}{Lai, M.}, \bibinfo{author}{Fan, Y.}, \bibinfo{author}{Xu, Y.}, \bibinfo{year}{2024}.
\newblock \bibinfo{title}{Nucleus-aware self-supervised pretraining using unpaired image-to-image translation for histopathology images}.
\newblock \bibinfo{journal}{IEEE Transactions on Medical Imaging} \bibinfo{volume}{43}, \bibinfo{pages}{459–472}.
\newblock \DOIprefix\doi{10.1109/tmi.2023.3309971}.
\bibitem[{Tao(2023)}]{TaoOP}
\bibinfo{author}{Tao, M.}, \bibinfo{year}{2023}.
\newblock \bibinfo{title}{Osteoarthritis prediction}.
\newblock \DOIprefix\doi{10.21227/EWX7-B315}.
\bibitem[{Tarvainen and Valpola(2017)}]{Tarvainen2017mt}
\bibinfo{author}{Tarvainen, A.}, \bibinfo{author}{Valpola, H.}, \bibinfo{year}{2017}.
\newblock \bibinfo{title}{Mean teachers are better role models: Weight-averaged consistency targets improve semi-supervised deep learning results}, in: \bibinfo{booktitle}{Proceedings of the 31st International Conference on Neural Information Processing Systems}, p. \bibinfo{pages}{1195–1204}.
\bibitem[{Teoh et~al.(2023)Teoh, Othmani, Lai, Goh and Usman}]{Teoh2023dhl}
\bibinfo{author}{Teoh, Y.X.}, \bibinfo{author}{Othmani, A.}, \bibinfo{author}{Lai, K.W.}, \bibinfo{author}{Goh, S.L.}, \bibinfo{author}{Usman, J.}, \bibinfo{year}{2023}.
\newblock \bibinfo{title}{Stratifying knee osteoarthritis features through multitask deep hybrid learning: Data from the osteoarthritis initiative}.
\newblock \bibinfo{journal}{Computer Methods and Programs in Biomedicine} \bibinfo{volume}{242}, \bibinfo{pages}{107807}.
\newblock \DOIprefix\doi{10.1016/j.cmpb.2023.107807}.
\bibitem[{Wang et~al.(2025)Wang, Cao, Lu, Pan, Tao, Huang, Wang, Huo and Wu}]{Wang2025oag}
\bibinfo{author}{Wang, F.}, \bibinfo{author}{Cao, Y.}, \bibinfo{author}{Lu, H.}, \bibinfo{author}{Pan, Y.}, \bibinfo{author}{Tao, Y.}, \bibinfo{author}{Huang, S.}, \bibinfo{author}{Wang, J.}, \bibinfo{author}{Huo, L.}, \bibinfo{author}{Wu, J.}, \bibinfo{year}{2025}.
\newblock \bibinfo{title}{Osteoarthritis incidence trends globally, regionally, and nationally, 1990–2019: An age‐period‐cohort analysis}.
\newblock \bibinfo{journal}{Musculoskeletal Care} \bibinfo{volume}{23}.
\newblock \DOIprefix\doi{10.1002/msc.70045}.
\bibitem[{Wang et~al.(2020)Wang, Wang, Du, Yang, Zhang, Ding, Mardziel and Hu}]{Wang2020scorecam}
\bibinfo{author}{Wang, H.}, \bibinfo{author}{Wang, Z.}, \bibinfo{author}{Du, M.}, \bibinfo{author}{Yang, F.}, \bibinfo{author}{Zhang, Z.}, \bibinfo{author}{Ding, S.}, \bibinfo{author}{Mardziel, P.}, \bibinfo{author}{Hu, X.}, \bibinfo{year}{2020}.
\newblock \bibinfo{title}{Score-cam: Score-weighted visual explanations for convolutional neural networks}, in: \bibinfo{booktitle}{2020 IEEE/CVF Conference on Computer Vision and Pattern Recognition Workshops (CVPRW)}, \bibinfo{publisher}{IEEE}. pp. \bibinfo{pages}{24--25}.
\newblock \DOIprefix\doi{10.1109/cvprw50498.2020.00020}.
\bibitem[{Wang et~al.(2022)Wang, Song, Wang, Rao, Wang and Wang}]{Wang2022MI-Se}
\bibinfo{author}{Wang, Y.}, \bibinfo{author}{Song, D.}, \bibinfo{author}{Wang, W.}, \bibinfo{author}{Rao, S.}, \bibinfo{author}{Wang, X.}, \bibinfo{author}{Wang, M.}, \bibinfo{year}{2022}.
\newblock \bibinfo{title}{Self-supervised learning and semi-supervised learning for multi-sequence medical image classification}.
\newblock \bibinfo{journal}{Neurocomputing} \bibinfo{volume}{513}, \bibinfo{pages}{383–394}.
\newblock \DOIprefix\doi{10.1016/j.neucom.2022.09.097}.
\bibitem[{Wu et~al.(2023)Wu, Hu, Yue, Li, Simic, Li, Xiang and Wang}]{Wu2023}
\bibinfo{author}{Wu, W.}, \bibinfo{author}{Hu, K.}, \bibinfo{author}{Yue, W.}, \bibinfo{author}{Li, W.}, \bibinfo{author}{Simic, M.}, \bibinfo{author}{Li, C.}, \bibinfo{author}{Xiang, W.}, \bibinfo{author}{Wang, Z.}, \bibinfo{year}{2023}.
\newblock \bibinfo{title}{Self-supervised multimodal fusion network for knee osteoarthritis severity grading}, in: \bibinfo{booktitle}{2023 International Conference on Digital Image Computing: Techniques and Applications (DICTA)}, \bibinfo{publisher}{IEEE}. p. \bibinfo{pages}{57–64}.
\newblock \DOIprefix\doi{10.1109/dicta60407.2023.00017}.
\bibitem[{Zagoruyko and Komodakis(2016)}]{Zagoruyko2016wresnet}
\bibinfo{author}{Zagoruyko, S.}, \bibinfo{author}{Komodakis, N.}, \bibinfo{year}{2016}.
\newblock \bibinfo{title}{Wide residual networks}, in: \bibinfo{booktitle}{Procedings of the British Machine Vision Conference 2016}, \bibinfo{publisher}{British Machine Vision Association}. pp. \bibinfo{pages}{87.1--87.12}.
\newblock \DOIprefix\doi{10.5244/c.30.87}.
\bibitem[{Zhuang et~al.(2023)Zhuang, Si, Wang, Xuan, Ouyang, Zhan, Xue, Zhang, Shen, Yao and Wang}]{Zhuang2023kcd}
\bibinfo{author}{Zhuang, Z.}, \bibinfo{author}{Si, L.}, \bibinfo{author}{Wang, S.}, \bibinfo{author}{Xuan, K.}, \bibinfo{author}{Ouyang, X.}, \bibinfo{author}{Zhan, Y.}, \bibinfo{author}{Xue, Z.}, \bibinfo{author}{Zhang, L.}, \bibinfo{author}{Shen, D.}, \bibinfo{author}{Yao, W.}, \bibinfo{author}{Wang, Q.}, \bibinfo{year}{2023}.
\newblock \bibinfo{title}{Knee cartilage defect assessment by graph representation and surface convolution}.
\newblock \bibinfo{journal}{IEEE Transactions on Medical Imaging} \bibinfo{volume}{42}, \bibinfo{pages}{368–379}.
\newblock \DOIprefix\doi{10.1109/tmi.2022.3206042}.

\end{thebibliography}

\end{document}